\documentclass[acmtog]{acmart}

\AtBeginDocument{%
  }

\acmSubmissionID{381}
\usepackage[inline]{enumitem}
\usepackage[tableposition=top]{caption}
\usepackage{subcaption}
\usepackage{tabularx}
\usepackage{xcolor}
\begin{document}

\title{Neural Parameterization for Dynamic Human Head Editing}
%   \footnote{Note that the field $V$ is not strictly a parameterization, but we use the terminology to indicate the operation for unwrap a 3D surface to a 2D texture.}

% \affiliation{%

\author{Li Ma}
\affiliation{%
  \institution{The Hong Kong University of Science and Technology}
  \city{Hong Kong}
  \country{China}}
\email{lmaag@connect.ust.hk}
\authornote{Author did this work during the internship at Tencent AI Lab.}

\author{Xiaoyu Li}
\affiliation{%
  \institution{Tencent AI Lab}
  \city{Shenzhen}
  \country{China}}
\email{lixiaoyu306@gmail.com}

\author{Jing Liao}
\affiliation{%
  \institution{City University of Hong Kong}
  \city{Hong Kong}
  \country{China}}
\email{jingliao@cityu.edu.hk}

\author{Xuan Wang}
\affiliation{%
  \institution{Tencent AI Lab}
  \city{Shenzhen}
  \country{China}}
\email{xwang.cv@gmail.com}

\author{Qi Zhang}
\affiliation{%
  \institution{Tencent AI Lab}
  \city{Shenzhen}
  \country{China}}
\email{nwpuqzhang@gmail.com}

\author{Jue Wang}
\affiliation{%
  \institution{Tencent AI Lab}
  \city{Shenzhen}
  \country{China}}
\email{arphid@gmail.com}

\author{Pedro V. Sander}
\affiliation{%
  \institution{The Hong Kong University of Science and Technology}
  \city{Hong Kong}
  \country{China}}
\email{psander@cse.ust.hk}

\renewcommand{\shortauthors}{Li et al.}

\begin{abstract}
 Implicit radiance functions emerged as a powerful scene representation for reconstructing and rendering photo-realistic views of a 3D scene. These representations, however, suffer from poor editability. On the other hand, explicit representations such as polygonal meshes allow easy editing but are not as suitable for reconstructing accurate details in dynamic human heads, such as fine facial features, hair, teeth, and eyes. In this work, we present Neural Parameterization (NeP), a hybrid representation that provides the advantages of both implicit and explicit methods. NeP is capable of photo-realistic rendering while allowing fine-grained editing of the scene geometry and appearance. We first disentangle the geometry and appearance by parameterizing the 3D geometry into 2D texture space. We enable geometric editability by introducing an explicit linear deformation blending layer. The deformation is controlled by a set of sparse key points, which can be explicitly and intuitively displaced to edit the geometry. For appearance, we develop a hybrid 2D texture consisting of an explicit texture map for easy editing and implicit view and time-dependent residuals to model temporal and view variations. We compare our method to several reconstruction and editing baselines. The results show that the NeP achieves almost the same level of rendering accuracy while maintaining high editability.
\end{abstract}


\begin{teaserfigure}
 \vspace{-4mm}
  \includegraphics[width=0.97\textwidth]{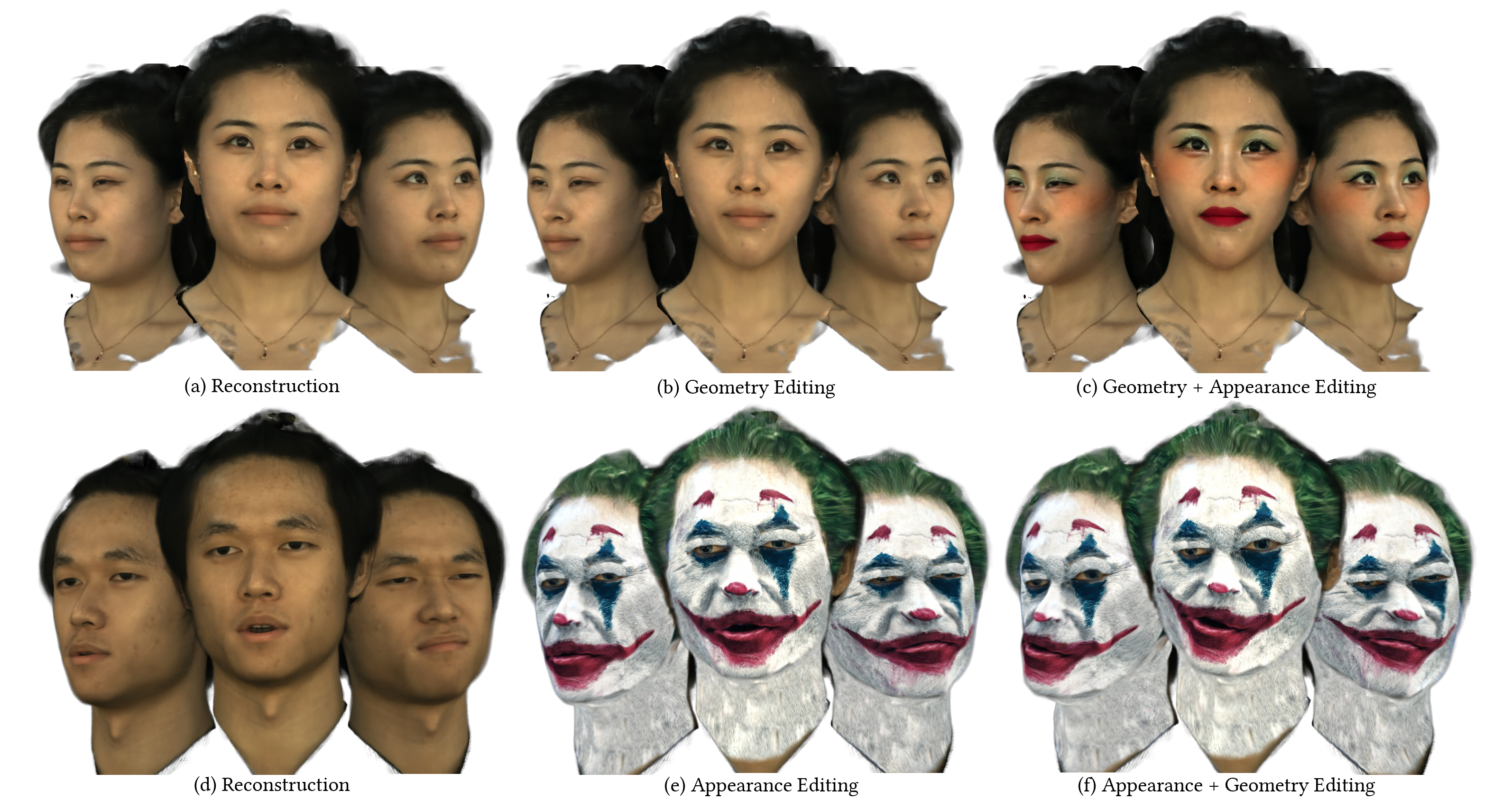}
  \vspace{-3mm}
  \caption{Given synchronized multi-view videos, our method reconstructs a volumetric representation (a,d) that enables geometry and appearance editing. We demonstrate applications of our method in geometry editing (b) appearance editing (e),  and joint editing (c,f). }
  \Description{Description of Neural Parameterization. }
  \label{fig:teaser}
\end{teaserfigure}

\maketitle

\section{Introduction}
% Current Background
%% implicit representation: pros: 1. compact, arbitrary resolution without constrains from the discrete grid 2. model complex scene as continuous function, 3. DR support, photorealistic.
Modeling, editing and rendering photo-realistic scenes have a wide range of applications in computer games, movie industry, virtual and augmented reality. Traditionally, polygonal meshes combined with texture maps have been adopted as the standard 3D representations in rendering pipelines. Since the geometry and appearance are explicitly modeled, editing can be done naturally by deforming the meshes or modifying the texture maps. However, despite the long-lasting development of mesh reconstruction methods~\cite{mesh_sfm_colmap1,mesh_sfm_deepsfm,mesh_sfm_pmvsnet,mesh_sfm_deepmvs}, it is still challenging to reconstruct accurate, well-aligned meshes for scenes like dynamic human heads from video, especially at the places of hair, teeth, and the eyes, which exhibit complex geometry and appearance.

To break through the limited representation capacity of mesh-based methods, implicit volumetric representations have gained considerable attention recently for photo-realistic rendering. In particular, neural radiance field (NeRF) \cite{NeRF} models scenes as some continuous functions defined densely in a 3D volume. By parameterizing the function using a multilayer perceptron (MLP), scenes are implicitly encoded into the parameters of the neural networks, which is more compact than explicitly storing values in a dense 3D grid, while allowing the volume to model complex scenes at arbitrary resolution. However, encoding the entire scene into the parameters of the neural networks implicitly makes this representation more obscure for interpreting and editing, which greatly restricts the practical usage of these methods.

In this work, we address the challenge of reconstructing high-fidelity dynamic scenes with implicit representations while allowing explicit editing of both the geometry and the appearance. Existing works usually train implicit representations that is conditioned on some latent codes. Scene editing can then be implemented by propagating the editing from the rendered results back to the latent codes \cite{NeRF_edit_cond,NeRF_edit_Deformed}, or using another encoder network to directly map the editing target to the parameters \cite{NeRF_edit_clip,NeRF_edit_styleHyper}. Similar ideas have also been introduced into implicit human head representations \cite{NeRF_edit_FENeRF,NeRFace_dynamic,NeRFace_headnerf,NeRFace_ImFace,NeRFace_IMAvatar}. However, latent-code-based editing usually does not allow fine-grained and out-of-domain editing.

% Moreover, the generative model usually requires a large collection of dataset to find a plausible distribution of the scene.
% The implicit representations can also be extended to model dynamic scene by introducing deformable field \cite{NeRF_dyn_nerfies,NeRF_dyn_dnerf,NeRF_dyn_neuralsceneflow} or time embedding \cite{NeRF_dyn_3Dvideo}. 

NeuTex~\cite{NeRF_neutex} proposes to model the appearance of implicit scene representation using 2D texture maps. The appearance of the scene can then be edited by modifying the 2D texture maps. However, NeuTex only focuses on static scenes, and how to edit the scene geometry is also unclear. Neural Atlas \cite{misc_neuralatlas} introduces a similar idea for consistent video editing. It synthesizes each frame of a video by mapping each pixel location to texture atlas space shared by the entire video sequence. The video can then be consistently modified by editing the texture atlas. %Editing on texture maps allows for fine-grained modification of the appearance. 
We adopt the same idea for appearance modeling and editing.

%we develop a novel scene representations based on implicit volume, while introduce some ideas from the explicit representation and thus improve the editability of the neural volumetric representation.

% In this work, ...
In this work, we present Neural Parameterization (NeP), a volumetric hybrid implicit-explicit 3D representation for dynamic human heads. NeP achieves photo-realistic rendering while allowing for explicit editing of both geometry and appearance. 
%% decompose
We draw inspiration from traditional texture mapping and propose to decompose a 3D dynamic human head into three components: a density volume, a UV volume and a 2D texture. The decomposition enables the disentanglement of geometry and appearance, which allows us to edit them separately.  
%% merge implicit with explicit
The density volume, UV volume and the texture are all modeled implicitly as MLPs in the first place to improve the reconstruction quality, and to reduce the memory requirement when modeling the dynamic view-dependent volume. To enable explicit editing of both geometry and appearance, we introduce explicit layers into the implicit MLPs for these two components. The explicit geometry layer is modeled as a temporally-varying 3D warping field controlled by semantic keypoints. This allows us to freely deform the geometry on one frame and propagate the deformation to the other frames for consistent geometry editing. For appearance, the editability is provided by using an explicit texture map that is shared by all the frames. By editing the texture map, we achieve consistent appearance editing for different views and time.

%% regularizations
We reconstruct the NeP from multi-view videos captured by a set of synchronized and calibrated cameras. Freely optimizing the hybrid representation using the multi-view videos will lead to sub-optimal solutions that is unsuitable for the purpose of editing, so we design several regularization methods and a two-stage optimizing strategy in the training process. 
%% experiments
%We demonstrate that the NeP achieves photo-realistic rendering results and outperforms several baselines in terms of editing. 
We also develop a user interface that enables interactive editing and preview using an extracted mesh representation. Extensive experiments show that NeP achieves photo-realistic rendering results and outperforms several baselines in terms of editing capability. 
% Our key contributions are
To summarize, our key technical contributions are:
\begin{itemize}
\item A novel hybrid representation NeP that enables intuitive and consistent editing on both geometry and appearance of dynamic human heads.
\item Regularization methods and a training strategy that greatly improves the editability of NeP while maintaining a high-fidelity reconstruction of human heads.
% \item We provide a UI for interactive editing and preview.
% \item An effective user interface for interactive geometry and appearance editing.
\item Several applications are made possible in 3D by using our method, like virtual makeup, artistic stylization and face shape editing, with a UI for interactive editing.

\end{itemize}

\section{Related work}
Since our editing method is based on a new scene representation, we review existing works on related scene representations, from explicit mesh to the current state-of-the-art implicit neural representation. In particular, we will focus more on 3D facial reconstruction and editing methods since our main interest is 3D human heads. 

%% mesh  reconstruct and editing, 
\paragraph{Mesh-Based Reconstruction and Editing.}
%%% mesh is explicit
%%% mesh reconstruction: 1. classical 2. DR 3. 3DMM as prior 4. appearance: texture + neural
%%% mesh + editing
%%% mesh + appearance model (texture + neural appearance)
The polygon mesh is the most commonly used geometry representation in real-time applications, such as in the game and movie industry. Mesh-based methods explicitly model the geometry of 3D surfaces. To reconstruct a 3D mesh from images captured from the real world, classical methods typically use structure from motion \cite{mesh_sfm_colmap1,mesh_sfm_colmap2,mesh_sfm_meshroom} to get a dense point cloud, and then convert it to the triangle mesh using surface reconstruction algorithms \cite{mesh_ballpivot,mesh_poisson}. To improve the reconstruction quality, there are several attempts that introduce differentiable rendering to the mesh rendering, such as rasterization \cite{mesh_dr1_opendr,mesh_dr1_softrasterizer,mesh_dr1_NMR,mesh_dr1_pytorch3d,mesh_dr1_sampling} or ray tracing \cite{mesh_dr2_MCraytracing,mesh_dr2_repearameterizing,mesh_dr2_svbrdf}. By making the rendering process differentiable, the loss can be back propagated from the rendering results to the scene parameters such as vertex positions. 

%Some works exploit signed distance field (SDF) as intermediate representation \cite{mesh_sfm_voxelhashing,sdf_DVR,sdf_IDR,sdf_NeuS,sdf_volsdf} and run ray marching \cite{sdf_mcube_deep,sdf_mcube_neural} to get the mesh representation. 

The human face is a strong prior that can be used to improve the reconstruction quality. The prior is usually modeled using a 3D morphable model (3DMM) \cite{3dmm_0,3dmm_bfm,3dmm_bfm2017,3dmm_LSFM,3dmm_FLAME}. The geometry of the face is determined by morphing a pre-defined template face mesh. In a typical linear 3DMM, the morphing is parameterized as a linear combination of a fixed set of basis offsets. By learning a set of principle basis offsets from a large 3D face dataset, the 3DMM model can be used to generate arbitrary faces using the coefficient vector. The problem of mesh reconstruction is then simplified to computing the coefficient vector of the face. The reconstruction of the 3DMM can be modeled as an analysis-by-synthesis problem using differentiable rendering \cite{3dmm_dr_unsupervised,3dmm_dr_faceverse,3dmm_dr_DECA,3dmm_dr_ganfit}, or directly regressed using deep neural networks \cite{3dmm_direct_3DDFAv2}. 
% To improve the details of the reconstructed mesh and texture, several works learn to produce high resolution displacement map and normal map alongside the 3DMM model. 
%Apart from 3DMM, the prior of the face can also be modeled using neural networks \cite{3dmm_nn_prnet,3dmm_nn_production,3dmm_nn_pixelcodec,3dmm_nn_tofu}. 

% Since the fitted mesh is morphed from the same template with same topology, the reconstructed meshes are densely aligned, making the editing of the mesh possible. 
With accurate reconstructed geometry, the mesh-based model can achieve impressive rendering results. Moreover, due to the explicitness of this representation, meshes with texture maps can be edited naturally. Many tools have been developed to edit the mesh representation, such as mesh deformation \cite{mesh_editing_asap}, smoothing \cite{mesh_editing_laplacian}, subdivision \cite{mesh_editing_catmullclark}. Unfortunately, despite the progress in generic surface and face reconstruction, an accurate mesh is still hard to acquire for human heads, especially for hair, eyes, and the mouth interior \cite{volume_MVP}. One option to compensate for the inaccurate reconstruction is to use neural texture representations \cite{mesh_appear_deferred,mesh_appear_freevs,mesh_appear_stablevs,3dmm_nn_pixelcodec} or deep appearance models \cite{mesh_appear_DAM,mesh_appear_DRAM}. However, the use of implicit textures corrupts the editability of the representation. 

%% Volumetric Representation, good reconstruction, poor editability
%% MPI, NV, MVP,
%% editing of volume is still unknown.
\paragraph{Volumetric Representation.}
Volumetric representation models the scene by densely storing parameters in a 3D regular grid, such as signed distance \cite{volume_sdf_mvsmachine}, colors \cite{volume_NV} and radiance function \cite{volume_plenoxels}. Neural Volumes \cite{volume_NV} use a decoder structure to produce a volume on the fly from a latent code. MVP \cite{volume_MVP} reduces the memory requirement by attaching small volumetric primitives on a coarse mesh proxy, so the empty space is not modeled. One variation is multiplane images (MPIs) \cite{MPI_stereomag,MPI_deepview,MPI_pushbound,MPI_LLFF}, which models the scene using several fronto-parallel RGB$\alpha$ image planes in the frustum of a reference camera. Unlike meshes, volumetric representation is able to model complex geometry like thin structures. Besides, the rendering of volumetric representation is naturally differentiable, which makes it easy to optimize a volume that achieves photorealistic rendering results. One drawback of the volumetric representation is the large memory consumption for densely storing the scene parameters, especially for dynamic scenes. In terms of editing, directly manipulating each voxel would be tedious and error-prone. There is a lack of more advanced tools to assist the editing of the volumetric representation.
%% implicit representation 
%%% 1. nerf, sdf, reflective 2. to model dynamic scene 3. poor editability, editable nerf (e.g. editable nerf, 3D GAN nerf, scene graph based, grid based NeRF, neutex 

\paragraph{Implicit Representation.}
Recently, many works propose to implicitly model a volume function by coordinate-based Multi-Layer Perceptrons (MLPs) instead of explicitly storing values in a regular grid. The implicit representation can be used to model signed distance function \cite{sdf_IDR,sdf_DVR,sdf_volsdf,sdf_NeuS}, occupancy function \cite{occupancynetwork}, radiance fields \cite{NeRF}, BRDF parameters \cite{nerf_relit_nerd,nerf_relit_nerv,nerf_relit_reflectance}. Especially, NeRF \cite{NeRF} proposes to model the 3D scene by mapping the spatial location and view direction to the density and color of the point. By optimizing the neural network using multi-view input images, NeRF achieves impressive novel view synthesize quality. To extend the NeRF to represent dynamic scenes, modulation-based methods \cite{NeRF_dyn_3Dvideo, xian2021space} directly provide time information to the MLP, while deformation-based methods \cite{NeRF_dyn_dnerf,NeRF_dyn_nerfies} introduce implicit deformation field to non-rigidly warp a template NeRF. \citet{NeRF_dyn_hypernerf} proposes to combine the two methods, where the template NeRF is also conditioned on the time information. 
% implicit + human faces
Implicit representation has also been used to help reconstruct the human head. A common approach is to condition the implicit field on some latent vectors, such as geometry and appearance code \cite{NeRFace_deform}, expression and identity code \cite{NeRFace_ImFace}, latent code sampled from a prior distribution \cite{NeRFace_H3D} or feature vectors \cite{zhang2022fdnerf}. HeadNeRF \cite{NeRFace_headnerf} reconstruct NeRFs for human heads, which is also conditioned on the identity, expression, and appearance parameters. \citet[]{NeRFace_dynamic} proposes to reconstruct a dynamic facial avatar by training a NeRF that depends on pre-estimated expression parameters.

One problem of the implicit representation is that it encodes the scene into the parameters of the neural networks, which makes it extremely difficult to edit directly. To edit the implicit representation, \cite{NeRF_edit_cond} train a NeRF conditioned on latent code and then fine-tune the latent code based on the editing target. This takes a large amount of time, even for a simple editing operation. CLIP-NeRF \cite{NeRF_edit_clip} introduces an image or text encoder that directly maps the editing target to the latent code. UV Volumes~\cite{chen2022uv} adopts feature volumes and neural texture stack for editing 3D humans. Other GAN-based NeRF methods \cite{NeRF_GAN_giraffe,NeRF_GAN_pigan,NeRF_GAN_stylenerf,NeRF_GAN_EG3D,NeRF_GAN_graf,NeRF_IDE3D} also have the potential to be manipulated by GAN inversion \cite{2D_GANinversion}. However, editing the latent code could not achieve fine-grained, and out-of-domain edits \cite{NeRF_edit_clip}. NeuTex \cite{NeRF_neutex} introduces the idea of texture mapping from the mesh-based representation into the implicit representation. The appearance of the scene can be edited by manipulating the texture map. This allows fine-grained editing, but the editing is limited to the static objects, and how to edit the geometry of the scene is not explored. In this work, we propose a method that can edit both the geometry and appearance of dynamic human heads.

%Another scheme to edit the implicit representation is by combining explicit components. For instance, \citet{NeRF_edit_scenegraph} and \citet{NeRF_edit_object} subdivide the whole scene into individual components, so that we can easily modify the scene by rearrange each component. ST-NeRF \cite{NeRF_edit_layered} further extends this idea to free-viewpoint videos. Editing per object is still a open question. 

% Several works that use feature grid to achieve faster training by reshuffling the grid. 

\begin{figure*}
    \centering
    \includegraphics[width=0.97\textwidth]{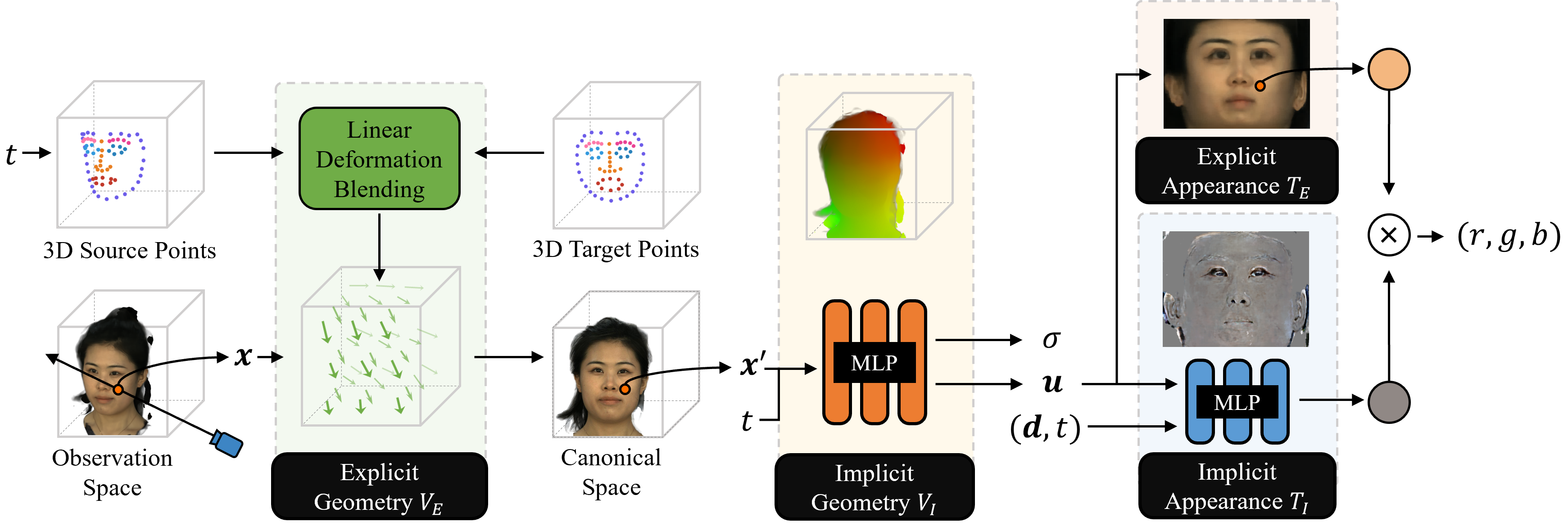}
    \vspace{-3mm}
    \caption{Our hybrid scene representation. We disentangle a dynamic radiance field to a geometry representation and appearance representation. The geometry is composed of an explicit deformation field controlled by sparse control points, and an implicit UV and density field that maps any 3D position to a 2D texture coordinate. The appearance is made up of an explicit texture map shared by all the frames, and a MLP that models all the view and time dependent effects on the map. }
    \vspace{-1mm}
    \label{fig:pipeline}
\end{figure*}

\section{Overview}
% two parts, geometry and appearance
% formally, the geometry: x, t, v->sigma, uv,  appearance: uv=>rgb
% To enable edit, we introduce explicit layers
% supervise using mse and reg. 
% to edit, ..., 
We propose a hybrid implicit and explicit representation to model a dynamic 3D volume of a human head. The blueprint is that the fundamental representation is implicit, so that it can achieve photorealistic reconstruction results while maintaining a certain level of editability through the introduction of explicit layers. Our goal is to model a volume that defines a dynamic 3D radiance field $F$ that maps the spatial 3D location $\mathbf{x} =(x,y,z) \in \mathbb{R}^3$, view direction $\mathbf{d} = (\theta, \phi) \in \mathbb{R}^2$ and time index $t \in \mathbb{Z}$ to a single RGB value $\mathbf{c} \in \mathbb{R}^3$, as well as the volume density $\sigma \in \mathbb{R}$. Formally,
\begin{equation}
    (\mathbf{c}, \sigma) = F(\mathbf{x}, \mathbf{d}, t).
\end{equation}
The time index $t$ is encoded into a latent code of length $256$ through a learnable embedding layer as proposed by \citet[]{dict_embed}. We apply Fourier position encoding to $\mathbf{x}$ and $\mathbf{d}$ to increase the model capacity of the implicit representation as illustrated in \cite{fourier_embed}. 

To be able to edit both the geometry and appearance of the human head, we disentangle the representation into the appearance representation, namely a neural texture $T$, as well as the geometry representation, which is essentially a UV and density field $V$:
\begin{equation}
    (\mathbf{u}, \sigma) = V(\mathbf{x}, t),  \mathbf{c} = T(\mathbf{u}, \mathbf{d}, t).
\end{equation}
where $\mathbf{u} = (u,v) \in  \mathbb{R}^2$ is the UV coordinate in texture space. % TODO 
We further decompose the geometry representation $V$ and the neural texture $T$ into implicit parts $V_I$ and $T_I$ for implicit reconstruction and explicit modules $V_E$ and $T_E$ for explicit editing. Fig. \ref{fig:pipeline} shows the system overview. We model the geometry using an explicit warping field cascaded with an implicit UV and density field, while the appearance is a two layered structure that is composed of an explicit texture map and a view and time dependent implicit texture.
% \remark{(xiaoyu: add some illustrations for the overview like the text in the caption)}

To construct such a hybrid representation, we take synchronized multi-view videos of a person's head as our main source of supervision. We also propose several regularizations to achieve better editability. The paper is organized as follows. In Section \ref{sec:data_acquisition}, we illustrate the data acquisition and its pre-processing. In Sections \ref{sec:geometry} and \ref{sec:appearance}, we explain the geometry and appearance aspects of our representation, respectively, and how to optimize the model parameters to achieve better reconstruction quality and editability. We introduce the rendering and training details in Sections 7 and 8. Then in Sections \ref{sec:experiments} and 10, we conduct comparisons and ablations and show related applications. Finally, we discuss limitations and conclude in Section \ref{sec:discussion}.

\section{Data Acquisition}
%% data capture
%% camera pose estimation
%% Face alignment
%% Background removal
\label{sec:data_acquisition}
The input to our system is a set of calibrated and synchronized multi-view videos of a dynamic head. We capture the videos using a camera array with $12$ cameras spread over approximately 120 degrees. We use $11$ views for constructing the model and $1$ view for evaluation. The videos are captured at 30fps and downsampled to the resolution of $512 \times 374$ for training. The camera poses and intrinsics are calibrated using the off-the-shelf Structure from Motion software \cite{mesh_sfm_CR} using the first frame of the videos. We also compute a loose bounding box $\mathcal{B}$ from the reconstructed point cloud as the boundary of the volume. Since our main focus is to model a dynamic head, we mask out the background of each video using the current state-of-the-art video matting algorithm \cite{misc_RVM}. Our method also requires a coarse tracked face mesh to initialize the UV mapping. To that end, we apply PRNet \cite{3dmm_nn_prnet} to predict the projection of densely aligned face mesh in image space for each view. We then construct the face mesh in 3D by minimizing the reprojection error in each view. We repeat the process for each frame to get a sequence of tracked 3D face mesh. From the mesh we extract the position of each vertex $\mathbf{p}_i \in \mathbb{R}^3$ and its corresponding UV coordinate $\mathbf{u}_i \in \mathbb{R}^2$. 
% Our method is robust to inaccurate mesh tracking.
% TODO：experiments that prove our method is robust to inaccurate mesh tracking

\section{Geometry Modeling}
\label{sec:geometry}
Recall that the representation $V$ is a composition of a UV field and a density field that maps the spatial location $\mathbf{x}$ and time $t$ to the UV coordinate $\mathbf{u}$ and volume density $\sigma$.
% \remove{Recall that the geometry representation is a UV field $V$ that maps the spatial location $\mathbf{x}$ and time $t$ to the UV coordinate $\mathbf{u}$ and volume density $\sigma$.} 
With $V$ modeled merely using MLPs, it would be hard to edit the geometry.
% another version
% Several works propose deformation fields for modeling dynamic scenes. The deformation can be modeled using MLPs, \cite{NeRF_dyn_dnerf,NeRF_dyn_nerfies,NeRF_dyn_hypernerf}, mixture of affine \cite{volume_NV,misc_oneshotFV}, or guided by mesh \cite{volume_MVP,misc_animatable}. We adopt the idea by also defining a deformation field $V_e$ that is determined by some explicitly defined parameters. Ideally, the explicit deformation field should have following properties for easy reconstruction and editing: 
Deformation-based NeRFs \cite{NeRF_dyn_dnerf,NeRF_dyn_nerfies,NeRF_dyn_hypernerf} use a deformation field that maps a spatial location $\mathbf{x}$ of one frame $t$ in the observation space $\mathcal{B}$ to $\mathbf{x}'$ in canonical space $\mathcal{B'}$. We adopt this idea by also defining a deformation field $V_E$ that is determined by some explicitly defined parameters. Ideally, the explicit deformation field should have the following properties for easy reconstruction and editing: 
\begin{enumerate*}
    \item To enable volumetric rendering, the field $V_E$ should be densely defined in the bounding box $\mathcal{B}$ for any time index $t$, so that for any query point $\mathbf{x}$, we could find the mapped new location $V_E(\mathbf{x}, t)$. 
    % TODO: whether optimize control points or not
    \item The $V_E$ should be differentiable with respect to the explicit parameters, so the parameters could be jointly optimized. 
    %\item The computation of $V_E$ should be memory and complexity feasible to support densely, since we need to densely sample points along rays dur=ing volumetric rendering. 
    \item The $V_E$ should be locally controlled by a portion of the parameters to allow for fine-grained editing.
\end{enumerate*}

To meet these requirements, we propose a simple but effective explicit deformation method. Specifically, since the head pose change can be modeled as a rigid transformation, we first transform the volume to a stable head space $\mathcal{\bar{B}}$ using a global rigid transformation $\mathbf{H} \in SE(3)$, i.e. $\mathbf{\bar{x}} = \mathbf{H} \mathbf{x}$. We then define the control parameters as set of sparse control points $\{\mathbf{\bar{s}}_i\}$ and corresponding target positions $\{\mathbf{\bar{z}}_i\}$. The explicit deformation field $V_E$ is then formulated as a linear deformation blending: 
\begin{align}
    V_E(\mathbf{x}) &= \mathbf{\bar{x}} + \frac{\sum_i \psi_i(\mathbf{\bar{x}}) (\mathbf{\bar{z}}_i - \mathbf{\bar{s}}_i)} {\sum_i \psi_i(\mathbf{\bar{x}})} \text{, where} \\
    \psi_i(\bar{\mathbf{x}}) &= \exp \left( - (\mathbf{\bar{x}} - \mathbf{\bar{s}}_i)^2 / r_i^2 \right).
\end{align}
Intuitively, $V_E$ is a linear interpolation of the set of displacements $\{\mathbf{\bar{z}}_i - \mathbf{\bar{s}}_i\}$, with the weights determined by a Gaussian radian basis function $\psi_i(\mathbf{\bar{x}})$ that has the influence radius $r_i$. The bar $\bar{\cdot}$ denotes that the control points and target positions are all defined in the stable head space $\mathcal{\bar{B}}$. Note that $V_E$ is differentiable with respect to the parameters $(\mathbf{H}, \mathbf{\bar{z}}_i, \mathbf{\bar{s}}_i, r_i)$, which means that it is possible to jointly optimize the explicit parameters using a stochastic gradient descent (SGD) algorithm. Since we are modeling a dynamic volume, we have the explicit deformation parameters $(\mathbf{H}, \mathbf{\bar{s}}_i, r_i)$ defined in every time index. The target positions $\{\mathbf{\bar{z}}_i\}$ are kept static for different time indexes to serve as a canonical target that is shared by all the frames. By editing the control points, we could explicitly manipulate the 3D deformation field in an intuitive way.
% $(\mathbf{H}^{(t)}, t_i^{(t)}, s_i^{(t)}, r_i^{(t)})$ except for the target

\begin{figure}
    \centering
    \includegraphics[width=\linewidth]{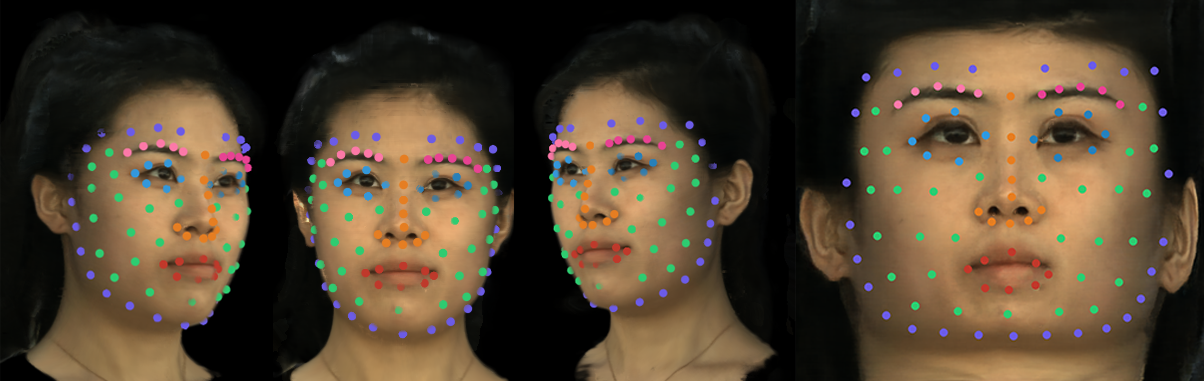}
    \vspace{-5mm}
    \caption{A visualization of the $96$ control points. The control points are manually selected and have rich semantic meaning.}
    \vspace{-2mm}
    \label{fig:control_point_vis}
\end{figure}

To edit the scene easily and consistently, we would like to make the control point semantically consistent. In other words, if each control point $\mathbf{s}_i^{(t)}$ has the same semantic meaning for all frames $t$, then we could edit the entire sequence by editing only one frame, and then propagating the editing result. An example of semantically consistent points would be tracked facial landmarks. In order to construct a set of semantically consistent control points, we predefined a subset of $96$ vertices in the tracked face mesh $\{\hat{\mathbf{s}}_i^{(t)}\}$ in frame $t$ as our initial control points as shown in Fig. \ref{fig:control_point_vis}, and apply a semantic loss over the course of optimization:
\begin{equation}
    \mathcal{L}_{semantic} = \sum_i \sum_t \|{\mathbf{s}_i^{(t)} - \hat{\mathbf{s}}_i^{(t)}}\|_2.
\end{equation}
% TODO: whether to optimize the control points
The reason why we only supervise the control points to be close to the vertices instead of directly using the vertices is that the tracked mesh may be inaccurate, and the deformation field constructed by the vertices may not be optimal. So we finetune the vertex locations using differentiable rendering. The global transformation $\mathbf{H}^{(t)}$ is also initialized from a least squares estimation of all the vertices of the tracked face mesh. $\{\mathbf{z}_i\}$ is fixed to be the same subset of vertices of a predefined canonical 3D face mesh. 

After the explicit deformation field deforms each location to a canonical space, we define our final UV field as:
\begin{equation}
    V = V_I(\mathbf{x'}, t),~\mathbf{x'} = V_E(\mathbf{x}, t).
\end{equation}
where the $V_I(\mathbf{x'}, t)$ is an implicit UV field that maps the canonical location $\mathbf{x}'$ and frame index $t$ to the UV coordinate $\mathbf{u}$ and density $\sigma$. We show that the explicit deformation field not only enables editing the geometry, but also helps to align the volume to a canonical space, which eases the complexity for reconstructing the dynamic head of the implicit representation. We conditioned the $V_I$ also on time index $t$ since the explicit deformation $V_E(\mathbf{x}, t)$ is a smooth and continuous deformation field that warps the geometry to the canonical space. We need $V_I$ to model temporal varying topological changes that require a discontinuity in the deformation field such as eye blink. 
% TODO: the stereoscopic mapping of the UV coordinate

\section{Appearance Modeling}
\label{sec:appearance}
%% about layered texture
Now that we have the UV coordinates, we seek to look up the RGB value given the UV coordinate, view direction $\mathbf{d}$, and time index $t$. We adopt the idea of view dependent textures and dynamic textures by allowing the appearance model $T$ to be conditioned on $\mathbf{d}$ and $t$. The view dependent textures \cite{mesh_appear_DAM,NeRF_neutex} are able to model view dependent effects such as specularity on the eyes, while compensating for inaccurate geometry and improving the realism of the rendering results. Our geometry representation $V$ is able to model the pose and the expression change of the head, and also fit most of the skin motions, such as mouth opening and eye blinking. However, dynamic textures \cite{mesh_dynamicasset,mesh_neuralasset,3dmm_dr_facescape,mesh_binocular,R2_practical,R2_DTexFusion} are still needed because not all temporal variations can be modeled by a dynamic UV field, such as the sudden appearance of ambient occlusion caused by wrinkles.

The view and time dependent textures can be simply implemented using implicit representations, but this will leads to poor editability. A desired texture should be simply a single static image, so that by editing the image, every frame of the 3D volume will be consistently edited. Therefore, we use a two-layer representation that combines the explicit and implicit texture maps. The main part of the texture is stored as an explicit texture map $T_E(\mathbf{u})$, while the view changes and temporal variations of the texture are modulated as residuals $T_I(\mathbf{u}, \mathbf{d}, t)$. Thus, the appearance component is computed as:
\begin{equation}
    T(\mathbf{u}, \mathbf{d}, t) = T_E(\mathbf{u}) * exp(T_I(\mathbf{u}, \mathbf{d}, t)).
\end{equation}
The lookup $T_E$ for an arbitrary floating point UV coordinate is performed using bilinear interpolation and the $*$ is the element-wise product that is individually applied on each RGB color channel. The use of $exp(\cdot)$ guarantees that the multiplier is greater than zero. $T_I$ can be interpreted as modulating the local lighting changes of the texture map. We would like the implicit texture map to only model the residual, thus we apply a sparsity loss:
\begin{equation}
    \mathcal{L}_{sparsity} = \sum_k |T_I(\mathbf{u}_k, \mathbf{d}_k, t)|,
\end{equation}
where $k$ is the index of the sampled point during rendering for a specific time $t$. This loss encourages the multiplier to be close to $1$, so that the main contribution of the final result is from the explicit texture map $T_E$. 
% TODO: We also find in experiments that this loss also encourages the texture map to be better aligned along time.
% TODO: There are multiple ways to combining the $T_E$ and $T_I$. But as shown in the experiment, the multiply operation has the best interpretability.

\section{Rendering}
Given a radiance field $F$ at frame $t$, we render a pixel using volume rendering \cite{volume_render}. Specifically, we first shoot rays from the camera center to the pixel locations and check whether the ray hit the volume boundary $\mathcal{B}$ using axis-aligned bounding boxes (AABB). Then, if the ray passes the AABB test, we then sample $N$ points along the ray between the hit near and far points. The final color is computed as:
\begin{equation}
    \mathbf{\hat{c}} = \sum_{i=1}^N T_i (1 - exp(-\sigma_i \delta_i)) \mathbf{c}_i \text{, where } T_i = exp\left(-\sum_{j=1}^{i-1} \sigma_j \delta_j \right)
\end{equation}
The terms $\sigma_i$ and $\mathbf{c}_i$ are the density and color of $i$-th point, and $\delta_i$ is the distance between adjacent samples. We adopt the same hierarchical volume sampling strategy as in \cite{NeRF}, which first uses $64$ stratified sample points on a coarse volume $V_{coarse}$, followed by another round of importance sampling of $64$ points on a fine volume $V_{fine}$. 

\section{Training}
\label{sec:training}
To reconstruct the representation $F$, we adopt an analysis by synthesis approach and train the model using Adam \cite{misc_adam} optimizer for 120k iterations, which takes about one day on three V100 GPUs. In each iteration, we randomly select a time index, and then randomly sample $B = 6000$ rays for optimization. Next, we describe the losses and the two-stage training process that we use to train our representation.
% main supervision is mse
% unsupervised texture unwarp
% alpha loss
% two stage training for alignment
\subsection{Main Supervision}
Our main source of supervision is the pixel reconstruction loss between the rendering result and the input images. 
\begin{equation}
    \mathcal{L}_{MSE} =  \sum_i^B \left(\|\mathbf{c}_i - \hat{\mathbf{c}}_i\|_2 + \|\alpha_i - \hat{\alpha}_i\|_2 \right).
\end{equation}
Note that we also supervise the rendered foreground mask to be close to the input mask, which is computed by using the accumulated transmittance. This prevents the network from using the background to fake shading effects that should be in the texture \cite{NeRF_neutex}.

\subsection{Texture Unwrap Regularization}
Given only the pixel reconstruction supervision, the radiance field is able to achieve good reconstruction quality. However, we find that freely optimizing the UV field and the texture will result in a noisy UV field that leads to poor editability. Thus, we introduce several regularizations to ensure the implicit representation is optimized to be suitable for editing. 

\begin{figure}
    \centering
    \begin{subfigure}{0.325\linewidth}
    \includegraphics[width=\linewidth]{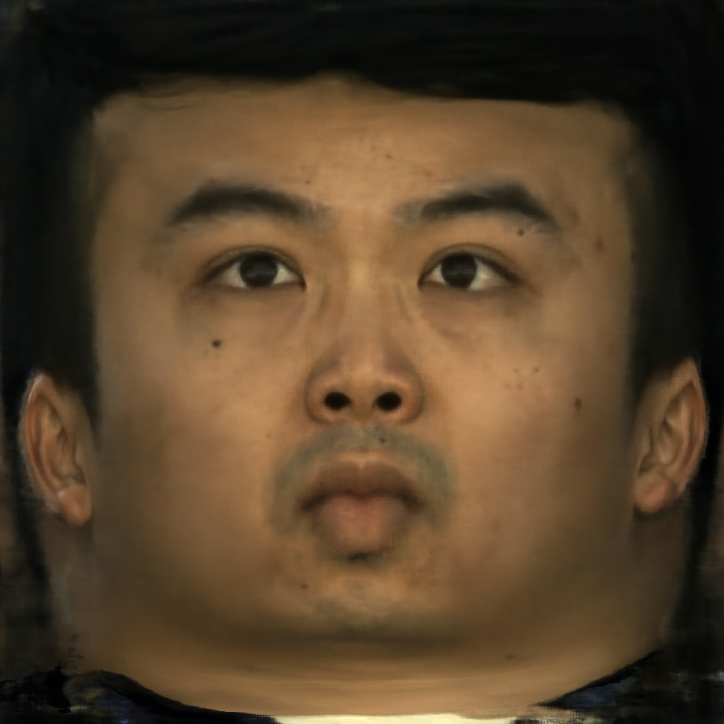}
    \caption{$\mathcal{L}_{uv}$ w/ decay}
    \label{subfig:ablation_uv_ours}
    \end{subfigure}
    \begin{subfigure}{0.325\linewidth}
    \includegraphics[width=\linewidth]{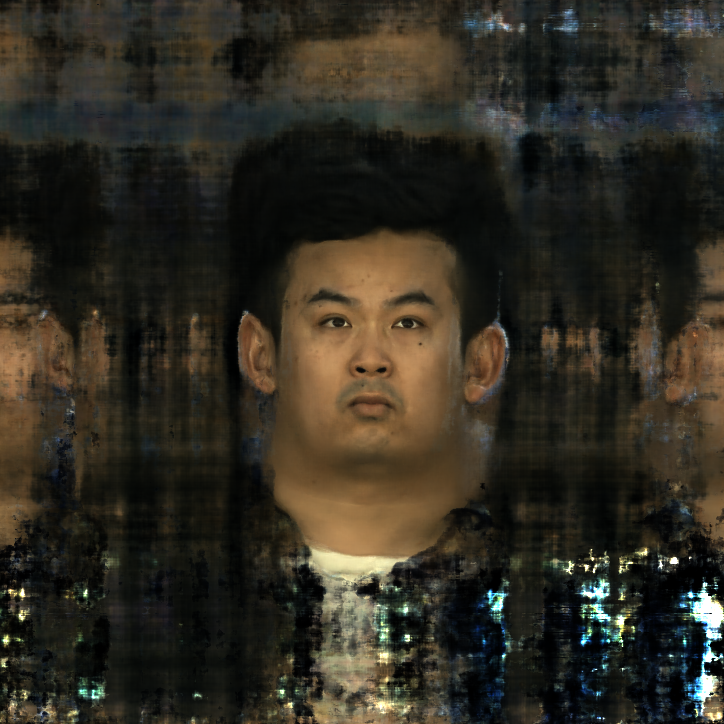}
    \caption{w/o $\mathcal{L}_{uv}$}
    \label{subfig:ablation_uv_nouv}
    \end{subfigure}
    \begin{subfigure}{0.325\linewidth}
    \includegraphics[width=\linewidth]{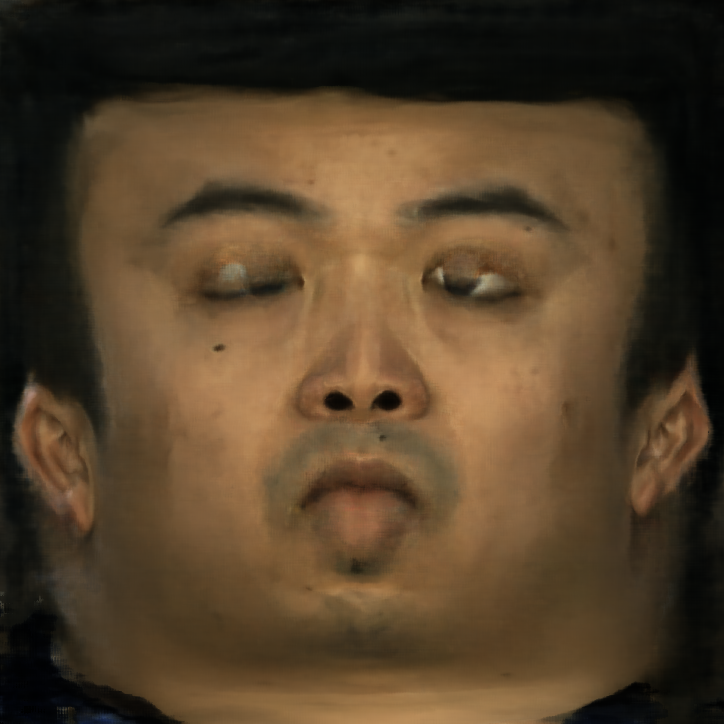}
    \caption{$\mathcal{L}_{uv}$ w/o decay}
    \label{subfig:ablation_uv_biguv}
    \end{subfigure}
    \vspace{-2mm}
    \caption{An example of different configurations of $\mathcal{L}_{uv}$. }
    \label{fig:ablation_uv}
\end{figure}

First, we would like to provide a coarse guidance of the UV field using the tracked 3D face mesh and its UV mapping, so that the model could have a big picture of where each component of the face maps to. Recall the tracked 3D face mesh has the form of a set of vertices $\{\mathbf{p}_i\}$ and its corresponding UV coordinate $\{\mathbf{u}_i\}$. We apply a $uv$ loss to the $\mathbf{u}$ output of UV field $V$:
\begin{equation}
    \mathcal{L}_{uv} = \sum_i^P \|V(\mathbf{p}_i, t) - \mathbf{u}_i\|_2,
\end{equation}
where $P$ is the total number of mesh vertices. Since the tracked mesh may be inaccurate, we only use it as initialization by setting the weight to exponentially decay to 0 at approximately 20,000 iterations. Fig. \ref{fig:ablation_uv} demonstrates an example of a texture with different configurations of the $\mathcal{L}_{uv}$. Without $\mathcal{L}_{uv}$, the texture mapping is freely optimized, which fails to utilize the full texture space as shown in Fig.~\ref{subfig:ablation_uv_nouv}. Without the weight decay, the inaccurate face mesh tracking leads to an incorrect mapping, and the model tries to compensate for the mapping error by generating artifacts on the texture map as shown in Fig.~\ref{subfig:ablation_uv_biguv}. 

Next, we would like the UV field to reasonably parameterize the surface of the head. We extend the cycle loss proposed by \citet[]{NeRF_neutex} to dynamic scenes. Specifically, we jointly optimize an inverse mapping network $V_I^{-1}$ that maps the UV coordinate $\mathbf{u}$ back into a 3D location:
\begin{equation}
    \hat{\mathbf{x}}' = V_I^{-1} (\mathbf{u}, t).
\end{equation}
Note that the output of $V_I^{-1}$ is a 3D location in canonical space $\mathcal{B}'$ instead of the observation space $\mathcal{B}$. This is because the canonical space contains fewer temporal variations, which eases the complexity of the inverse mapping network. The cycle loss is then defined as:
\begin{equation}
    \mathcal{L}_{cycle} = \sum_i^{B}
    \|\mathbf{x}_i' - \hat{\mathbf{x}}_i'\|_2.
\end{equation}
To reduce the computational overhead, rather than applying the cycle loss to all the sampling points, we only apply it to the points $\mathbf{x}_i'$ that have the maximum contribution to the rendering results along each ray. Specifically, when rendering a ray in the volume, the final color is formulated as a weighted sum of all sampled colors. We then determine the maximum contribution point by selecting the point that has the maximum weight during rendering.

The $\mathcal{L}_{cycle}$ successfully constrains the surface of the head to be mapped to a 2D texture. However, in practice we find that the smoothness of $(u, v)$ along the surface direction is not guaranteed using only the cycle loss. We would like the UV field to have the ability to model high frequency discontinuities, such as closing of the eyelids and lip, while also being piece-wise smooth in regions like the cheek and jaws for better editability. This discontinuity can be achieved by using the positional encoding illustrated in \cite{fourier_embed}. To ensure piece-wise smoothness, we employ an angle preserving loss:
\begin{equation}
    \mathcal{L}_{angle} = \sum_i^{B} \frac{|\nabla_\mathbf{x}u_\perp \cdot \nabla_\mathbf{x}v_\perp| }{\|\nabla_\mathbf{x}u_\perp\| \|\nabla_\mathbf{x}v_\perp\|},
    \label{eq:l_angle}
\end{equation}
% check here
where $\nabla_\mathbf{x}u_\perp$ and $\nabla_\mathbf{x}v_\perp$ are the gradient of the texture coordinate $(u, v)$, i.e. the output of the UV field $V$, with respect to the input position $\mathbf{x}$, after being projected to the plane perpendicular to the surface normal at $\mathbf{x}$. The surface normal is defined as the gradient of the density $\nabla_\mathbf{x}\sigma$. This loss regularizes the UV mapping along the surface to be as conformal as possible. This term is also applied to the sampling points with maximum contribution along a ray. We will show in the experiment that this loss leads to the smoothness of the UV field. 

% \begin{table*}[t]
% \begin{tabular}{l|l|lll|lll|l}
%           &          & \multicolumn{3}{c|}{Face region}           & \multicolumn{3}{c|}{Full head}             \\
%           & Editable & 
%         %   MSE$\downarrow$  & 
%           PSNR$\uparrow$     & SSIM$\uparrow$  & LPIPS$\downarrow$  & 
%         %   MSE$\downarrow$  & 
%           PSNR$\uparrow$ & SSIM$\uparrow$ & LPIPS$\downarrow$ & \textit{ASTD}$\downarrow$   \\
% \hline
% HyperNeRF & No   & 
% % 53.84 & 
% 31.31 & 0.8078   & 0.04270 & 
% % 112.8  &
% 29.11 & 0.8092 & 0.0979 & -\\
% NeRF+T    & No        & 
% % \textbf{37.27} &
% \textbf{32.61} & \textbf{0.8381}  & 0.03960 & 
% % \textbf{102.5} &
% \textbf{29.53} & \textbf{0.8313} & 0.09273 & -\\
% Ours      & Yes        & 
% % 60.27 & 
% 30.62 & 0.7998 & 0.03382 & 
% % 127.7 &
% 28.38 & 0.7964 & \textbf{0.08892} & \textbf{2.004} \\
% DFNRMVS   & Yes        & 
% % 92.67 & 
% 28.75  & 0.7778 & \textbf{0.02633} & 
% % -        &
% -        & -        & -   & 4.984 \\
% HiFi3D    & Yes        & 
% % 413.8 &
% 22.46 & 0.5601 & 0.07053 & 
% % -        &
% -        & -        & -     & 5.616
% \end{tabular}
% \caption{Reconstruction and texture temporal alignment. Ours is slightly worse than the NeRF-based methods due to the regularization for improving the editability. Invalid cells are marked as $-$. $\uparrow$ means higher is better. Ours achieved the best trade-off between reconstruction and editability.}
% \label{tab:comparison_reconstruct}
% \end{table*}

\begin{table*}[t]
\caption{Reconstruction and texture temporal alignment. Our approach is slightly inferior to the NeRF-based methods due to the regularization for improving editability. Overall, our method achieves the best trade-off between reconstruction and editability. (Invalid cells are denoted as $-$. $\uparrow$ means higher values are better.)}
\label{tab:comparison_reconstruct}
\vspace{-6mm}
\begin{tabular}{l|l|llll|lll}
          &          & \multicolumn{4}{c|}{Face region}           & \multicolumn{3}{c}{Full head}   \\
          & Editable & 
        %   MSE$\downarrow$  & 
          PSNR$\uparrow$     & SSIM$\uparrow$  & LPIPS$\downarrow$  & \textit{ASTD}$\downarrow$ & 
        %   MSE$\downarrow$  & 
          PSNR$\uparrow$ & SSIM$\uparrow$ & LPIPS$\downarrow$   \\
\hline
HyperNeRF & No   & 
% 53.84 & 
31.31 & 0.8078   & 0.04270 & - & 
% 112.8  &
29.11 & 0.8092 & 0.0979 \\
DyNeRF   & No        & 
% \textbf{37.27} &
\textbf{32.61} & \textbf{0.8381}  & 0.03960 & -& 
% \textbf{102.5} &
\textbf{29.53} & \textbf{0.8313} & 0.09273 \\
Ours      & Yes        & 
% 60.27 & 
30.62 & 0.7998 & 0.03382 & \textbf{2.713} & 
% 127.7 &
28.38 & 0.7964 & \textbf{0.08892}  \\
DFNRMVS   & Yes        & 
% 92.67 & 
28.75  & 0.7778 & \textbf{0.02633} & 4.984 & 
% -        &
-        & -        & -   \\
HiFi3D    & Yes        & 
% 413.8 &
22.46 & 0.5601 & 0.07053 & 5.616 & 
% -        &
-        & -        & -  \\
PRNet   & Yes        & 
% 92.67 & 
27.48  & 0.7745 & 0.05148 & 11.60 & 
% -        &
-        & -        & -   \\
\end{tabular}
\end{table*}

\subsection{Two-stage Training}
The time variation of the scene can be modeled either by the dynamic UV field $V$ or the dynamic texture $T$. In order to achieve consistent appearance editing, we would like the dynamic textures to be as static as possible, while leaving the majority of the time variation to be fitted using the UV field. In an ideal representation, the points of the same feature on the face for different frames should be mapped to the same texture coordinate.
% the dynamic texture only models dynamics that UV field can not modeled, such as the self occlusion that is caused by wrinkles of the skin. In other world, the textures of different time stamps should be pixel-wisely aligned. Then the editing of the textures does not introduce misalignment. 
To improve the alignment of the dynamic texture, we propose a two-stage training procedure. In the first stage, we fix the texture $T$ to be time independent. This is implemented by setting the embedding of the time index to be the same for all frames. Thus, the dynamic UV field tries its best to model the dynamic human head at this stage. After the training converges (at 80,000 iterations in all of our experiments), we start optimizing a dynamic texture to model the rest of the time variations that a dynamic UV field fails to model. 

The final loss is the combination of all the losses introduced above:
\begin{align}
\begin{split}
    \mathcal{L} = \mathcal{L}_{MSE} &+ \lambda_{uv} \mathcal{L}_{uv} + \lambda_{cycle} \mathcal{L}_{cycle} + \lambda_{angle} \mathcal{L}_{angle} \\
    &+ \lambda_{sparsity} \mathcal{L}_{sparsity} + \lambda_{semantic} \mathcal{L}_{semantic}.
\end{split}
\end{align}
where $\lambda_{(\cdot)}$ is the corresponding weight for each loss. We set $\lambda_{cycle} = 1$ as suggested in \cite{NeRF_neutex}. We set the $\lambda_{uv}$ to $1$ and quickly decrease to 0 at 20,000 iterations as described above. In all the experiments we set $\lambda_{angle} = \lambda_{sparsity} = \lambda_{semantic} = 0.05$. We ablate the weighting scheme in the experiment section.

\begin{figure*}[t]
\newcommand\ws{0.138}
\newcommand\hs{-1.6mm}
\newcommand\vs{-1mm}
\centering

%%%%%%%%%%%%%%%%%%%%%%%%%%%%%%
\hfill
\begin{subfigure}{\ws\linewidth}
\includegraphics[width=\linewidth]{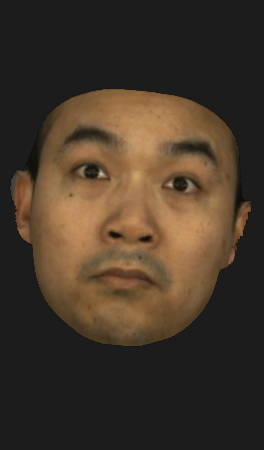}
\end{subfigure}
\hspace{\hs}
\begin{subfigure}{\ws\linewidth}
\includegraphics[width=\linewidth]{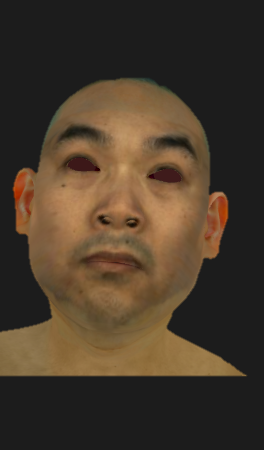}
\end{subfigure}
\hspace{\hs}
\begin{subfigure}{\ws\linewidth}
\includegraphics[width=\linewidth]{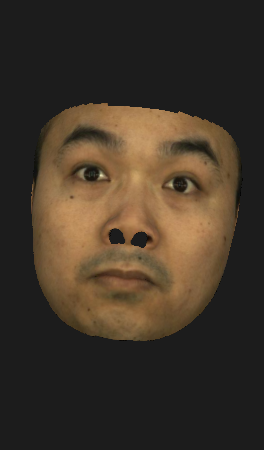}
\end{subfigure}
\hspace{\hs}
\begin{subfigure}{\ws\linewidth}
\includegraphics[width=\linewidth]{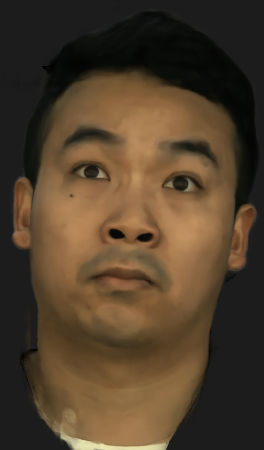}
\end{subfigure}
\hspace{\hs}
\begin{subfigure}{\ws\linewidth}
\includegraphics[width=\linewidth]{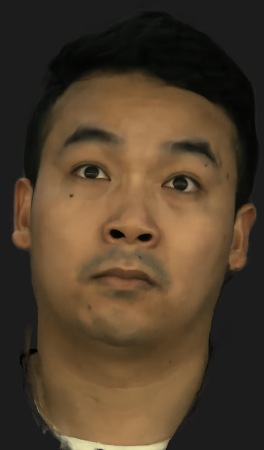}
\end{subfigure}
\hspace{\hs}
\begin{subfigure}{\ws\linewidth}
\includegraphics[width=\linewidth]{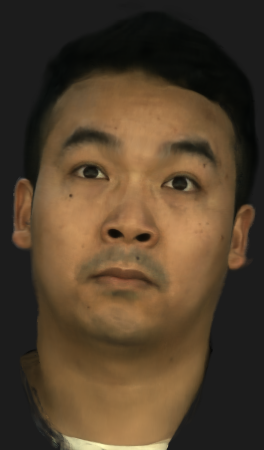}
\end{subfigure}
\hspace{\hs}
\begin{subfigure}{\ws\linewidth}
\includegraphics[width=\linewidth]{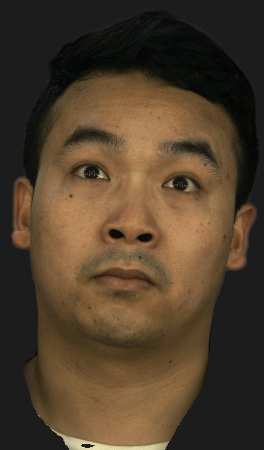}
\end{subfigure}
\vspace{-0.7cm}
\hfill

%%%%%%%%%%%%%%%%%%%%%%%%%%%%%
\hfill
\begin{subfigure}{\ws\linewidth}
\includegraphics[width=\linewidth]{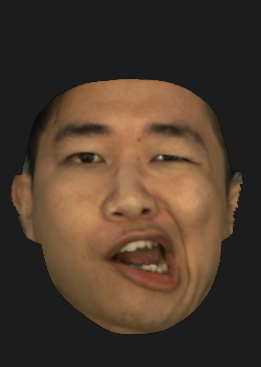}
\end{subfigure}
\hspace{\hs}
\begin{subfigure}{\ws\linewidth}
\includegraphics[width=\linewidth]{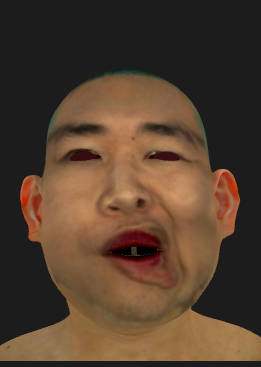}
\end{subfigure}
\hspace{\hs}
\begin{subfigure}{\ws\linewidth}
\includegraphics[width=\linewidth]{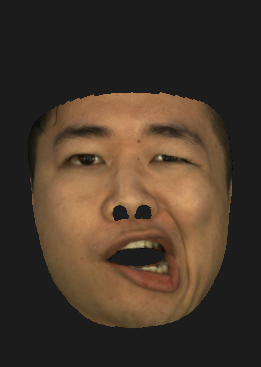}
\end{subfigure}
\hspace{\hs}
\begin{subfigure}{\ws\linewidth}
\includegraphics[width=\linewidth]{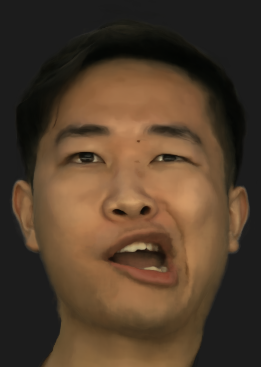}
\end{subfigure}
\hspace{\hs}
\begin{subfigure}{\ws\linewidth}
\includegraphics[width=\linewidth]{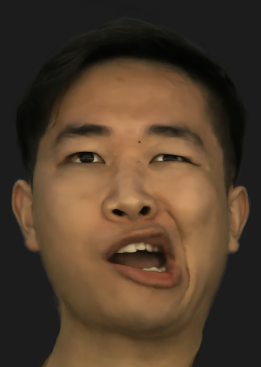}
\end{subfigure}
\hspace{\hs}
\begin{subfigure}{\ws\linewidth}
\includegraphics[width=\linewidth]{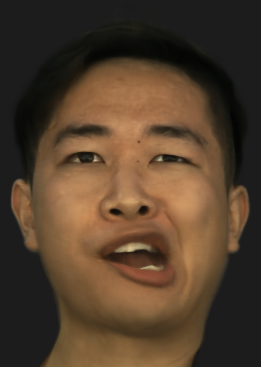}
\end{subfigure}
\hspace{\hs}
\begin{subfigure}{\ws\linewidth}
\includegraphics[width=\linewidth]{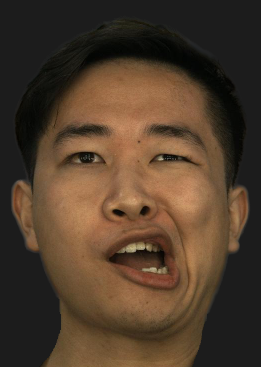}
\end{subfigure}
\vspace{\vs}
\hfill

%%%%%%%%%%%%%%%%%%%%%%%%%%%%%%
\hfill
\begin{subfigure}{\ws\linewidth}
\includegraphics[width=\linewidth]{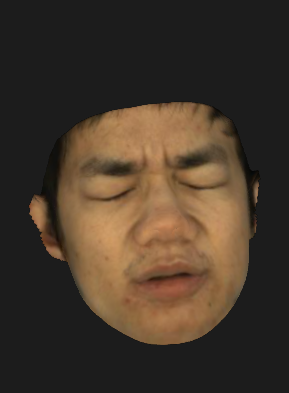}
\caption{PRNet}
\end{subfigure}
\hspace{\hs}
\begin{subfigure}{\ws\linewidth}
\includegraphics[width=\linewidth]{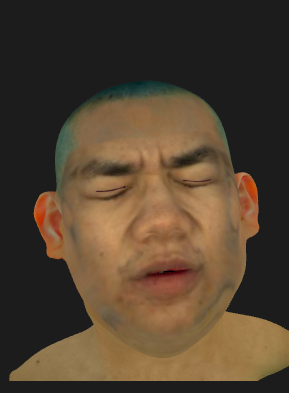}
\caption{HiFi3D}
\end{subfigure}
\hspace{\hs}
\begin{subfigure}{\ws\linewidth}
\includegraphics[width=\linewidth]{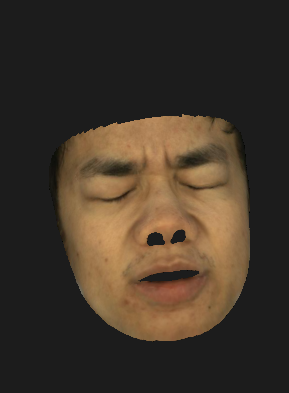}
\caption{DFNRMVS}
\end{subfigure}
\hspace{\hs}
\begin{subfigure}{\ws\linewidth}
\includegraphics[width=\linewidth]{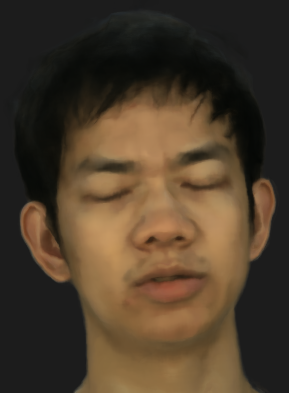}
\caption{HyperNeRF}
\end{subfigure}
\hspace{\hs}
\begin{subfigure}{\ws\linewidth}
\includegraphics[width=\linewidth]{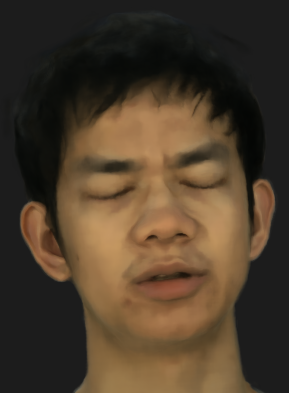}
\caption{DyNeRF}
\end{subfigure}
\hspace{\hs}
\begin{subfigure}{\ws\linewidth}
\includegraphics[width=\linewidth]{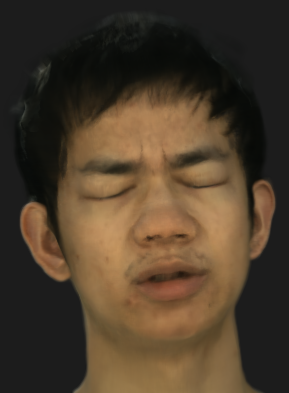}
\caption{Ours}
\end{subfigure}
\hspace{\hs}
\begin{subfigure}{\ws\linewidth}
\includegraphics[width=\linewidth]{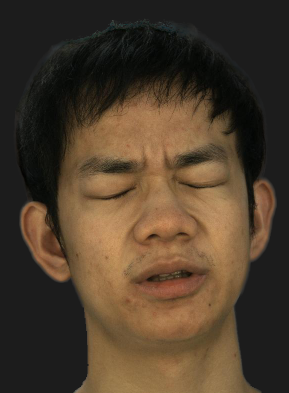}
\caption{Ground Truth}
\end{subfigure}
\vspace{\vs}
\hfill

\caption{Novel view synthesis results. Each row shows the results of one subject. Ours enables editing of the full head while achieving similar rendering results to NeRF-based methods. Some regions of ours, such as the teeth are less detailed, while other regions often outperform NeRF-based methods due to the use of the explicit deformation field. }
\label{fig:recon_fig}
\end{figure*}

\section{Experiments}
\label{sec:experiments}
In this section, we perform quantitative and qualitative experiments to validate our model on the captured multi-view video dataset, which contains 10 video sequences,  120 - 170 frames each. We implement both $V_I$ and $T_I$ using MLPs that have 6 layers and 256 neurons per layer, with a skip connect at the 4th layer. We first describe several metrics that we use in the experiments and then compare reconstruction and appearance editing quality with several baselines. We also conduct a series of ablations to validate the effectiveness of our model. 

\subsection{Metrics}
For the reconstruction quality, we use the task of novel view synthesis, and compare the synthesized novel view and ground truth. To measure editability, we develop several metrics to compute how well the model could be used for editing. First, since we want to consistently edit the entire sequence by editing on one texture, we measure the quality of the alignment by computing the average standard deviation (\textit{ASTD}) of each pixel color in the face region over the entire sequence of a dynamic texture. A bigger \textit{ASTD} indicates bigger time variations of the dynamic texture, which leads to poor editability. In the experiment, we find that conformal UV mapping greatly affects the editing results. We measure the average angle preserving error (\textit{Angle}) similar to how we compute $\mathcal{L}_{angle}$, but averaged over all the pixels in the test view. 

\begin{figure*}
\newcommand\ws{0.15}
\newcommand\hs{-1.6mm}
\newcommand\vs{-0.1mm}
\centering

\hfill
\rotatebox[origin=c]{90}{\makebox{\centering DFNRMVS}}
\begin{subfigure}{\ws\linewidth}
\includegraphics[width=\linewidth]{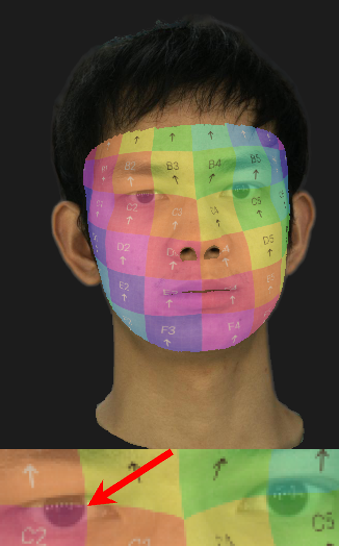}
\end{subfigure}
\hspace{\hs}
\begin{subfigure}{\ws\linewidth}
\includegraphics[width=\linewidth]{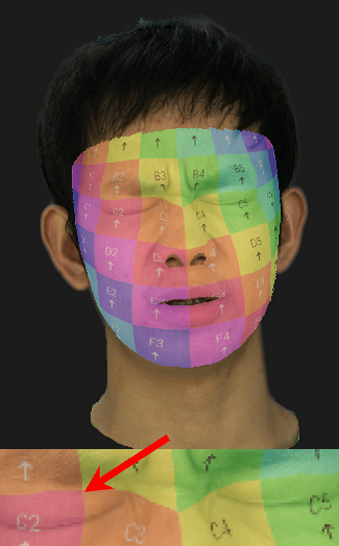}
\end{subfigure}
\hspace{\hs}
\begin{subfigure}{\ws\linewidth}
\includegraphics[width=\linewidth]{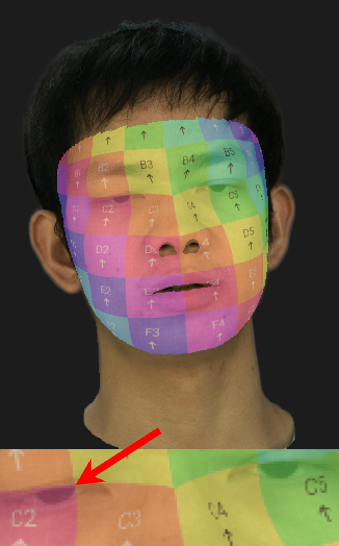}
\end{subfigure}
\vspace{\vs}
\hfill
\hfill
\rotatebox[origin=c]{90}{\makebox{\centering HiFi3D}}
\begin{subfigure}{\ws\linewidth}
\includegraphics[width=\linewidth]{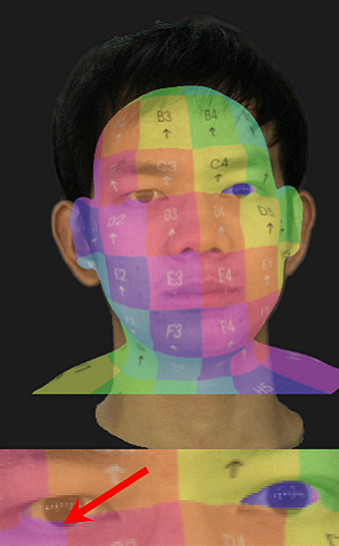}
\end{subfigure}
\hspace{\hs}
\begin{subfigure}{\ws\linewidth}
\includegraphics[width=\linewidth]{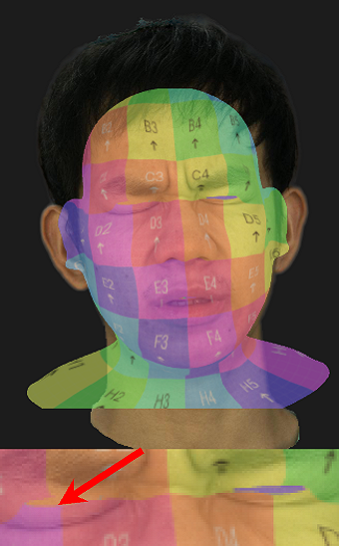}
\end{subfigure}
\hspace{\hs}
\begin{subfigure}{\ws\linewidth}
\includegraphics[width=\linewidth]{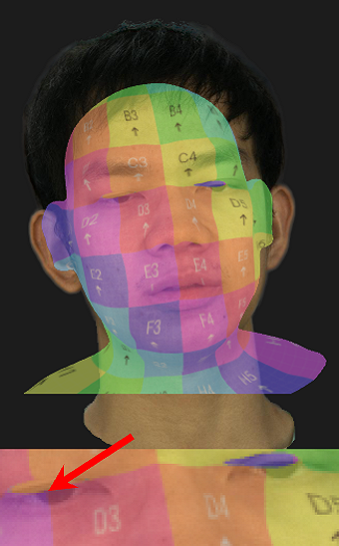}
\end{subfigure}
\hfill

\vspace{0.5mm}

\hfill
\rotatebox[origin=c]{90}{\makebox{\centering PRNet}}
\begin{subfigure}{\ws\linewidth}
\includegraphics[width=\linewidth]{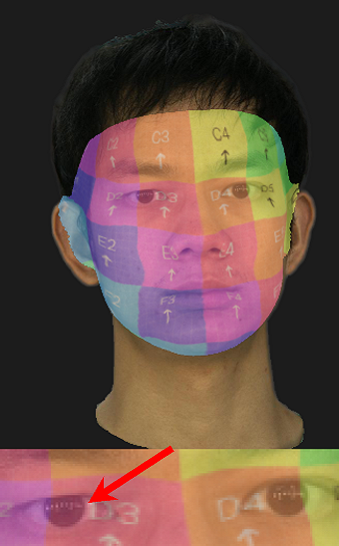}
\caption*{Frame 1}
\end{subfigure}
\hspace{\hs}
\begin{subfigure}{\ws\linewidth}
\includegraphics[width=\linewidth]{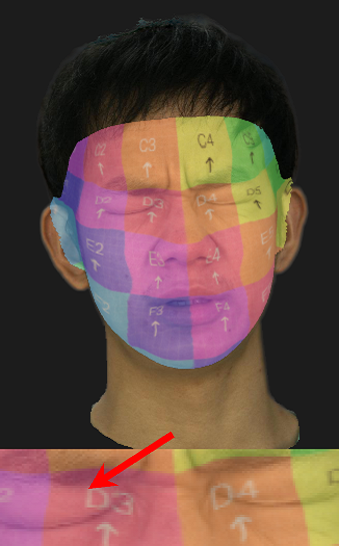}
\caption*{Frame 26}
\end{subfigure}
\hspace{\hs}
\begin{subfigure}{\ws\linewidth}
\includegraphics[width=\linewidth]{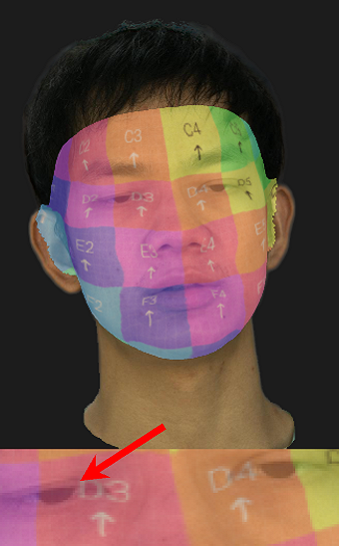}
\caption*{Frame 118}
\end{subfigure}
\vspace{\vs}
\hfill
\hfill
\rotatebox[origin=c]{90}{\makebox{\centering Ours}}
\begin{subfigure}{\ws\linewidth}
\includegraphics[width=\linewidth]{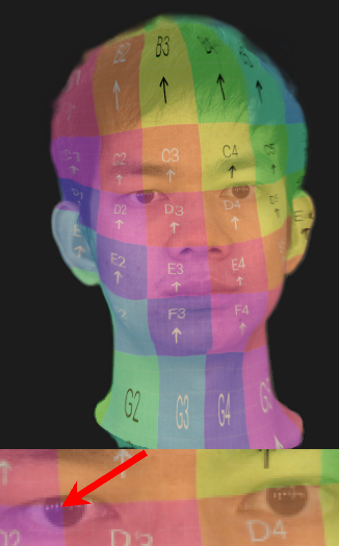}
\caption*{Frame 1}
\end{subfigure}
\hspace{\hs}
\begin{subfigure}{\ws\linewidth}
\includegraphics[width=\linewidth]{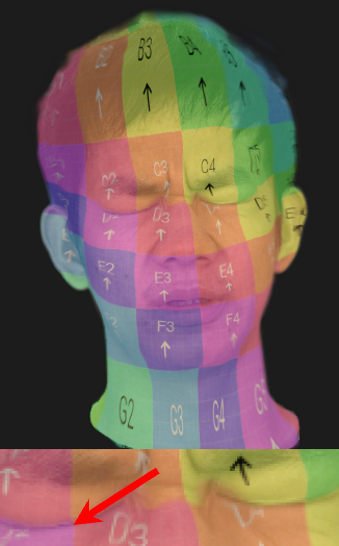}
\caption*{Frame 26}
\end{subfigure}
\hspace{\hs}
\begin{subfigure}{\ws\linewidth}
\includegraphics[width=\linewidth]{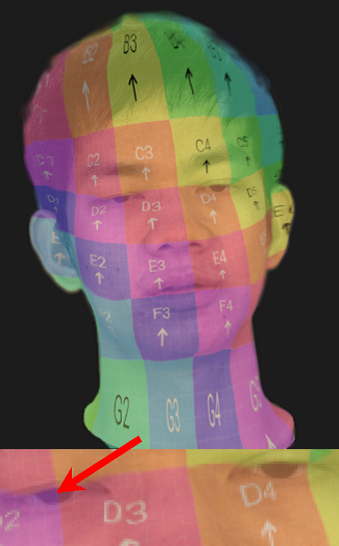}
\caption*{Frame 118}
\end{subfigure}
\hfill

\vspace{-2mm}
\caption{Visualizations of UV checker overlay. In each column, we show a frame of the same sequence. We highlight an equivalent point on the UV checker with a red arrow. Our UV mapping achieves better temporal consistency.}
\label{fig:edit_checker}
\end{figure*}

\begin{figure}
\newcommand\ws{0.24}
\newcommand\hs{-1.6mm}
\newcommand\vs{-1mm}
\centering
\rotatebox[origin=c]{90}{\makebox{\centering Frame 1}}
\begin{subfigure}{\ws\linewidth}
\includegraphics[width=\linewidth]{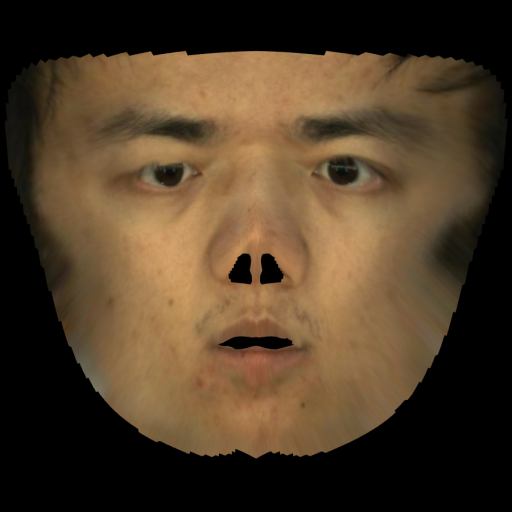}
\end{subfigure}
\hspace{\hs}
\begin{subfigure}{\ws\linewidth}
\includegraphics[width=\linewidth]{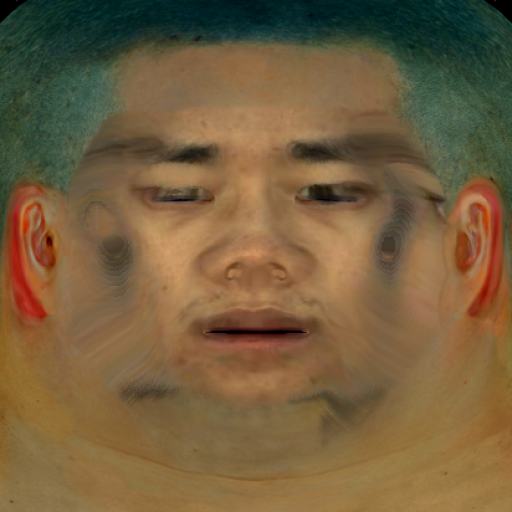}
\end{subfigure}
\hspace{\hs}
\begin{subfigure}{\ws\linewidth}
\includegraphics[width=\linewidth]{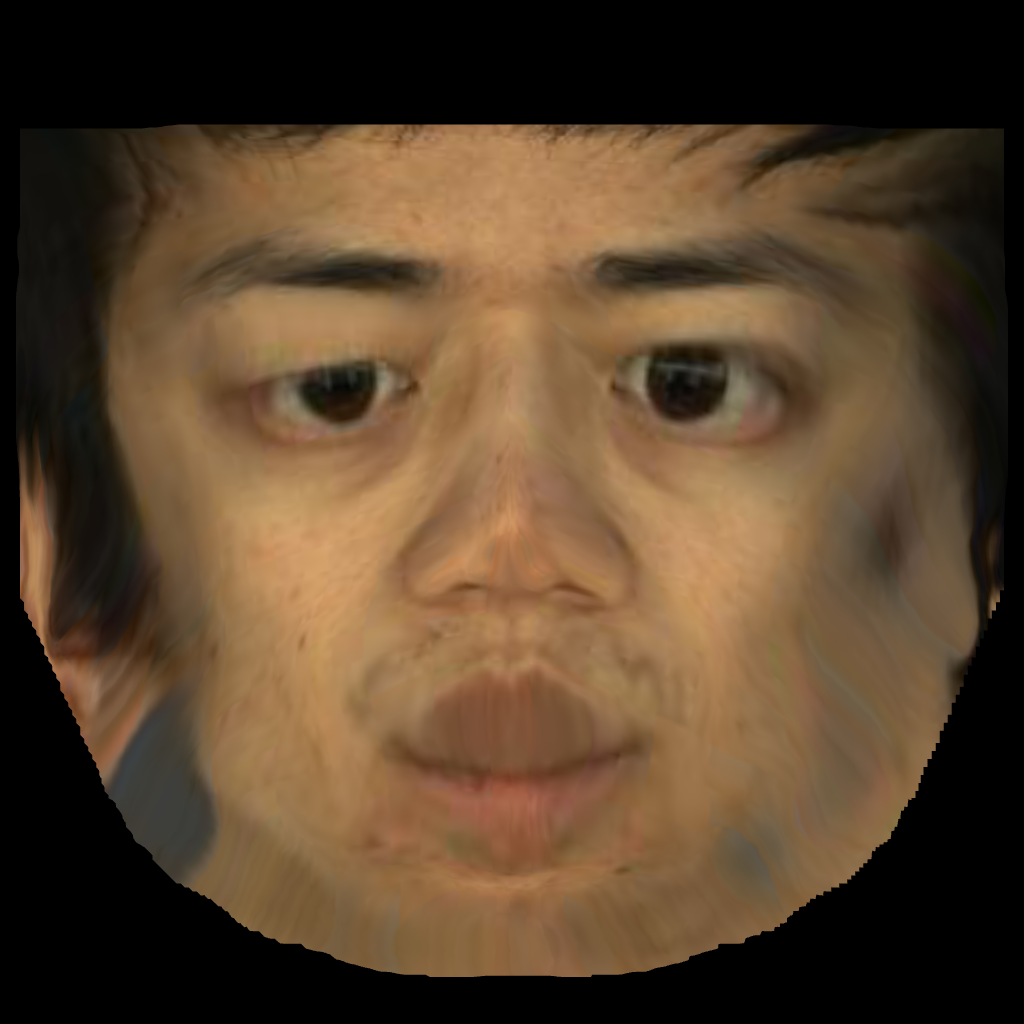}
\end{subfigure}
\hspace{\hs}
\begin{subfigure}{\ws\linewidth}
\includegraphics[width=\linewidth]{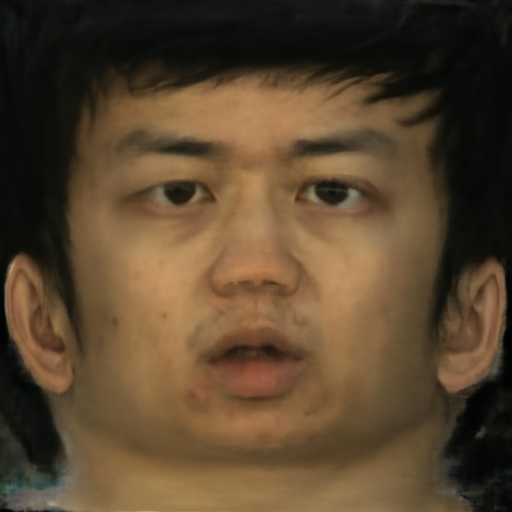}
\end{subfigure}
\vspace{\vs}

\rotatebox[origin=c]{90}{\makebox{\centering Frame 26}}
\begin{subfigure}{\ws\linewidth}
\includegraphics[width=\linewidth]{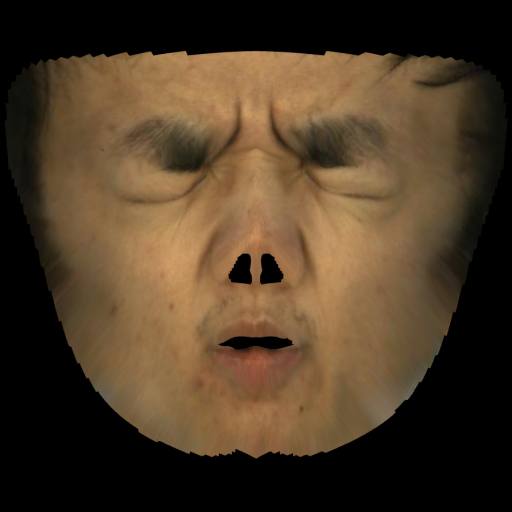}
\end{subfigure}
\hspace{\hs}
\begin{subfigure}{\ws\linewidth}
\includegraphics[width=\linewidth]{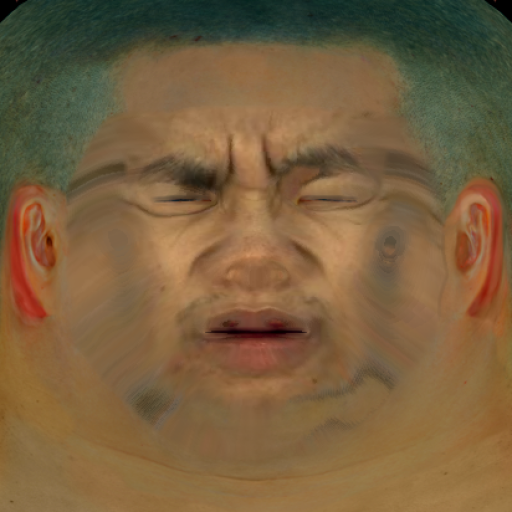}
\end{subfigure}
\hspace{\hs}
\begin{subfigure}{\ws\linewidth}
\includegraphics[width=\linewidth]{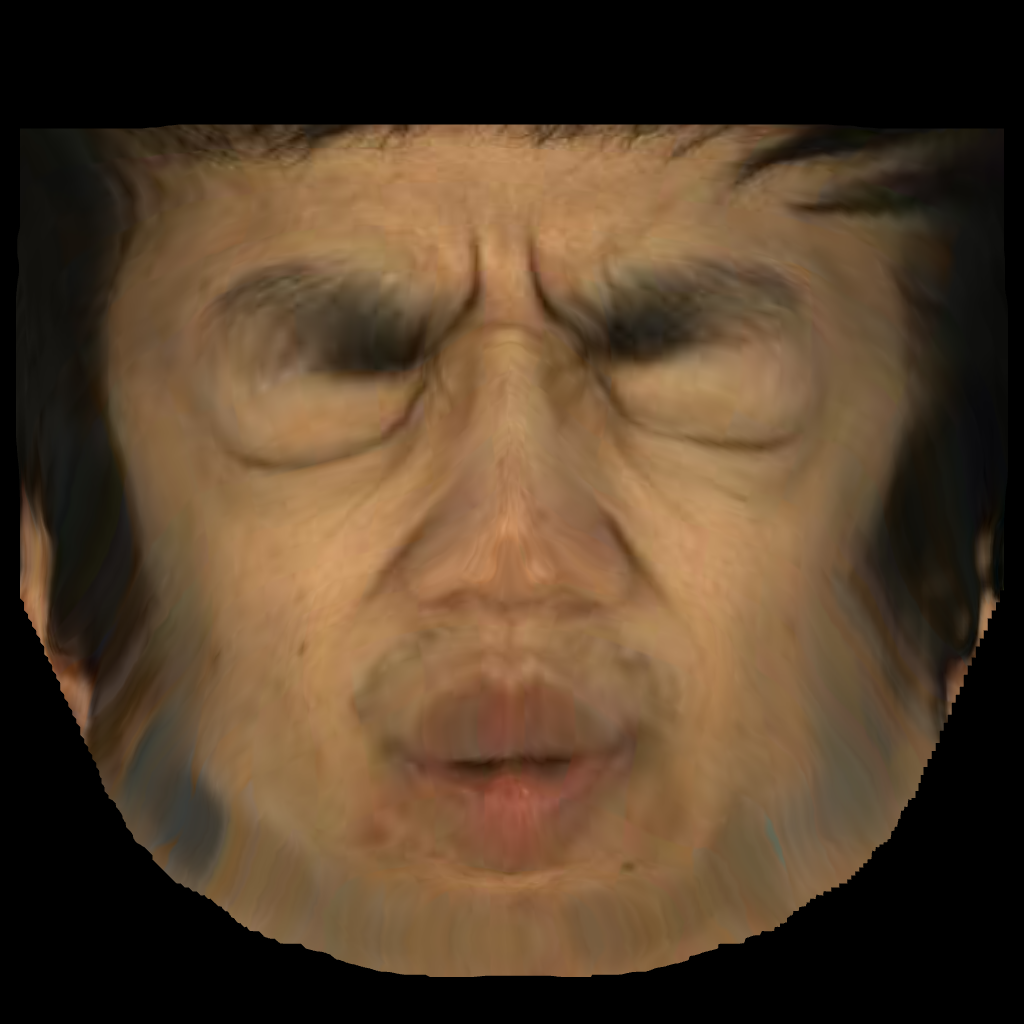}
\end{subfigure}
\hspace{\hs}
\begin{subfigure}{\ws\linewidth}
\includegraphics[width=\linewidth]{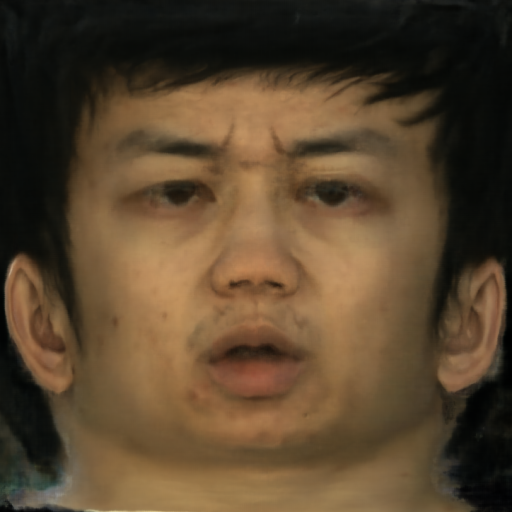}
\end{subfigure}
\vspace{\vs}

\rotatebox[origin=c]{90}{\makebox{\centering Frame 118}}
\begin{subfigure}{\ws\linewidth}
\includegraphics[width=\linewidth]{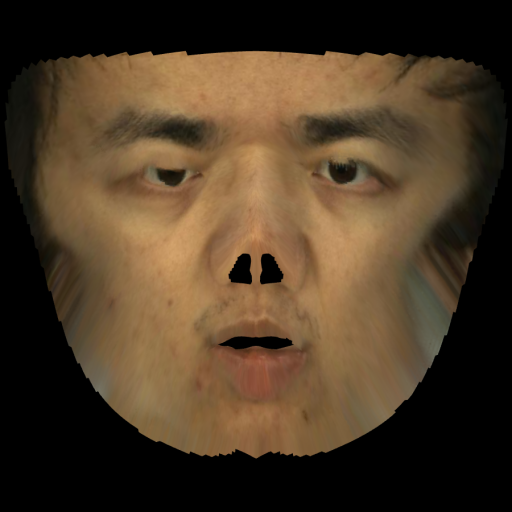}
\end{subfigure}
\hspace{\hs}
\begin{subfigure}{\ws\linewidth}
\includegraphics[width=\linewidth]{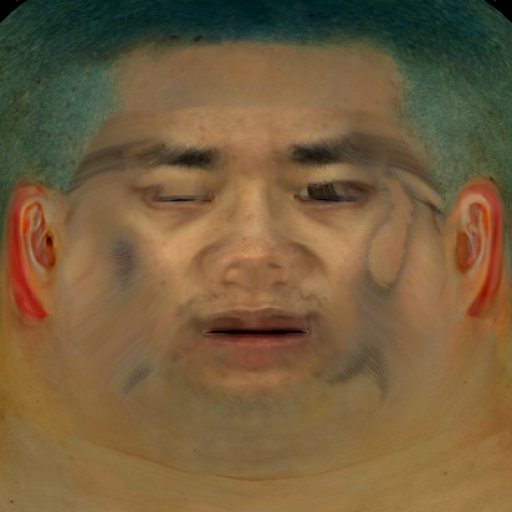}
\end{subfigure}
\hspace{\hs}
\begin{subfigure}{\ws\linewidth}
\includegraphics[width=\linewidth]{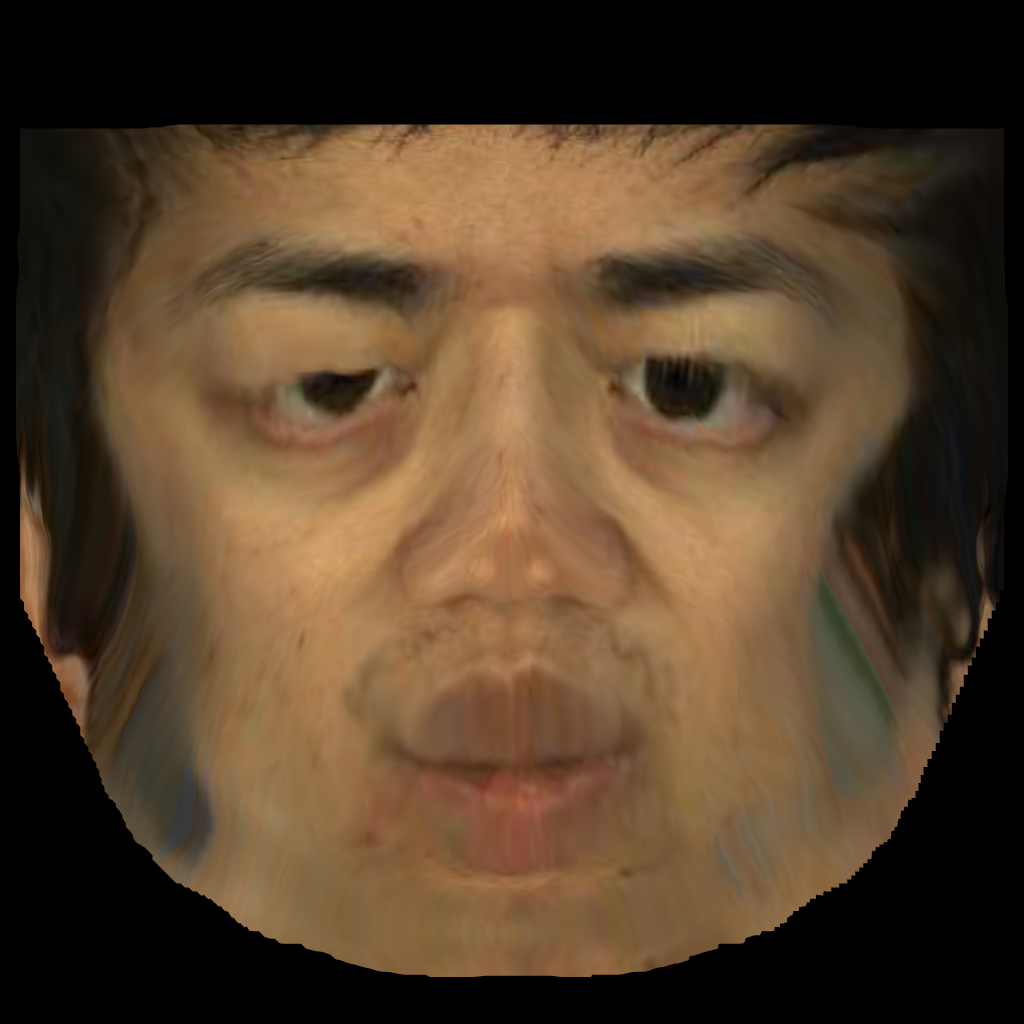}
\end{subfigure}
\hspace{\hs}
\begin{subfigure}{\ws\linewidth}
\includegraphics[width=\linewidth]{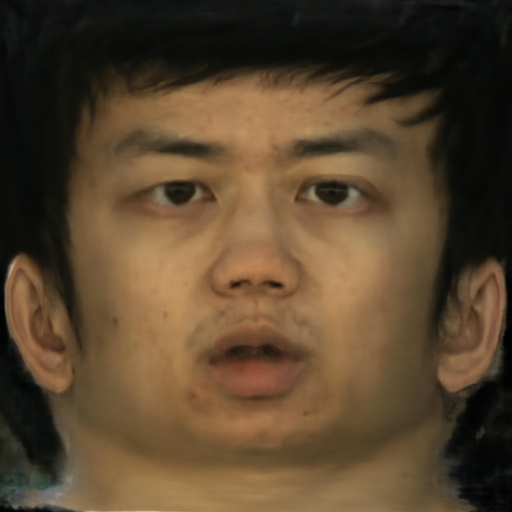}
\end{subfigure}
\vspace{\vs}

\rotatebox[origin=c]{90}{\makebox{\centering \textit{STD}}}
\begin{subfigure}{\ws\linewidth}
\includegraphics[width=\linewidth]{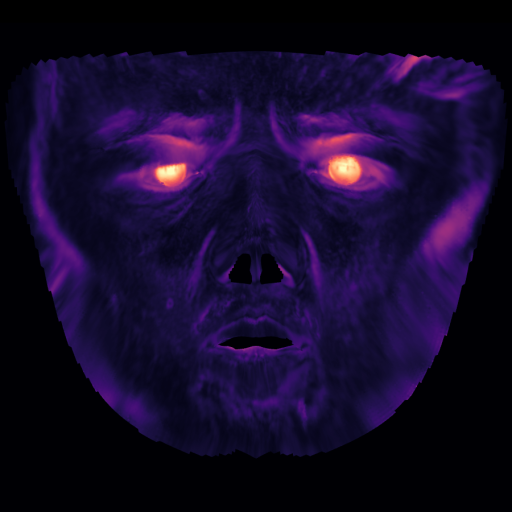}
\caption{DFNRMVS}
\end{subfigure}
\hspace{\hs}
\begin{subfigure}{\ws\linewidth}
\includegraphics[width=\linewidth]{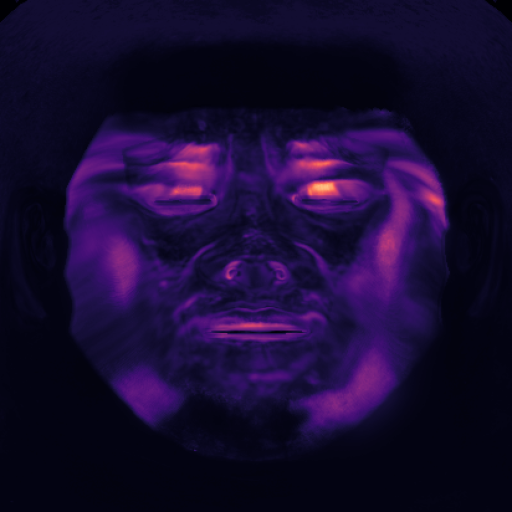}
\caption{HiFi3D}
\end{subfigure}
\hspace{\hs}
\begin{subfigure}{\ws\linewidth}
\includegraphics[width=\linewidth]{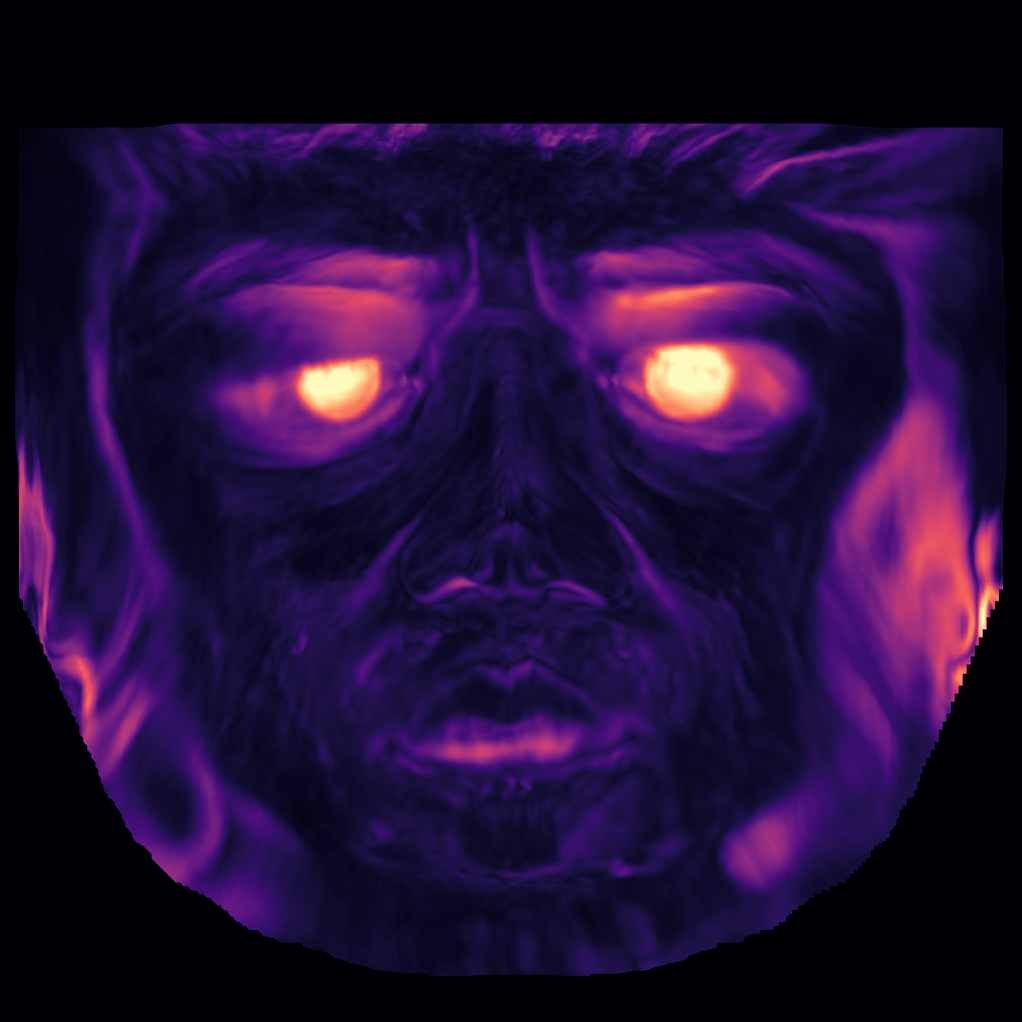}
\caption{PRNet}
\end{subfigure}
\hspace{\hs}
\begin{subfigure}{\ws\linewidth}
\includegraphics[width=\linewidth]{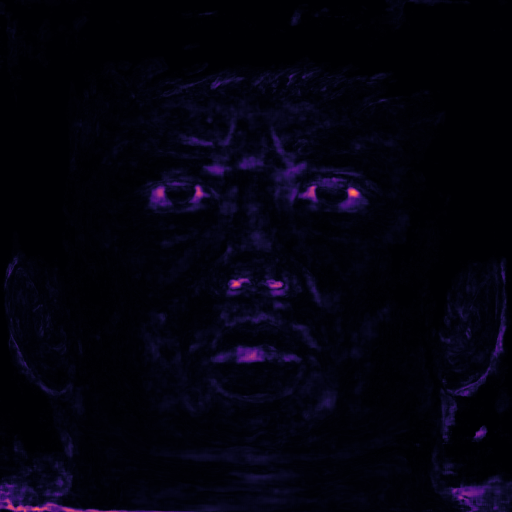}
\caption{Ours}
\end{subfigure}
\vspace{\vs}

\caption{Visualizations of textures in different frames. In the last row, we visualize the \textit{STD} maps of each pixel color over the entire sequence. It can be seen that ours achieves the best texture alignment.}
\label{fig:edit_texture}
\end{figure}

\subsection{Comparison}
\paragraph{Novel View Synthesis.}
We first evaluate the novel view synthesis quality, which reflects the reconstruction performance of the models. We compare with two dynamic NeRF baselines, namely \textit{HyperNeRF} \cite{NeRF_dyn_hypernerf} and \textit{DyNeRF} \cite{NeRF_dyn_3Dvideo}. We do not compare with NeRF-like human head reconstruction methods such as NerFace \cite{NeRFace_dynamic} and IM Avatar \cite{NeRFace_IMAvatar}, since these methods take monocular input and exploit the head pose changes to collect multi-view information. We find that extending these methods for multi-view input is a non-trivial problem.
We then compare with mesh-based methods, \textit{HiFi3D} \cite{3dmm_dr_hifi3d},  \textit{DFNRMVS} \cite{mesh_face_dfnrmvs}, as well as the PRNet \cite{3dmm_nn_prnet} that our method uses as the initialization. We additionally provide results compared on the Facescape \cite{facescape} and Beeler \cite{Beeler11} datasets in the supplementary material.
% \remove{We also compare with mesh-based methods, \textit{HiFi3D} \cite{3dmm_dr_hifi3d} and \textit{DFNRMVS} \cite{mesh_face_dfnrmvs}.} 
For the \textit{DFNRMVS} method, which reconstructs facial meshes from multi-views, we find that blending multiple textures from multi-views results in blurry textures and worsens the performance. Therefore, we compute the mesh texture from a nearest neighbor view, which has roughly the same self-occlusion as the test view. For both mesh-based methods, we apply the algorithm frame by frame. We could have compared our method with other multi-view face reconstruction methods \cite{mesh_appear_DAM,mesh_mvstopology,NeRFace_deform}, but the source codes are not publicly available before the submission of this paper. We also try to compare with Tofu \cite{3dmm_nn_tofu}, but we find the reconstruction quality is poor, possibly due to the different camera, lighting, and subject settings. We report novel view synthesis results in Tab. \ref{tab:comparison_reconstruct}. We separately report reconstruction metrics in the face region and full head. Results show that our method is slightly worse than the NeRF-based method regarding the reconstruction quality. This is expected since HyperNeRF and DyNeRF focus on reconstruction only and do not support editing, while we apply strong regularizations to our volume to guarantee the editing quality. Our method outperforms the mesh-based methods (except for the LPIPS value in the face region) either on the reconstruction quality or the texture temporal alignment (\textit{ASTD}). The results also show that the initial fitting (PRNet) is inaccurate and that our method is robust to the inaccurate initialization and could further improve the accuracy based on the initialization. We also show several qualitative results in Fig. \ref{fig:recon_fig}. It can be seen that compared to mesh-based methods, our method naturally reconstructs the hair and the inside of the mouth. Our method lost some of the details in the teeth compared with the NeRF-based method. We also find that in cases where the head moves, our method predicts sharper textures. This is probably because of the use of the explicit deformation field $V_E$, which aligns the head.
% \change{We also compare our method on other datasets More comparisons with pre-registered mesh can be found in the supplementary material.}

\vspace{-2mm}
\paragraph{Editing.}
We then compare the quality of appearance editing with mesh-based methods. Although mesh-based methods focus mainly on the reconstruction, we could achieve appearance editing easily using tracked mesh by overlaying the editing on top of the input. Here we focus on the temporal consistency of the UV mapping, which reflects the temporal consistency of the editing. We first show the UV checkered pattern (UV checker) overlays on Fig.~\ref{fig:edit_checker}. For video results, please see the supplementary material. Note the points highlighted in the red arrow, which is the same point in the UV checker. Although mesh-based methods achieve plausible UV mapping in the face region for a single frame, temporal consistency is not guaranteed. The same point on the texture can be mapped to different locations of the face (inside the eye and on the eyelid) in the same video sequence. Besides, ours achieves mapping for the entire head, while HiFi3D fails to generate a plausible mapping in the hair, ear, and neck. We show the textures of different frames in Fig.~\ref{fig:edit_texture}. We also show the texture \textit{STD} maps alongside the textures. Ours demonstrates smaller temporal variations of the dynamic textures, which also indicates that ours has a more consistent mapping compared to mesh-based methods.

% How to edit

\begin{figure}
\newcommand\ws{0.25}
% * 308 / 488
\newcommand\wsa{0.15778689}
\newcommand\hs{-1.6mm}
\newcommand\vs{-1mm}
\centering

\hfill
\rotatebox[origin=c]{90}{\makebox{\centering $0$}}
\begin{subfigure}{\ws\linewidth}
\includegraphics[width=\linewidth]{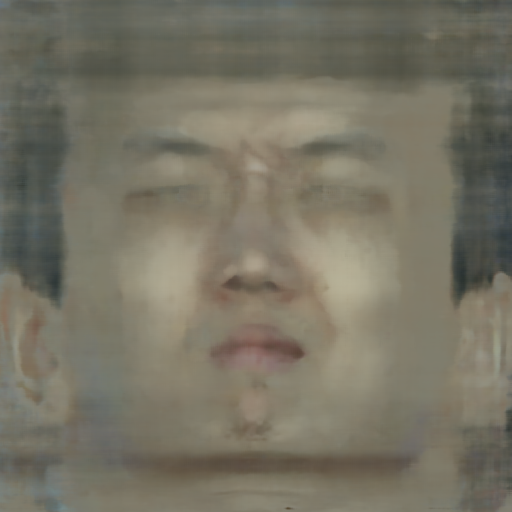}
\end{subfigure}
\hspace{\hs}
\begin{subfigure}{\ws\linewidth}
\includegraphics[width=\linewidth]{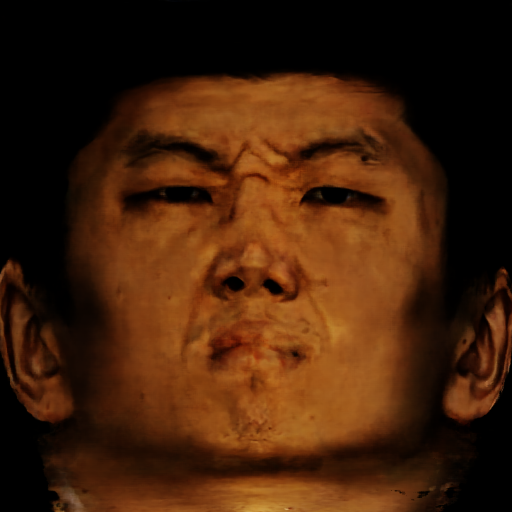}
\end{subfigure}
\hspace{\hs}
\begin{subfigure}{\ws\linewidth}
\includegraphics[width=\linewidth]{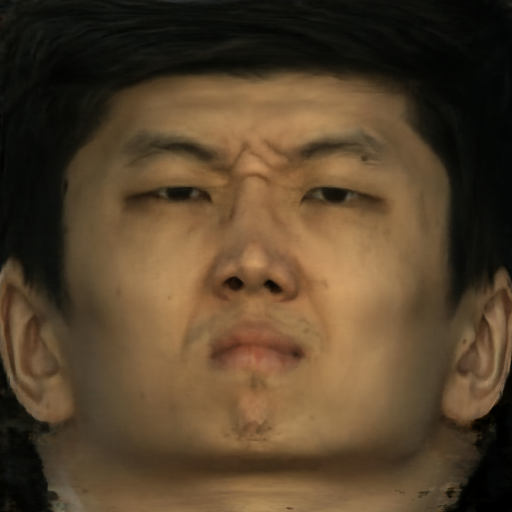}
\end{subfigure}
\hspace{\hs}
\begin{subfigure}{\wsa\linewidth}
\includegraphics[width=\linewidth]{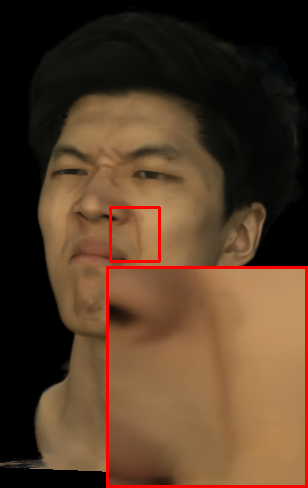}
\end{subfigure}
\vspace{\vs}
\hfill

\hfill
\rotatebox[origin=c]{90}{\makebox{\centering $0.005$}}
\begin{subfigure}{\ws\linewidth}
\includegraphics[width=\linewidth]{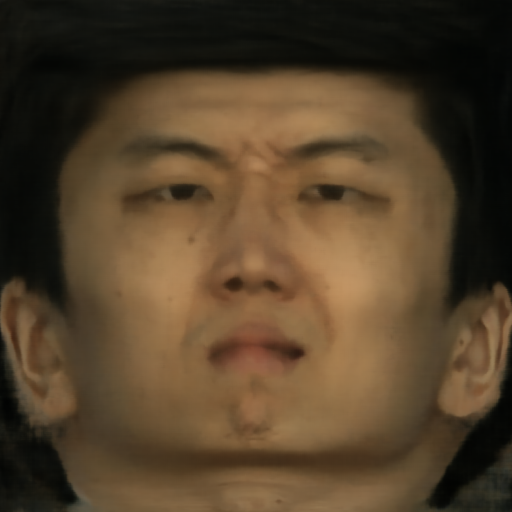}
\end{subfigure}
\hspace{\hs}
\begin{subfigure}{\ws\linewidth}
\includegraphics[width=\linewidth]{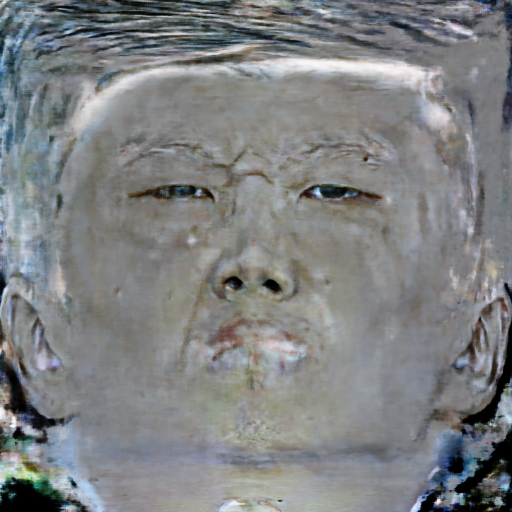}
\end{subfigure}
\hspace{\hs}
\begin{subfigure}{\ws\linewidth}
\includegraphics[width=\linewidth]{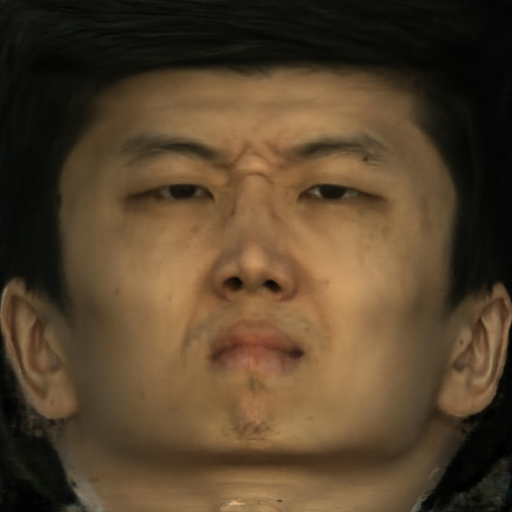}
\end{subfigure}
\hspace{\hs}
\begin{subfigure}{\wsa\linewidth}
\includegraphics[width=\linewidth]{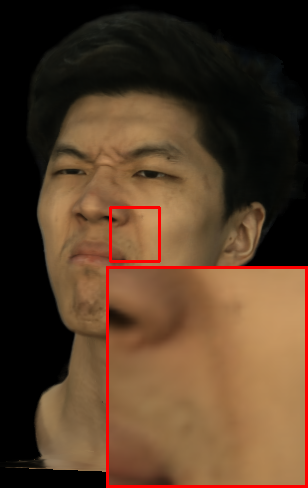}
\end{subfigure}
\vspace{\vs}
\hfill

\hfill
\rotatebox[origin=c]{90}{\makebox{\centering $0.05$}}
\begin{subfigure}{\ws\linewidth}
\includegraphics[width=\linewidth]{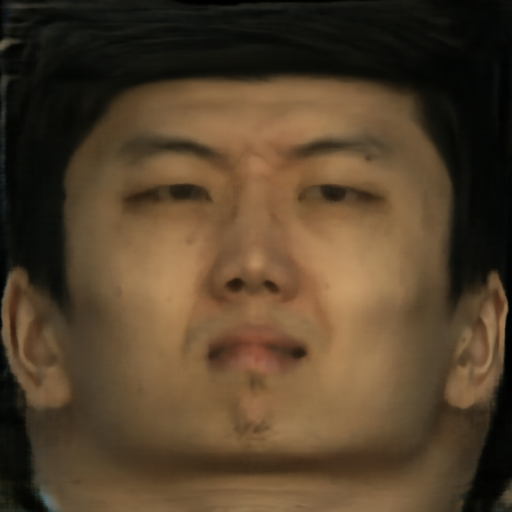}
\end{subfigure}
\hspace{\hs}
\begin{subfigure}{\ws\linewidth}
\includegraphics[width=\linewidth]{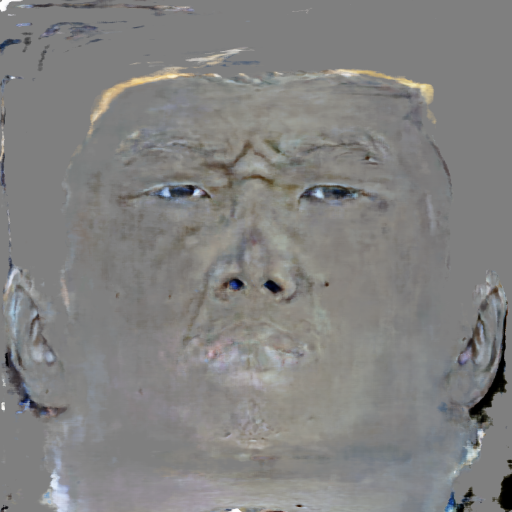}
\end{subfigure}
\hspace{\hs}
\begin{subfigure}{\ws\linewidth}
\includegraphics[width=\linewidth]{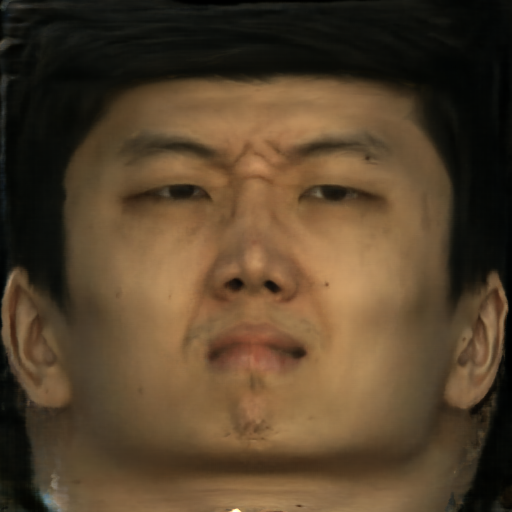}
\end{subfigure}
\hspace{\hs}
\begin{subfigure}{\wsa\linewidth}
\includegraphics[width=\linewidth]{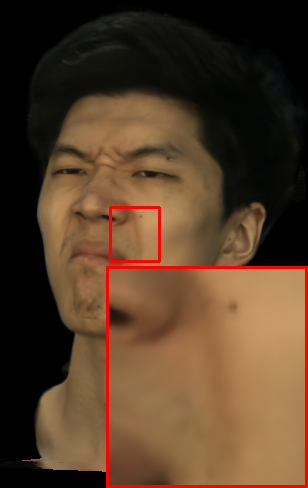}
\end{subfigure}
\vspace{\vs}
\hfill

\hfill
\rotatebox[origin=c]{90}{\makebox{\centering $\lambda_{sparsity} = 0.5$}}
\begin{subfigure}{\ws\linewidth}
\includegraphics[width=\linewidth]{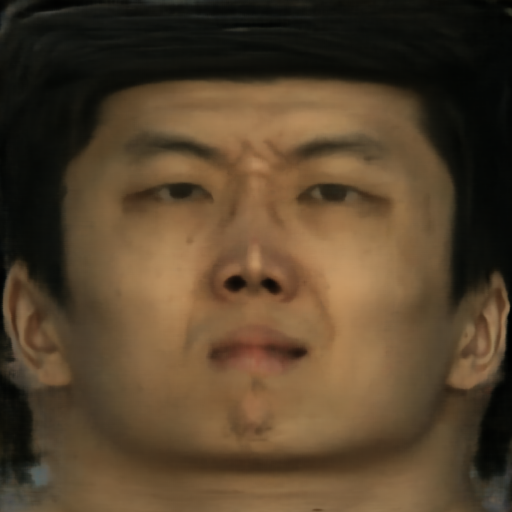}
\caption{$T_E$}
\end{subfigure}
\hspace{\hs}
\begin{subfigure}{\ws\linewidth}
\includegraphics[width=\linewidth]{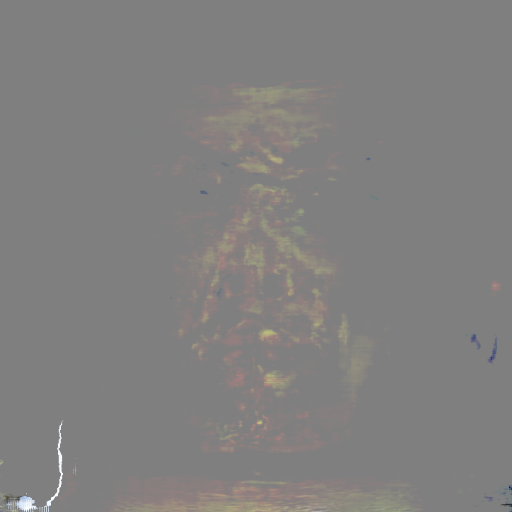}
\caption{$T_I$}
\end{subfigure}
\hspace{\hs}
\begin{subfigure}{\ws\linewidth}
\includegraphics[width=\linewidth]{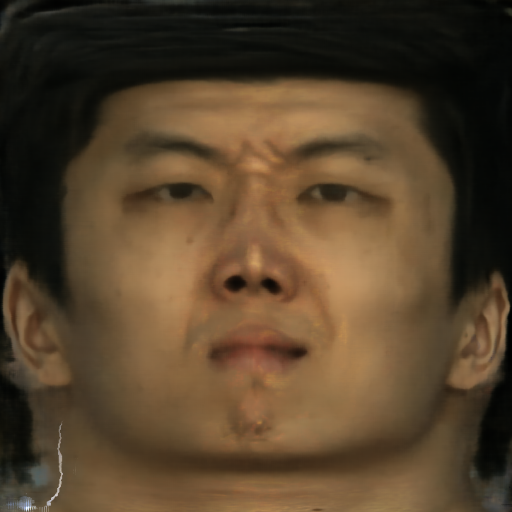}
\caption{$T$}
\end{subfigure}
\hspace{\hs}
\begin{subfigure}{\wsa\linewidth}
\includegraphics[width=\linewidth]{imgs/ablationalpha_bigalpha_render.png}
\caption{$RGB$}
\end{subfigure}
\vspace{\vs}
\hfill

\caption{Visualizations of the explicit texture $T_E$, the implicit texture $T_I$, the final texture $T$ and the rendering results using different $\lambda_{sparsity}$ settings. A darker color in $T_I$ indicates a smaller value, and the gray color indicates the $T_I = 0$. We use $\lambda_{sparsity} = 0.05$ for all of the experiments, which successfully decomposes the temporal variations to $T_I$. }
\label{fig:ablation_sparsity}
\end{figure}
\begin{figure}
\newcommand\ws{0.3}
% * 308 / 488
\newcommand\wsa{0.15778689}
\newcommand\hs{-1.6mm}
\newcommand\vs{-1mm}
\centering

\hfill
\rotatebox[origin=c]{90}{\makebox{\centering Ours}}
\begin{subfigure}{\ws\linewidth}
\includegraphics[width=\linewidth]{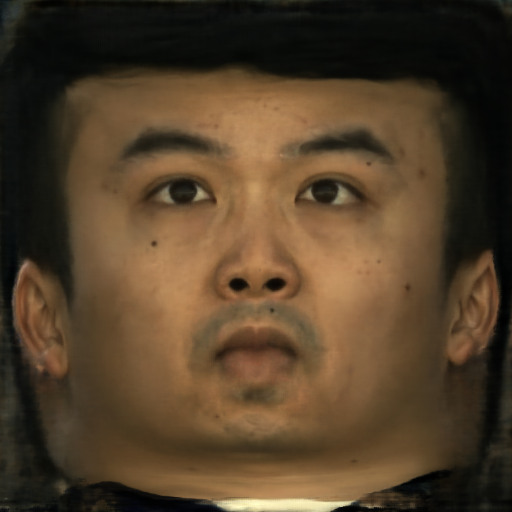}
\end{subfigure}
\hspace{\hs}
\begin{subfigure}{\ws\linewidth}
\includegraphics[width=\linewidth]{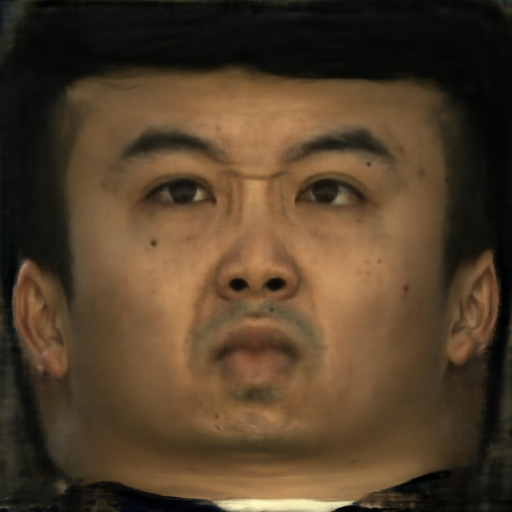}
\end{subfigure}
\hspace{\hs}
\begin{subfigure}{\ws\linewidth}
\includegraphics[width=\linewidth]{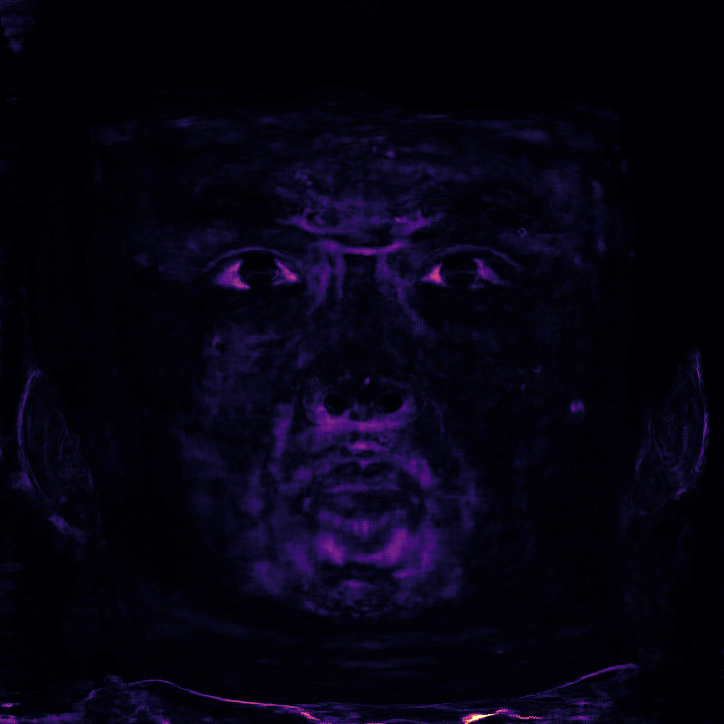}
\end{subfigure}
\vspace{\vs}
\hfill

\hfill
\rotatebox[origin=c]{90}{\makebox{\centering w/o 2 stage}}
\begin{subfigure}{\ws\linewidth}
\includegraphics[width=\linewidth]{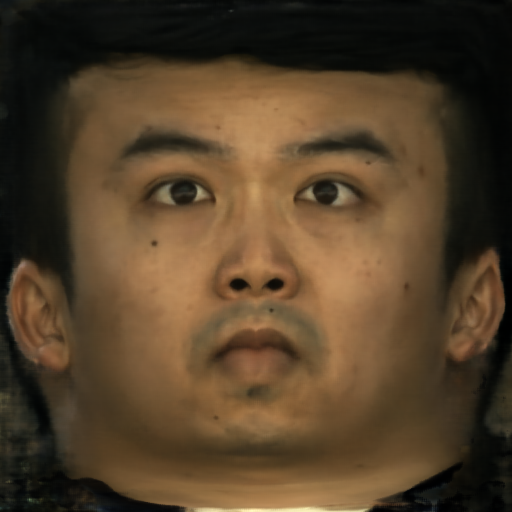}
\caption{Frame 1}
\end{subfigure}
\hspace{\hs}
\begin{subfigure}{\ws\linewidth}
\includegraphics[width=\linewidth]{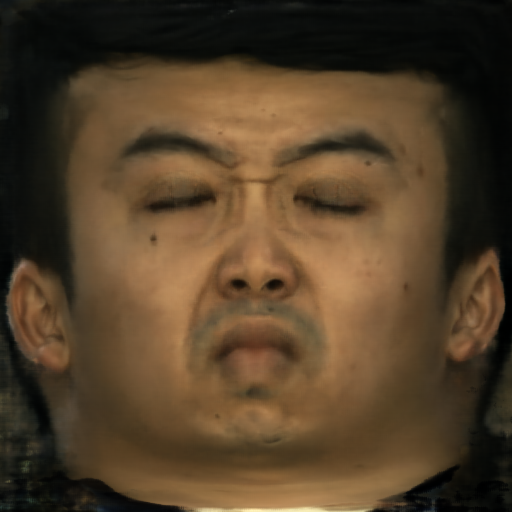}
\caption{Frame 25}
\end{subfigure}
\hspace{\hs}
\begin{subfigure}{\ws\linewidth}
\includegraphics[width=\linewidth]{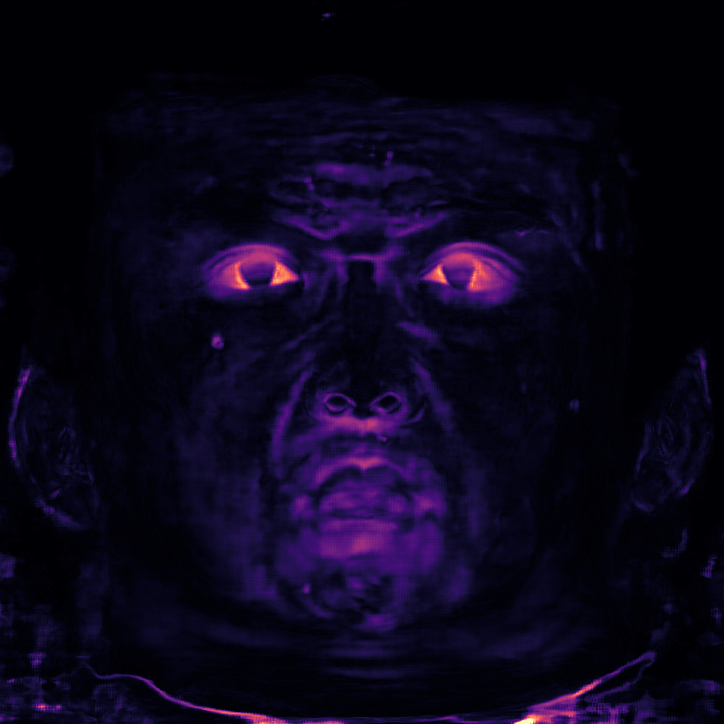}
\caption{\textit{STD}}
\end{subfigure}
\vspace{\vs}
\hfill

\caption{Visualizations of the texture $T$ in different frames of the same sequence and the \textit{STD} maps. With the two-stage training strategy, textures tend to have better temporal alignment. }
\vspace{-0.5cm}
\label{fig:ablation_1stage}
\end{figure}
% \input{figs_and_tabs/ablation_table}

% \begin{figure*}[t]
% \centering

% \hfill
% \begin{subfigure}{0.49\linewidth}
% \includegraphics[width=\linewidth]{imgs/ablationangle_plot.png}
% \caption{Plot of PSNR and \textit{Area} metrics under different $\lambda_{angle}$.}
% \label{subfig:ablation_angle_plot}
% \end{subfigure}
% \begin{subfigure}{0.49\linewidth}
% \includegraphics[width=\linewidth]{imgs/ablationangle_visual.png}
% \caption{Visualizations under $\lambda_{angle} = 0.0001$ and $\lambda_{angle} = 0.05$.}
% \label{subfig:ablation_angle_vis}
% \end{subfigure}
% \hfill

% \caption{Ablations of the $\mathcal{L}_{angle}$. Smaller loss weight $\lambda_{angle}$ has better reconstruction quality, but leads to noisy $uv$ mapping and poor angle and area persevering, while bigger weight leads to diverge of the model. }
% \end{figure*}

\begin{figure}[t]
\centering

\includegraphics[width=\linewidth]{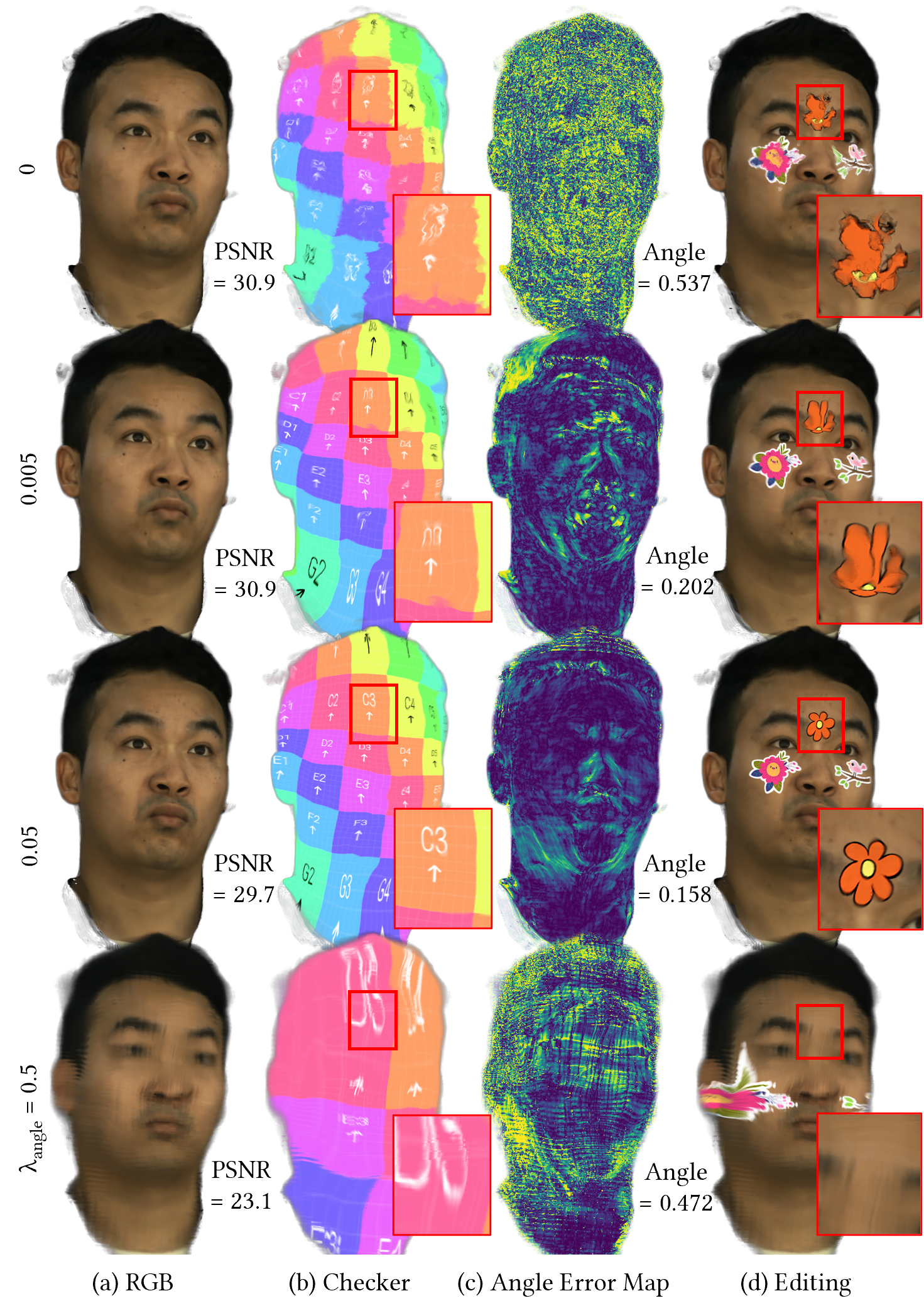}
\caption{Visualizations under different settings of $\lambda_{angle}$. The Angle Error Map is a color-coded map for the per-pixel angle error, the darker color indicates a smaller error. A smaller weight $\lambda_{angle}$ results in better reconstruction quality, but leads to noisy UV mapping and poor angle and area preservation. On the other hand, larger weight leads to divergence of the model.}
\label{subfig:ablation_angle_vis}
\vspace{-0.3cm}
\end{figure}

\subsection{Ablations}
% We conduct ablations on the multi-view videos dataset that consists of $5$ subjects. We remove or modify one component at a time to validate the effectiveness of that component. We show numeric results in Tab.~\ref{tab:ablation}.
We conduct a series of ablations on the multi-view videos dataset, where we remove or modify one component at a time to validate the effectiveness of each component.

\vspace{-2mm}
\paragraph{Sparsity Loss.}
We first validate the necessity of $\mathcal{L}_{sparsity}$. We visualize the explicit textures $T_E$, implicit textures $T_I$, the final textures $T$ and the rendering results of different loss weights $\lambda_{sparsity}$ in Fig.~\ref{fig:ablation_sparsity}. Recall that we want the temporal variations to be modeled as residuals using $T_I$ so we can edit the explicit texture $T_E$. Results show that a smaller $\lambda_{sparsity}$ leaves too much information on $T_I$ and a bigger $\lambda_{sparsity}$ eliminates important temporal variations such as the wrinkles.
% We also find that applying $\mathcal{L}_{sparsity}$ helps to improve the temporal consistency and quality of the UV mapping.

\vspace{-2mm}
\paragraph{Two-Stage Training.}
Then we test the effectiveness of the proposed two-stage training strategy. We show the textures with and without two-stage training in Fig.~\ref{fig:ablation_1stage}. Results show that the two-stage strategy reduces the temporal variations in the dynamic textures and achieves better temporal alignment, especially in the eye and mouth regions. 

\vspace{-2mm}
\paragraph{Angle Loss.}
We experiment with different $\lambda_{angle}$ and 
% plot the relations between the reconstruction quality (PSNR) and the editability (\textit{Area}) in Fig.~\ref{subfig:ablation_angle_plot}. We also 
visualize the rendering results, UV checker visualizations, \textit{Angle} error maps, and editing examples in Fig.~\ref{subfig:ablation_angle_vis}. We note that a small weight of $\mathcal{L}_{angle}$ will produce better reconstruction quality since the UV mapping is less regularized. However, this will leads to a noisy UV mapping, which is not suitable for editing since the editing results will also be noisy. It also has poor performance in terms of angle preservation. On the other hand, we find that the model does not converge to plausible results after increasing $\lambda_{angle}$ beyond a threshold. In summary, setting $\lambda_{angle} = 0.05$ achieves the best trade-off between reconstruction and editability. We used this setting in all of our experiments. 

\begin{figure*}[t]
    \centering
    \includegraphics[width=\linewidth]{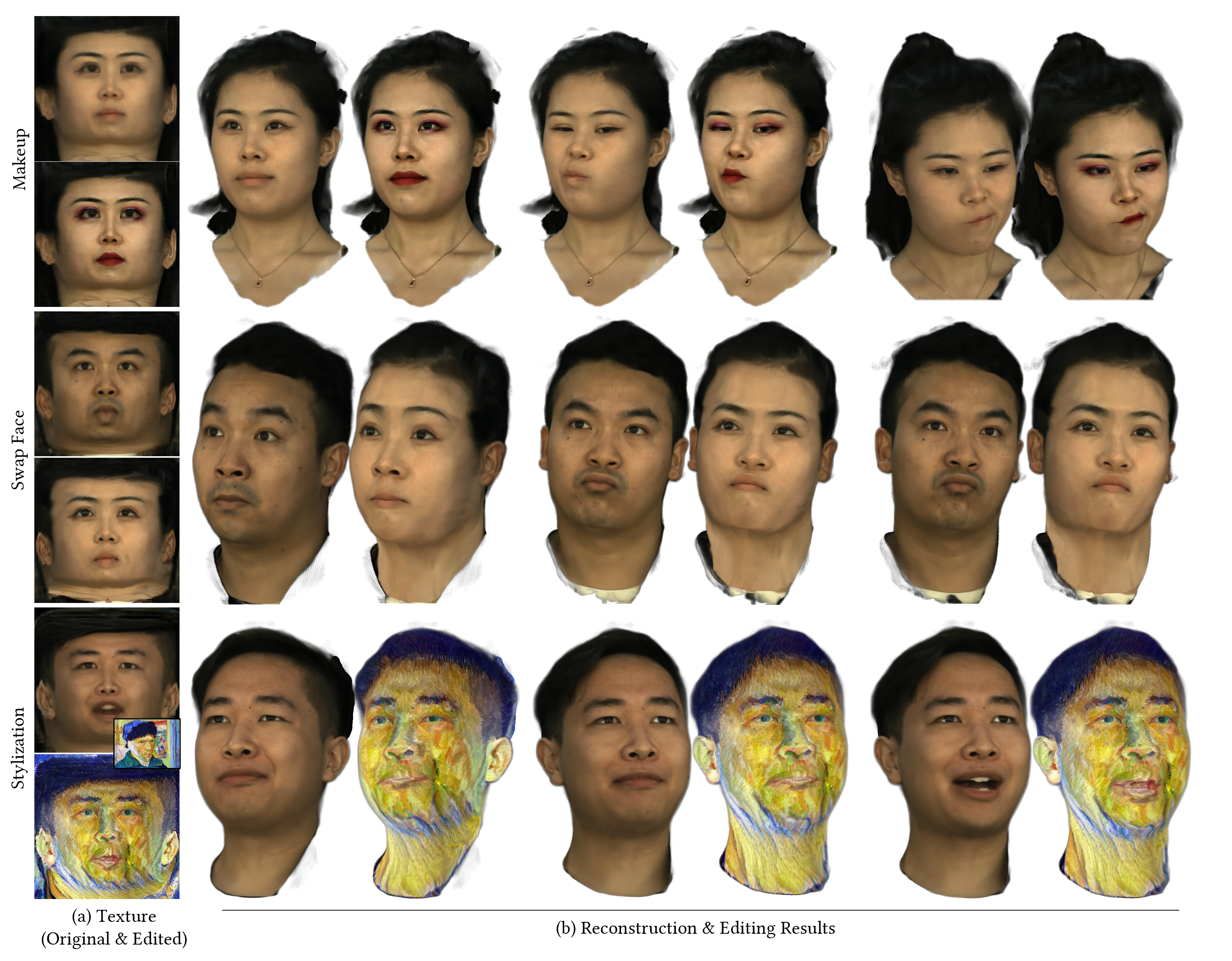}
    \vspace{-7mm}
    \caption{Examples of appearance editing.}
    \vspace{-4mm}
    \label{fig:app_app}
\end{figure*}

\begin{figure}[t]
    \centering
    \includegraphics[width=\linewidth]{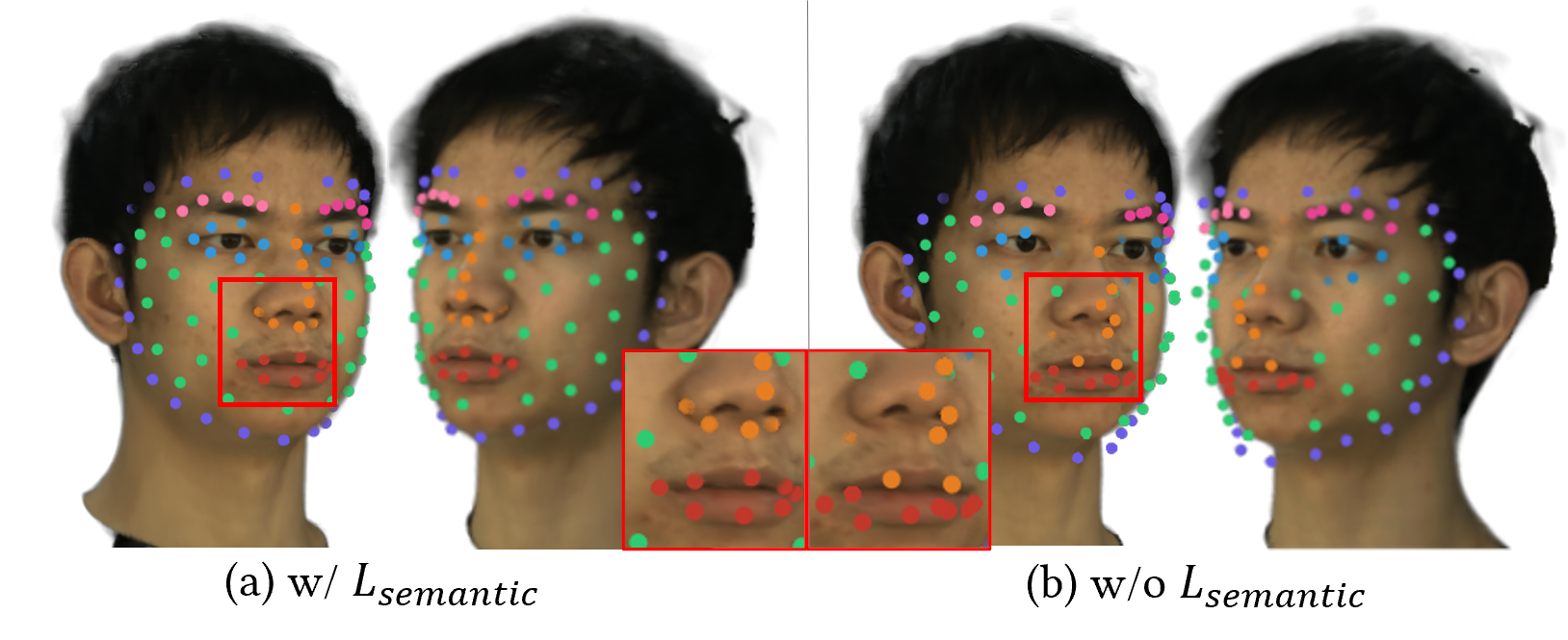}
    \caption{Visualization of control points with and without $\mathcal{L}_{semantic}$.}
    \label{fig:ablation_semantic}
    \vspace{-0.5cm}
\end{figure}
\begin{figure}
    \centering
    \includegraphics[width=\linewidth]{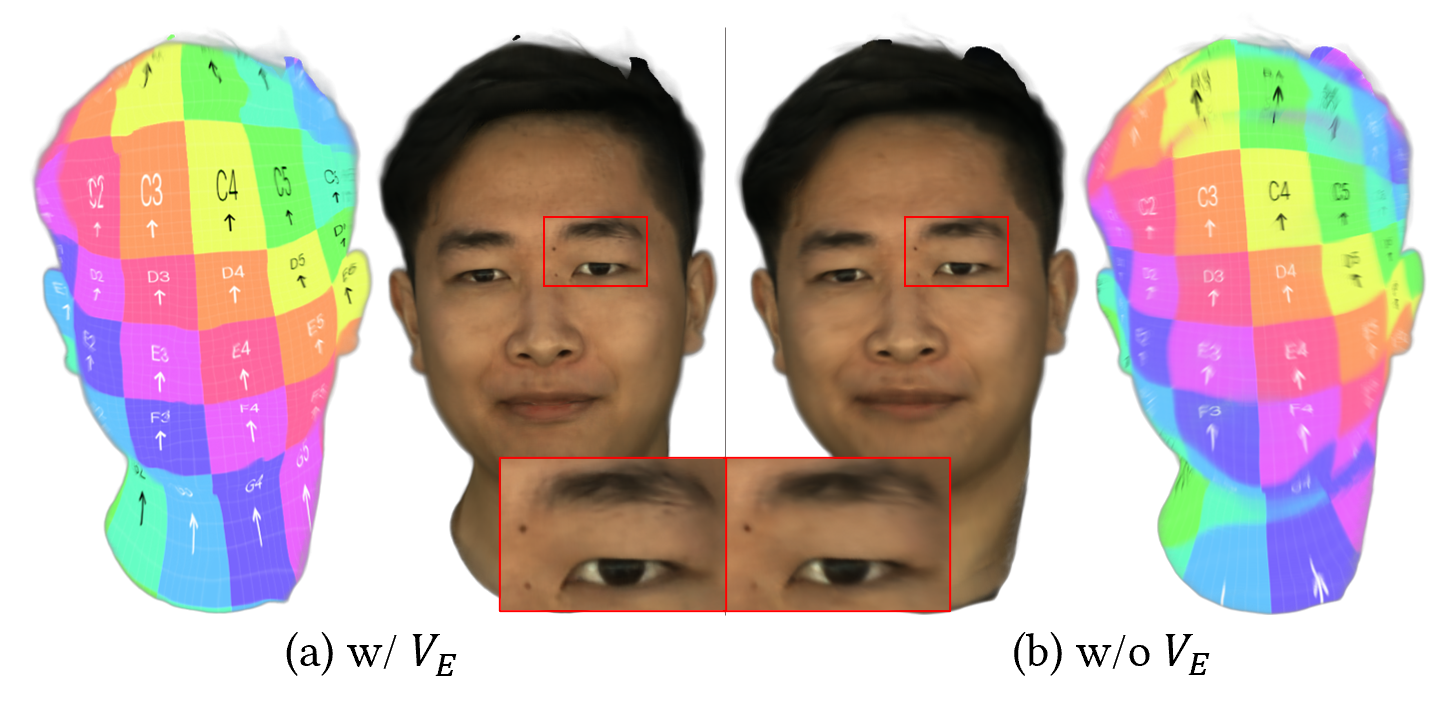}
    \caption{Ablations with and without explicit warping module $V_E$. Without $V_E$, the reconstruction tends to be blurrier.}
    \label{fig:ablation_geowarp}
\end{figure}

% Uncommment this to show the figures
\begin{figure*}[t]
    \centering
    \includegraphics[width=0.95\linewidth]{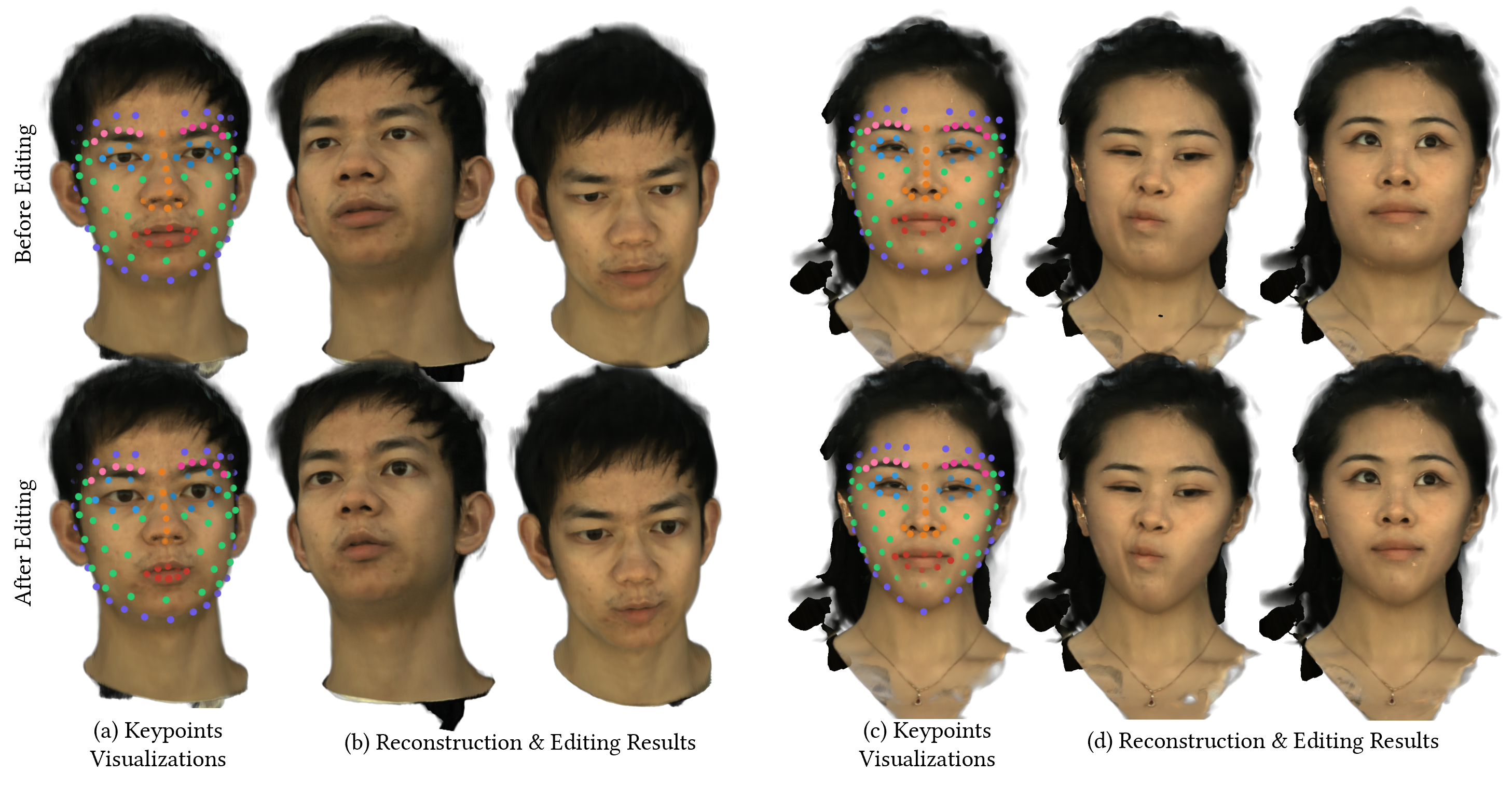}
    \vspace{-0.4cm}
    \caption{Examples of geometry editing.}
    \vspace{-0.4cm}
    \label{fig:app_geo}
\end{figure*}

\begin{figure*}[t]
    \centering
    \includegraphics[width=0.95\linewidth]{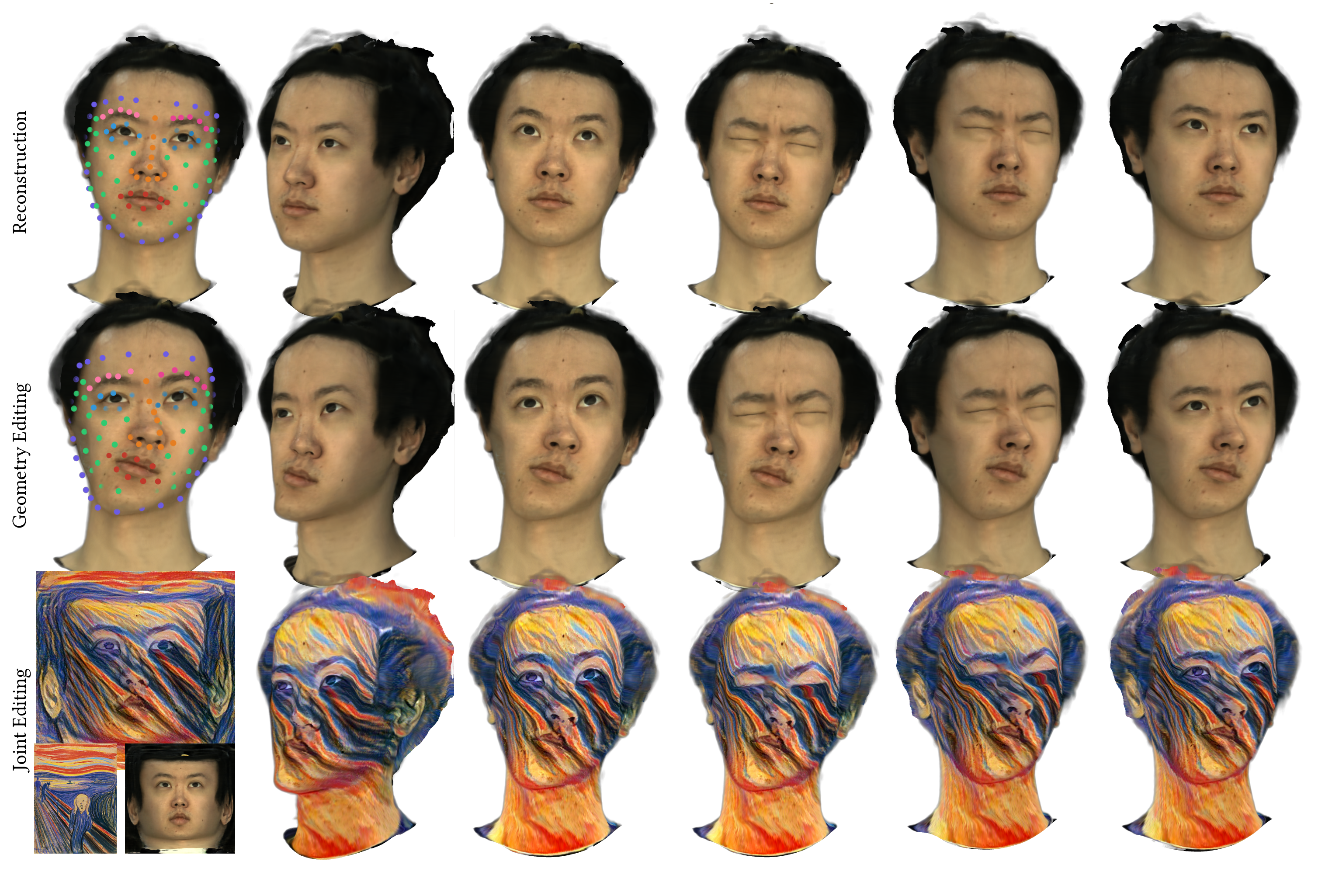}
    \vspace{-0.4cm}
    \caption{Examples of joint appearance editing and geometry editing. }
    \vspace{-0.4cm}
    \label{fig:app_joint}
\end{figure*}

\begin{figure}[t]
    \centering
    \includegraphics[width=\linewidth]{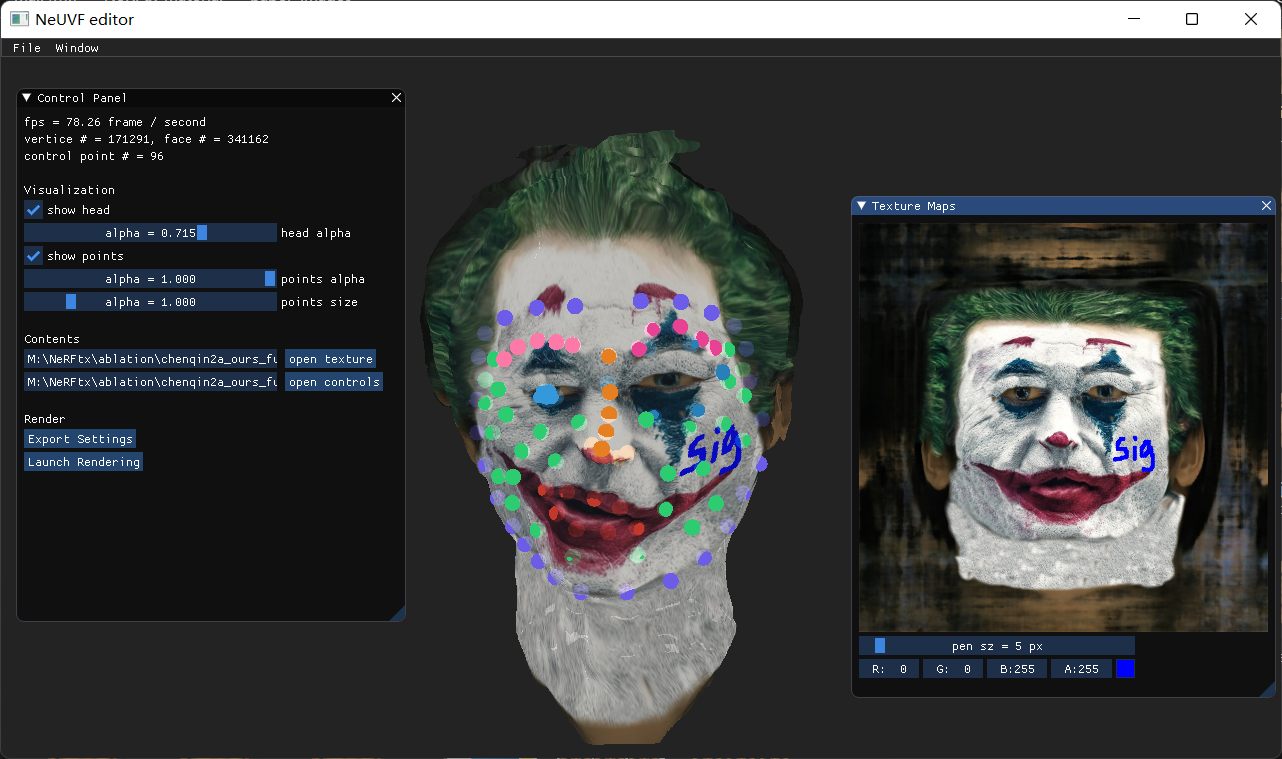}
    \vspace{-6mm}
    \caption{A screen shot of our user interface.}
    \vspace{-4mm}
    \label{fig:app_ui}
\end{figure}

\vspace{-2mm}
\paragraph{Explicit Deformation Layer.}
We first demonstrate the necessity of $\mathcal{L}_{semantic}$ by training the model without this term. We show one such example in Fig.~\ref{fig:ablation_semantic}. It can be seen that freely optimizing the explicit deformation parameters will not converge to semantically meaningful control points, which makes the editing inconvenient.

We also find that although the design of the explicit deformation layer $V_E$ is to enable geometry editing, it actually helps improve the reconstruction quality of the radiance field. We show one such example in Fig.~\ref{fig:ablation_geowarp}. The reconstruction is sharper with the explicit deformation layer. This is due to the fact that the explicit deformation is initialized from the mesh tracking results, which helps to make the training of the entire model easier.

\section{Applications}
\label{sec:application}

In this section, we demonstrate several applications that are made possible by using our model. These include geometry editing (Fig.~\ref{fig:teaser}b, Fig.~\ref{fig:app_geo}), appearance editing (Fig.~\ref{fig:teaser}e, Fig.~\ref{fig:app_app}) and joint editing (Fig.~\ref{fig:teaser}cf, Fig.~\ref{fig:app_joint}). We edit the appearance by editing the explicit texture $T_E$. We do the face swapping by manually aligning the source texture map to the texture map of the editing target using photo processing software. And we use the method proposed by \citet[]{misc_STROTSS} for stylizing the textures. For geometry editing, we manually adjust the control points in one frame using our UI, and then propagate the delta positions of each control point to the other frames. As shown in Fig.~\ref{fig:app_ui}, we develop an interactive UI for intuitively editing the control points and the texture map. Video editing results, as well as an editing UI demo, can be found in the supplementary video.
% For appearance editing, we provide an interface for loading arbitrary textures, and users could use any existing image editing software to edit the texture, and then load into our UI. 
To produce a preview of the rendering results, we perform ray marching on the density field and extract a coarse mesh. We then forward our model to get the UV coordinates of each vertex, along with the texture map. In the UI, users can freely adjust the control points. We deform all the vertices of the coarse mesh using the control points based on the method described in Sec.~\ref{sec:geometry}. The UI is implemented in OpenGL and runs at over 70fps on a commercial laptop. After the editing, offline volume rendering is used to render the final results.

%% TODO: how to edit
\section{Discussion and Limitations}
\label{sec:discussion}
In this paper, we present a method for reconstructing and editing 3D dynamic human heads using multi-view videos as input. The method extends the implicit representation and improves editability by introducing explicit geometry and appearance layers. We show that our method achieves photorealistic reconstruction while enabling consistent, intuitive, and powerful editing of both appearance and geometry. Although our main focus here is the human head, it is relatively easy to extend to other objects as long as a tracked coarse mesh is available. 

% slow training
Our method comes with some limitations. Like most NeRF-based methods, training a model takes days, which may prevent it from practical use in some settings. We believe this could be improved with the help of existing methods for accelerating NeRF training \cite{misc_instantNGP,volume_plenoxels}. 

% reconstruction error
\begin{figure}[t]
    \centering
    \includegraphics[width=0.97\linewidth]{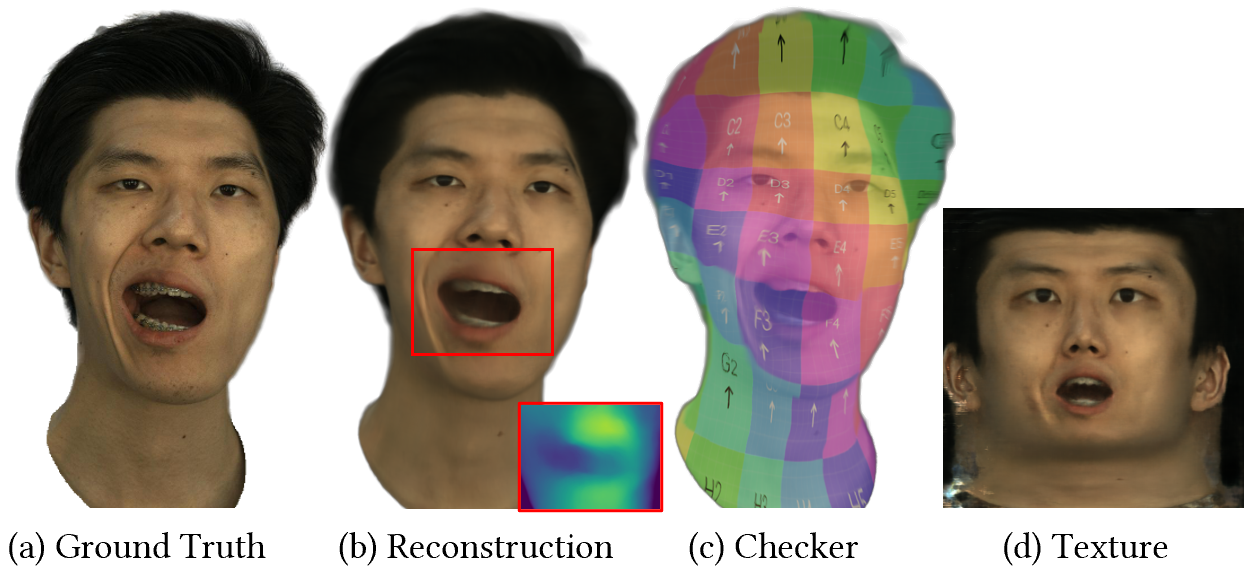}
    \vspace{-0.2cm}
    \caption{Results of a challenging mouth example. The inset visualizes the reconstructed depth map of the mouth region. Our method tends to model the mouth interior using coarse geometry and view dependent textures.}
    \vspace{-0.4cm}
    \label{fig:discuss_mouthrecon}
\end{figure}
Our method sometimes could not achieve the same level of reconstruction accuracy as other dynamic NeRF methods, which focus only on reconstruction, as illustrated in the experiment section. 
For challenging area such as mouth interior, our method tends to fit using coarse geometry and view-dependent textures. One example is shown in Fig.~\ref{fig:discuss_mouthrecon}. It can be seen that the reconstructed depth map in the mouth region is over-smoothed.
Furthermore, our appearance representation bakes in the lighting to the view-dependent texture. Therefore, it is still challenging to edit the lighting and the material. An interesting future work would be to introduce a lighting model and physically based rendering techniques so that we could have more control over the shading.

\begin{figure}[t]
    \centering
    \includegraphics[width=0.97\linewidth]{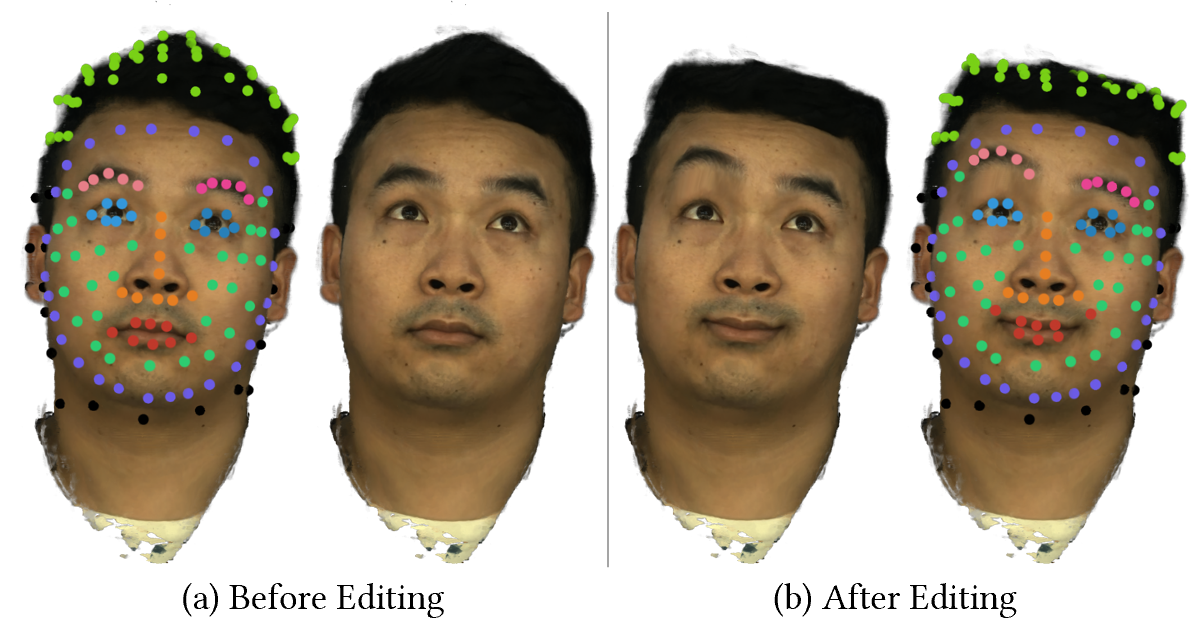}
    \vspace{-0.4cm}
    \caption{An example of expression and hair editing. Extra control points are used for editing the hair.}
    \vspace{-0.4cm}
    \label{fig:discuss_moreediting}
\end{figure}
% poor geometry editing
To enable geometry editing, we adopt a simple yet intuitive geometric editing module driven by sparse control points. While our primary focus is to consistently edit the facial attribute of a dynamic head (e.g. size of the eye, shape of face contour), we show that it is possible to extend our method for editing one's expression. By adding more control points on the head we could perform basic hair editing. We show one example of expression and hair editing in Fig.~\ref{fig:discuss_moreediting}. However, one limitation of our geometry editing module is that it cannot handle topology changes, such as opening the mouth. A potential solution is to develop more advanced interpolation techniques that allow modeling discontinuities while also supporting explicit editing.
% \remove{For the geometry editing, since our could only model a smooth deformation field, it is still incapable of handling editing that contains topology changes, such as opening a mouth. A potential solution is to develop more advanced interpolation technologies that allow modeling discontinuity while also supporting explicit editing. }

\begin{acks}
The authors from HKUST were partly supported by the Hong Kong Research Grants Council (RGC). The author from CityU was supported by DON\_RMG grant from CityU, Hong Kong (Project No. 9229064).  
% We also want to thank Yuze He, Qin Chen, Weiyu Li, Xin Huang, Qihang Fang and Benwang Chen for providing the face data.
\end{acks}

\bibliographystyle{ACM-Reference-Format}
\bibliography{sample-base}


\begin{thebibliography}{100}

%%% the Web address.

\ifx \showCODEN    \undefined \def \showCODEN     #1{\unskip}     \fi
\ifx \showDOI      \undefined \def \showDOI       #1{#1}\fi
\ifx \showISBNx    \undefined \def \showISBNx     #1{\unskip}     \fi
\ifx \showISBNxiii \undefined \def \showISBNxiii  #1{\unskip}     \fi
\ifx \showISSN     \undefined \def \showISSN      #1{\unskip}     \fi
\ifx \showLCCN     \undefined \def \showLCCN      #1{\unskip}     \fi
\ifx \shownote     \undefined \def \shownote      #1{#1}          \fi
\ifx \showarticletitle \undefined \def \showarticletitle #1{#1}   \fi
\ifx \showURL      \undefined \def \showURL       {\relax}        \fi
% invisible to TeX
\providecommand\bibfield[2]{#2}
\providecommand\bibinfo[2]{#2}
\providecommand\natexlab[1]{#1}
\providecommand\showeprint[2][]{arXiv:#2}

\bibitem[Bai et~al\mbox{.}(2020)]%
        {mesh_face_dfnrmvs}
\bibfield{author}{\bibinfo{person}{Ziqian Bai}, \bibinfo{person}{Zhaopeng Cui},
  \bibinfo{person}{Jamal~Ahmed Rahim}, \bibinfo{person}{Xiaoming Liu}, {and}
  \bibinfo{person}{Ping Tan}.} \bibinfo{year}{2020}\natexlab{}.
\newblock \showarticletitle{Deep facial non-rigid multi-view stereo}. In
  \bibinfo{booktitle}{\emph{Proceedings of the IEEE/CVF Conference on Computer
  Vision and Pattern Recognition}}. \bibinfo{pages}{5850--5860}.
\newblock


\bibitem[Bao et~al\mbox{.}(2020)]%
        {3dmm_dr_hifi3d}
\bibfield{author}{\bibinfo{person}{Linchao Bao}, \bibinfo{person}{Xiangkai
  Lin}, \bibinfo{person}{Yajing Chen}, \bibinfo{person}{Haoxian Zhang},
  \bibinfo{person}{Sheng Wang}, \bibinfo{person}{Xuefei Zhe},
  \bibinfo{person}{Di Kang}, \bibinfo{person}{Haozhi Huang},
  \bibinfo{person}{Xinwei Jiang}, \bibinfo{person}{Jue Wang}, {et~al\mbox{.}}}
  \bibinfo{year}{2020}\natexlab{}.
\newblock \showarticletitle{High-Fidelity 3D Digital Human Head Creation from
  RGB-D Selfies}.
\newblock \bibinfo{journal}{\emph{arXiv preprint arXiv:2010.05562}}
  (\bibinfo{year}{2020}).
\newblock


\bibitem[Beeler et~al\mbox{.}(2011)]%
        {Beeler11}
\bibfield{author}{\bibinfo{person}{Thabo Beeler}, \bibinfo{person}{Fabian
  Hahn}, \bibinfo{person}{Derek Bradley}, \bibinfo{person}{Bernd Bickel},
  \bibinfo{person}{Paul Beardsley}, \bibinfo{person}{Craig Gotsman},
  \bibinfo{person}{Robert~W. Sumner}, {and} \bibinfo{person}{Markus Gross}.}
  \bibinfo{year}{2011}\natexlab{}.
\newblock \showarticletitle{High-quality passive facial performance capture
  using anchor frames}.
\newblock \bibinfo{journal}{\emph{ACM Trans. Graph.}}  \bibinfo{volume}{30},
  Article \bibinfo{articleno}{75} (\bibinfo{date}{August}
  \bibinfo{year}{2011}), \bibinfo{numpages}{10}~pages.
\newblock
Issue 4.
\showISSN{0730-0301}
\urldef\tempurl%
\url{https://doi.org/10.1145/2010324.1964970}
\showDOI{\tempurl}


\bibitem[Bernardini et~al\mbox{.}(1999)]%
        {mesh_ballpivot}
\bibfield{author}{\bibinfo{person}{Fausto Bernardini}, \bibinfo{person}{Joshua
  Mittleman}, \bibinfo{person}{Holly Rushmeier}, \bibinfo{person}{Cl{\'a}udio
  Silva}, {and} \bibinfo{person}{Gabriel Taubin}.}
  \bibinfo{year}{1999}\natexlab{}.
\newblock \showarticletitle{The ball-pivoting algorithm for surface
  reconstruction}.
\newblock \bibinfo{journal}{\emph{IEEE transactions on visualization and
  computer graphics}} \bibinfo{volume}{5}, \bibinfo{number}{4}
  (\bibinfo{year}{1999}), \bibinfo{pages}{349--359}.
\newblock


\bibitem[Bi et~al\mbox{.}(2021)]%
        {mesh_appear_DRAM}
\bibfield{author}{\bibinfo{person}{Sai Bi}, \bibinfo{person}{Stephen Lombardi},
  \bibinfo{person}{Shunsuke Saito}, \bibinfo{person}{Tomas Simon},
  \bibinfo{person}{Shih-En Wei}, \bibinfo{person}{Kevyn Mcphail},
  \bibinfo{person}{Ravi Ramamoorthi}, \bibinfo{person}{Yaser Sheikh}, {and}
  \bibinfo{person}{Jason Saragih}.} \bibinfo{year}{2021}\natexlab{}.
\newblock \showarticletitle{Deep relightable appearance models for animatable
  faces}.
\newblock \bibinfo{journal}{\emph{ACM Transactions on Graphics (TOG)}}
  \bibinfo{volume}{40}, \bibinfo{number}{4} (\bibinfo{year}{2021}),
  \bibinfo{pages}{1--15}.
\newblock


\bibitem[Bi et~al\mbox{.}(2020)]%
        {nerf_relit_reflectance}
\bibfield{author}{\bibinfo{person}{Sai Bi}, \bibinfo{person}{Zexiang Xu},
  \bibinfo{person}{Pratul Srinivasan}, \bibinfo{person}{Ben Mildenhall},
  \bibinfo{person}{Kalyan Sunkavalli}, \bibinfo{person}{Miloš Hašan},
  \bibinfo{person}{Yannick Hold-Geoffroy}, \bibinfo{person}{David Kriegman},
  {and} \bibinfo{person}{Ravi Ramamoorthi}.} \bibinfo{year}{2020}\natexlab{}.
\newblock \bibinfo{title}{Neural Reflectance Fields for Appearance
  Acquisition}.
\newblock
\newblock
\showeprint[arxiv]{2008.03824}~[cs.CV]


\bibitem[Blanz and Vetter(1999)]%
        {3dmm_0}
\bibfield{author}{\bibinfo{person}{Volker Blanz} {and} \bibinfo{person}{Thomas
  Vetter}.} \bibinfo{year}{1999}\natexlab{}.
\newblock \showarticletitle{A Morphable Model for the Synthesis of 3D Faces}.
  In \bibinfo{booktitle}{\emph{Proceedings of the 26th Annual Conference on
  Computer Graphics and Interactive Techniques}}
  \emph{(\bibinfo{series}{SIGGRAPH '99})}. \bibinfo{publisher}{ACM
  Press/Addison-Wesley Publishing Co.}, \bibinfo{address}{USA},
  \bibinfo{pages}{187–194}.
\newblock
\showISBNx{0201485605}
\urldef\tempurl%
\url{https://doi.org/10.1145/311535.311556}
\showDOI{\tempurl}


\bibitem[Bojanowski et~al\mbox{.}(2017)]%
        {dict_embed}
\bibfield{author}{\bibinfo{person}{Piotr Bojanowski}, \bibinfo{person}{Armand
  Joulin}, \bibinfo{person}{David Lopez-Paz}, {and} \bibinfo{person}{Arthur
  Szlam}.} \bibinfo{year}{2017}\natexlab{}.
\newblock \bibinfo{title}{Optimizing the Latent Space of Generative Networks}.
\newblock
\newblock
\showeprint[arxiv]{1707.05776}~[stat.ML]


\bibitem[Booth et~al\mbox{.}(2018)]%
        {3dmm_LSFM}
\bibfield{author}{\bibinfo{person}{James Booth}, \bibinfo{person}{Anastasios
  Roussos}, \bibinfo{person}{Allan Ponniah}, \bibinfo{person}{David Dunaway},
  {and} \bibinfo{person}{Stefanos Zafeiriou}.} \bibinfo{year}{2018}\natexlab{}.
\newblock \showarticletitle{Large scale 3d morphable models}.
\newblock \bibinfo{journal}{\emph{International Journal of Computer Vision}}
  \bibinfo{volume}{126}, \bibinfo{number}{2} (\bibinfo{year}{2018}),
  \bibinfo{pages}{233--254}.
\newblock


\bibitem[Boss et~al\mbox{.}(2021)]%
        {nerf_relit_nerd}
\bibfield{author}{\bibinfo{person}{Mark Boss}, \bibinfo{person}{Raphael Braun},
  \bibinfo{person}{Varun Jampani}, \bibinfo{person}{Jonathan~T. Barron},
  \bibinfo{person}{Ce Liu}, {and} \bibinfo{person}{Hendrik~P.A. Lensch}.}
  \bibinfo{year}{2021}\natexlab{}.
\newblock \showarticletitle{NeRD: Neural Reflectance Decomposition from Image
  Collections}.
\newblock \bibinfo{journal}{\emph{2021 IEEE/CVF International Conference on
  Computer Vision (ICCV)}} (\bibinfo{date}{Oct} \bibinfo{year}{2021}).
\newblock
\urldef\tempurl%
\url{https://doi.org/10.1109/iccv48922.2021.01245}
\showDOI{\tempurl}


\bibitem[Cao et~al\mbox{.}(2021)]%
        {mesh_binocular}
\bibfield{author}{\bibinfo{person}{Chen Cao}, \bibinfo{person}{Vasu Agrawal},
  \bibinfo{person}{Fernando De~La~Torre}, \bibinfo{person}{Lele Chen},
  \bibinfo{person}{Jason Saragih}, \bibinfo{person}{Tomas Simon}, {and}
  \bibinfo{person}{Yaser Sheikh}.} \bibinfo{year}{2021}\natexlab{}.
\newblock \showarticletitle{Real-time 3D neural facial animation from binocular
  video}.
\newblock \bibinfo{journal}{\emph{ACM Transactions on Graphics (TOG)}}
  \bibinfo{volume}{40}, \bibinfo{number}{4} (\bibinfo{year}{2021}),
  \bibinfo{pages}{1--17}.
\newblock


\bibitem[Catmull and Clark(1978)]%
        {mesh_editing_catmullclark}
\bibfield{author}{\bibinfo{person}{Edwin Catmull} {and} \bibinfo{person}{James
  Clark}.} \bibinfo{year}{1978}\natexlab{}.
\newblock \showarticletitle{Recursively generated B-spline surfaces on
  arbitrary topological meshes}.
\newblock \bibinfo{journal}{\emph{Computer-aided design}} \bibinfo{volume}{10},
  \bibinfo{number}{6} (\bibinfo{year}{1978}), \bibinfo{pages}{350--355}.
\newblock


\bibitem[Chan et~al\mbox{.}(2021a)]%
        {NeRF_GAN_EG3D}
\bibfield{author}{\bibinfo{person}{Eric~R. Chan}, \bibinfo{person}{Connor~Z.
  Lin}, \bibinfo{person}{Matthew~A. Chan}, \bibinfo{person}{Koki Nagano},
  \bibinfo{person}{Boxiao Pan}, \bibinfo{person}{Shalini~De Mello},
  \bibinfo{person}{Orazio Gallo}, \bibinfo{person}{Leonidas Guibas},
  \bibinfo{person}{Jonathan Tremblay}, \bibinfo{person}{Sameh Khamis},
  \bibinfo{person}{Tero Karras}, {and} \bibinfo{person}{Gordon Wetzstein}.}
  \bibinfo{year}{2021}\natexlab{a}.
\newblock \bibinfo{title}{Efficient Geometry-aware 3D Generative Adversarial
  Networks}.
\newblock
\newblock
\showeprint[arxiv]{2112.07945}~[cs.CV]


\bibitem[Chan et~al\mbox{.}(2021b)]%
        {NeRF_GAN_pigan}
\bibfield{author}{\bibinfo{person}{Eric~R. Chan}, \bibinfo{person}{Marco
  Monteiro}, \bibinfo{person}{Petr Kellnhofer}, \bibinfo{person}{Jiajun Wu},
  {and} \bibinfo{person}{Gordon Wetzstein}.} \bibinfo{year}{2021}\natexlab{b}.
\newblock \showarticletitle{pi-GAN: Periodic Implicit Generative Adversarial
  Networks for 3D-Aware Image Synthesis}.
\newblock \bibinfo{journal}{\emph{2021 IEEE/CVF Conference on Computer Vision
  and Pattern Recognition (CVPR)}} (\bibinfo{date}{Jun} \bibinfo{year}{2021}).
\newblock
\urldef\tempurl%
\url{https://doi.org/10.1109/cvpr46437.2021.00574}
\showDOI{\tempurl}


\bibitem[Chen et~al\mbox{.}(2022)]%
        {chen2022uv}
\bibfield{author}{\bibinfo{person}{Yue Chen}, \bibinfo{person}{Xuan Wang},
  \bibinfo{person}{Qi Zhang}, \bibinfo{person}{Xiaoyu Li},
  \bibinfo{person}{Xingyu Chen}, \bibinfo{person}{Yu Guo}, \bibinfo{person}{Jue
  Wang}, {and} \bibinfo{person}{Fei Wang}.} \bibinfo{year}{2022}\natexlab{}.
\newblock \showarticletitle{UV Volumes for Real-time Rendering of Editable
  Free-view Human Performance}.
\newblock \bibinfo{journal}{\emph{arXiv preprint arXiv:2203.14402}}
  (\bibinfo{year}{2022}).
\newblock


\bibitem[Chiang et~al\mbox{.}(2022)]%
        {NeRF_edit_styleHyper}
\bibfield{author}{\bibinfo{person}{Pei-Ze Chiang}, \bibinfo{person}{Meng-Shiun
  Tsai}, \bibinfo{person}{Hung-Yu Tseng}, \bibinfo{person}{Wei-Sheng Lai},
  {and} \bibinfo{person}{Wei-Chen Chiu}.} \bibinfo{year}{2022}\natexlab{}.
\newblock \showarticletitle{Stylizing 3D Scene via Implicit Representation and
  HyperNetwork}. In \bibinfo{booktitle}{\emph{Proceedings of the IEEE/CVF
  Winter Conference on Applications of Computer Vision}}.
  \bibinfo{pages}{1475--1484}.
\newblock


\bibitem[Cole et~al\mbox{.}(2021)]%
        {mesh_dr1_sampling}
\bibfield{author}{\bibinfo{person}{Forrester Cole}, \bibinfo{person}{Kyle
  Genova}, \bibinfo{person}{Avneesh Sud}, \bibinfo{person}{Daniel Vlasic},
  {and} \bibinfo{person}{Zhoutong Zhang}.} \bibinfo{year}{2021}\natexlab{}.
\newblock \showarticletitle{Differentiable surface rendering via
  non-differentiable sampling}. In \bibinfo{booktitle}{\emph{Proceedings of the
  IEEE/CVF International Conference on Computer Vision}}.
  \bibinfo{pages}{6088--6097}.
\newblock


\bibitem[Deng et~al\mbox{.}(2021)]%
        {NeRF_edit_Deformed}
\bibfield{author}{\bibinfo{person}{Yu Deng}, \bibinfo{person}{Jiaolong Yang},
  {and} \bibinfo{person}{Xin Tong}.} \bibinfo{year}{2021}\natexlab{}.
\newblock \showarticletitle{Deformed implicit field: Modeling 3d shapes with
  learned dense correspondence}. In \bibinfo{booktitle}{\emph{Proceedings of
  the IEEE/CVF Conference on Computer Vision and Pattern Recognition}}.
  \bibinfo{pages}{10286--10296}.
\newblock


\bibitem[{Epic Games}(2018)]%
        {mesh_sfm_CR}
\bibfield{author}{\bibinfo{person}{{Epic Games}}.}
  \bibinfo{year}{2018}\natexlab{}.
\newblock \bibinfo{title}{CapturingReality}.
\newblock
\newblock
\urldef\tempurl%
\url{https://www.capturingreality.com}
\showURL{%
\tempurl}


\bibitem[Feng et~al\mbox{.}(2021)]%
        {3dmm_dr_DECA}
\bibfield{author}{\bibinfo{person}{Yao Feng}, \bibinfo{person}{Haiwen Feng},
  \bibinfo{person}{Michael~J Black}, {and} \bibinfo{person}{Timo Bolkart}.}
  \bibinfo{year}{2021}\natexlab{}.
\newblock \showarticletitle{Learning an animatable detailed 3D face model from
  in-the-wild images}.
\newblock \bibinfo{journal}{\emph{ACM Transactions on Graphics (TOG)}}
  \bibinfo{volume}{40}, \bibinfo{number}{4} (\bibinfo{year}{2021}),
  \bibinfo{pages}{1--13}.
\newblock


\bibitem[Feng et~al\mbox{.}(2018)]%
        {3dmm_nn_prnet}
\bibfield{author}{\bibinfo{person}{Yao Feng}, \bibinfo{person}{Fan Wu},
  \bibinfo{person}{Xiaohu Shao}, \bibinfo{person}{Yanfeng Wang}, {and}
  \bibinfo{person}{Xi Zhou}.} \bibinfo{year}{2018}\natexlab{}.
\newblock \showarticletitle{Joint 3d face reconstruction and dense alignment
  with position map regression network}. In
  \bibinfo{booktitle}{\emph{Proceedings of the European conference on computer
  vision (ECCV)}}. \bibinfo{pages}{534--551}.
\newblock


\bibitem[Flynn et~al\mbox{.}(2019)]%
        {MPI_deepview}
\bibfield{author}{\bibinfo{person}{John Flynn}, \bibinfo{person}{Michael
  Broxton}, \bibinfo{person}{Paul Debevec}, \bibinfo{person}{Matthew DuVall},
  \bibinfo{person}{Graham Fyffe}, \bibinfo{person}{Ryan Overbeck},
  \bibinfo{person}{Noah Snavely}, {and} \bibinfo{person}{Richard Tucker}.}
  \bibinfo{year}{2019}\natexlab{}.
\newblock \showarticletitle{Deepview: View synthesis with learned gradient
  descent}. In \bibinfo{booktitle}{\emph{Proceedings of the IEEE/CVF Conference
  on Computer Vision and Pattern Recognition}}. \bibinfo{pages}{2367--2376}.
\newblock


\bibitem[Fyffe et~al\mbox{.}(2017)]%
        {mesh_mvstopology}
\bibfield{author}{\bibinfo{person}{Graham Fyffe}, \bibinfo{person}{Koki
  Nagano}, \bibinfo{person}{Loc Huynh}, \bibinfo{person}{Shunsuke Saito},
  \bibinfo{person}{Jay Busch}, \bibinfo{person}{Andrew Jones},
  \bibinfo{person}{Hao Li}, {and} \bibinfo{person}{Paul Debevec}.}
  \bibinfo{year}{2017}\natexlab{}.
\newblock \showarticletitle{Multi-View Stereo on Consistent Face Topology}. In
  \bibinfo{booktitle}{\emph{Computer Graphics Forum}},
  Vol.~\bibinfo{volume}{36}. Wiley Online Library, \bibinfo{pages}{295--309}.
\newblock


\bibitem[Gafni et~al\mbox{.}(2021)]%
        {NeRFace_dynamic}
\bibfield{author}{\bibinfo{person}{Guy Gafni}, \bibinfo{person}{Justus Thies},
  \bibinfo{person}{Michael Zollhofer}, {and} \bibinfo{person}{Matthias
  Nie{\ss}ner}.} \bibinfo{year}{2021}\natexlab{}.
\newblock \showarticletitle{Dynamic neural radiance fields for monocular 4d
  facial avatar reconstruction}. In \bibinfo{booktitle}{\emph{Proceedings of
  the IEEE/CVF Conference on Computer Vision and Pattern Recognition}}.
  \bibinfo{pages}{8649--8658}.
\newblock


\bibitem[Gecer et~al\mbox{.}(2019)]%
        {3dmm_dr_ganfit}
\bibfield{author}{\bibinfo{person}{Baris Gecer}, \bibinfo{person}{Stylianos
  Ploumpis}, \bibinfo{person}{Irene Kotsia}, {and} \bibinfo{person}{Stefanos
  Zafeiriou}.} \bibinfo{year}{2019}\natexlab{}.
\newblock \showarticletitle{Ganfit: Generative adversarial network fitting for
  high fidelity 3d face reconstruction}. In
  \bibinfo{booktitle}{\emph{Proceedings of the IEEE/CVF conference on computer
  vision and pattern recognition}}. \bibinfo{pages}{1155--1164}.
\newblock


\bibitem[Genova et~al\mbox{.}(2018)]%
        {3dmm_dr_unsupervised}
\bibfield{author}{\bibinfo{person}{Kyle Genova}, \bibinfo{person}{Forrester
  Cole}, \bibinfo{person}{Aaron Maschinot}, \bibinfo{person}{Aaron Sarna},
  \bibinfo{person}{Daniel Vlasic}, {and} \bibinfo{person}{William~T Freeman}.}
  \bibinfo{year}{2018}\natexlab{}.
\newblock \showarticletitle{Unsupervised training for 3d morphable model
  regression}. In \bibinfo{booktitle}{\emph{Proceedings of the IEEE Conference
  on Computer Vision and Pattern Recognition}}. \bibinfo{pages}{8377--8386}.
\newblock


\bibitem[Gerig et~al\mbox{.}(2018)]%
        {3dmm_bfm2017}
\bibfield{author}{\bibinfo{person}{Thomas Gerig}, \bibinfo{person}{Andreas
  Morel-Forster}, \bibinfo{person}{Clemens Blumer}, \bibinfo{person}{Bernhard
  Egger}, \bibinfo{person}{Marcel Luthi}, \bibinfo{person}{Sandro
  Sch{\"o}nborn}, {and} \bibinfo{person}{Thomas Vetter}.}
  \bibinfo{year}{2018}\natexlab{}.
\newblock \showarticletitle{Morphable face models-an open framework}. In
  \bibinfo{booktitle}{\emph{2018 13th IEEE International Conference on
  Automatic Face \& Gesture Recognition (FG 2018)}}. IEEE,
  \bibinfo{pages}{75--82}.
\newblock


\bibitem[Gotardo et~al\mbox{.}(2018)]%
        {R2_practical}
\bibfield{author}{\bibinfo{person}{Paulo Gotardo},
  \bibinfo{person}{J{\'e}r{\'e}my Riviere}, \bibinfo{person}{Derek Bradley},
  \bibinfo{person}{Abhijeet Ghosh}, {and} \bibinfo{person}{Thabo Beeler}.}
  \bibinfo{year}{2018}\natexlab{}.
\newblock \showarticletitle{Practical dynamic facial appearance modeling and
  acquisition}.
\newblock  (\bibinfo{year}{2018}).
\newblock


\bibitem[Griwodz et~al\mbox{.}(2021)]%
        {mesh_sfm_meshroom}
\bibfield{author}{\bibinfo{person}{Carsten Griwodz}, \bibinfo{person}{Simone
  Gasparini}, \bibinfo{person}{Lilian Calvet}, \bibinfo{person}{Pierre
  Gurdjos}, \bibinfo{person}{Fabien Castan}, \bibinfo{person}{Benoit Maujean},
  \bibinfo{person}{Gregoire~De Lillo}, {and} \bibinfo{person}{Yann Lanthony}.}
  \bibinfo{year}{2021}\natexlab{}.
\newblock \showarticletitle{{A}liceVision {M}eshroom: An open-source {3D}
  reconstruction pipeline}. In \bibinfo{booktitle}{\emph{Proceedings of the
  12th ACM Multimedia Systems Conference - {MMSys '21}}}.
  \bibinfo{publisher}{ACM Press}.
\newblock
\urldef\tempurl%
\url{https://doi.org/10.1145/3458305.3478443}
\showDOI{\tempurl}


\bibitem[Gu et~al\mbox{.}(2021)]%
        {NeRF_GAN_stylenerf}
\bibfield{author}{\bibinfo{person}{Jiatao Gu}, \bibinfo{person}{Lingjie Liu},
  \bibinfo{person}{Peng Wang}, {and} \bibinfo{person}{Christian Theobalt}.}
  \bibinfo{year}{2021}\natexlab{}.
\newblock \showarticletitle{Stylenerf: A style-based 3d-aware generator for
  high-resolution image synthesis}.
\newblock \bibinfo{journal}{\emph{arXiv preprint arXiv:2110.08985}}
  (\bibinfo{year}{2021}).
\newblock


\bibitem[Guo et~al\mbox{.}(2020)]%
        {3dmm_direct_3DDFAv2}
\bibfield{author}{\bibinfo{person}{Jianzhu Guo}, \bibinfo{person}{Xiangyu Zhu},
  \bibinfo{person}{Yang Yang}, \bibinfo{person}{Fan Yang},
  \bibinfo{person}{Zhen Lei}, {and} \bibinfo{person}{Stan~Z Li}.}
  \bibinfo{year}{2020}\natexlab{}.
\newblock \showarticletitle{Towards fast, accurate and stable 3d dense face
  alignment}. In \bibinfo{booktitle}{\emph{European Conference on Computer
  Vision}}. Springer, \bibinfo{pages}{152--168}.
\newblock


\bibitem[Hong et~al\mbox{.}(2022)]%
        {NeRFace_headnerf}
\bibfield{author}{\bibinfo{person}{Yang Hong}, \bibinfo{person}{Bo Peng},
  \bibinfo{person}{Haiyao Xiao}, \bibinfo{person}{Ligang Liu}, {and}
  \bibinfo{person}{Juyong Zhang}.} \bibinfo{year}{2022}\natexlab{}.
\newblock \showarticletitle{HeadNeRF: A Real-time NeRF-based Parametric Head
  Model}.
\newblock  (\bibinfo{year}{2022}).
\newblock


\bibitem[Huang et~al\mbox{.}(2018)]%
        {mesh_sfm_deepmvs}
\bibfield{author}{\bibinfo{person}{Po-Han Huang}, \bibinfo{person}{Kevin
  Matzen}, \bibinfo{person}{Johannes Kopf}, \bibinfo{person}{Narendra Ahuja},
  {and} \bibinfo{person}{Jia-Bin Huang}.} \bibinfo{year}{2018}\natexlab{}.
\newblock \showarticletitle{DeepMVS: Learning Multi-View Stereopsis}. In
  \bibinfo{booktitle}{\emph{IEEE Conference on Computer Vision and Pattern
  Recognition (CVPR)}}.
\newblock


\bibitem[Kar et~al\mbox{.}(2017)]%
        {volume_sdf_mvsmachine}
\bibfield{author}{\bibinfo{person}{Abhishek Kar}, \bibinfo{person}{Christian
  H{\"a}ne}, {and} \bibinfo{person}{Jitendra Malik}.}
  \bibinfo{year}{2017}\natexlab{}.
\newblock \showarticletitle{Learning a multi-view stereo machine}.
\newblock \bibinfo{journal}{\emph{Advances in neural information processing
  systems}}  \bibinfo{volume}{30} (\bibinfo{year}{2017}).
\newblock


\bibitem[Kasten et~al\mbox{.}(2021)]%
        {misc_neuralatlas}
\bibfield{author}{\bibinfo{person}{Yoni Kasten}, \bibinfo{person}{Dolev Ofri},
  \bibinfo{person}{Oliver Wang}, {and} \bibinfo{person}{Tali Dekel}.}
  \bibinfo{year}{2021}\natexlab{}.
\newblock \showarticletitle{Layered neural atlases for consistent video
  editing}.
\newblock \bibinfo{journal}{\emph{ACM Transactions on Graphics (TOG)}}
  \bibinfo{volume}{40}, \bibinfo{number}{6} (\bibinfo{year}{2021}),
  \bibinfo{pages}{1--12}.
\newblock


\bibitem[Kato et~al\mbox{.}(2018)]%
        {mesh_dr1_NMR}
\bibfield{author}{\bibinfo{person}{Hiroharu Kato}, \bibinfo{person}{Yoshitaka
  Ushiku}, {and} \bibinfo{person}{Tatsuya Harada}.}
  \bibinfo{year}{2018}\natexlab{}.
\newblock \showarticletitle{Neural 3d mesh renderer}. In
  \bibinfo{booktitle}{\emph{Proceedings of the IEEE conference on computer
  vision and pattern recognition}}. \bibinfo{pages}{3907--3916}.
\newblock


\bibitem[Kazhdan et~al\mbox{.}(2006)]%
        {mesh_poisson}
\bibfield{author}{\bibinfo{person}{Michael Kazhdan}, \bibinfo{person}{Matthew
  Bolitho}, {and} \bibinfo{person}{Hugues Hoppe}.}
  \bibinfo{year}{2006}\natexlab{}.
\newblock \showarticletitle{Poisson surface reconstruction}. In
  \bibinfo{booktitle}{\emph{Proceedings of the fourth Eurographics symposium on
  Geometry processing}}, Vol.~\bibinfo{volume}{7}.
\newblock


\bibitem[Kingma and Ba(2014)]%
        {misc_adam}
\bibfield{author}{\bibinfo{person}{Diederik~P. Kingma} {and}
  \bibinfo{person}{Jimmy Ba}.} \bibinfo{year}{2014}\natexlab{}.
\newblock \bibinfo{title}{Adam: A Method for Stochastic Optimization}.
\newblock
\newblock
\showeprint[arxiv]{1412.6980}~[cs.LG]


\bibitem[Kolkin et~al\mbox{.}(2019)]%
        {misc_STROTSS}
\bibfield{author}{\bibinfo{person}{Nicholas Kolkin}, \bibinfo{person}{Jason
  Salavon}, {and} \bibinfo{person}{Gregory Shakhnarovich}.}
  \bibinfo{year}{2019}\natexlab{}.
\newblock \showarticletitle{Style transfer by relaxed optimal transport and
  self-similarity}. In \bibinfo{booktitle}{\emph{Proceedings of the IEEE/CVF
  Conference on Computer Vision and Pattern Recognition}}.
  \bibinfo{pages}{10051--10060}.
\newblock


\bibitem[Levoy(1990)]%
        {volume_render}
\bibfield{author}{\bibinfo{person}{Marc Levoy}.}
  \bibinfo{year}{1990}\natexlab{}.
\newblock \showarticletitle{Efficient ray tracing of volume data}.
\newblock \bibinfo{journal}{\emph{ACM Transactions on Graphics (TOG)}}
  \bibinfo{volume}{9}, \bibinfo{number}{3} (\bibinfo{year}{1990}),
  \bibinfo{pages}{245--261}.
\newblock


\bibitem[Li et~al\mbox{.}(2020)]%
        {mesh_dynamicasset}
\bibfield{author}{\bibinfo{person}{Jiaman Li}, \bibinfo{person}{Zhengfei
  Kuang}, \bibinfo{person}{Yajie Zhao}, \bibinfo{person}{Mingming He},
  \bibinfo{person}{Karl Bladin}, {and} \bibinfo{person}{Hao Li}.}
  \bibinfo{year}{2020}\natexlab{}.
\newblock \showarticletitle{Dynamic facial asset and rig generation from a
  single scan.}
\newblock \bibinfo{journal}{\emph{ACM Trans. Graph.}} \bibinfo{volume}{39},
  \bibinfo{number}{6} (\bibinfo{year}{2020}), \bibinfo{pages}{215--1}.
\newblock


\bibitem[Li et~al\mbox{.}(2021a)]%
        {NeRFace_deform}
\bibfield{author}{\bibinfo{person}{Moran Li}, \bibinfo{person}{Haibin Huang},
  \bibinfo{person}{Yi Zheng}, \bibinfo{person}{Mengtian Li},
  \bibinfo{person}{Nong Sang}, {and} \bibinfo{person}{Chongyang Ma}.}
  \bibinfo{year}{2021}\natexlab{a}.
\newblock \showarticletitle{Implicit Neural Deformation for Multi-View Face
  Reconstruction}.
\newblock \bibinfo{journal}{\emph{arXiv preprint arXiv:2112.02494}}
  (\bibinfo{year}{2021}).
\newblock


\bibitem[Li et~al\mbox{.}(2017)]%
        {3dmm_FLAME}
\bibfield{author}{\bibinfo{person}{Tianye Li}, \bibinfo{person}{Timo Bolkart},
  \bibinfo{person}{Michael~J Black}, \bibinfo{person}{Hao Li}, {and}
  \bibinfo{person}{Javier Romero}.} \bibinfo{year}{2017}\natexlab{}.
\newblock \showarticletitle{Learning a model of facial shape and expression
  from 4D scans.}
\newblock \bibinfo{journal}{\emph{ACM Trans. Graph.}} \bibinfo{volume}{36},
  \bibinfo{number}{6} (\bibinfo{year}{2017}), \bibinfo{pages}{194--1}.
\newblock


\bibitem[Li et~al\mbox{.}(2021b)]%
        {3dmm_nn_tofu}
\bibfield{author}{\bibinfo{person}{Tianye Li}, \bibinfo{person}{Shichen Liu},
  \bibinfo{person}{Timo Bolkart}, \bibinfo{person}{Jiayi Liu},
  \bibinfo{person}{Hao Li}, {and} \bibinfo{person}{Yajie Zhao}.}
  \bibinfo{year}{2021}\natexlab{b}.
\newblock \showarticletitle{Topologically Consistent Multi-View Face Inference
  Using Volumetric Sampling}. In \bibinfo{booktitle}{\emph{Proceedings of the
  IEEE/CVF International Conference on Computer Vision}}.
  \bibinfo{pages}{3824--3834}.
\newblock


\bibitem[Li et~al\mbox{.}(2021c)]%
        {NeRF_dyn_3Dvideo}
\bibfield{author}{\bibinfo{person}{Tianye Li}, \bibinfo{person}{Mira
  Slavcheva}, \bibinfo{person}{Michael Zollhoefer}, \bibinfo{person}{Simon
  Green}, \bibinfo{person}{Christoph Lassner}, \bibinfo{person}{Changil Kim},
  \bibinfo{person}{Tanner Schmidt}, \bibinfo{person}{Steven Lovegrove},
  \bibinfo{person}{Michael Goesele}, {and} \bibinfo{person}{Zhaoyang Lv}.}
  \bibinfo{year}{2021}\natexlab{c}.
\newblock \showarticletitle{Neural 3d video synthesis}.
\newblock \bibinfo{journal}{\emph{arXiv preprint arXiv:2103.02597}}
  (\bibinfo{year}{2021}).
\newblock


\bibitem[Li et~al\mbox{.}(2018)]%
        {mesh_dr2_MCraytracing}
\bibfield{author}{\bibinfo{person}{Tzu-Mao Li}, \bibinfo{person}{Miika
  Aittala}, \bibinfo{person}{Fr{\'e}do Durand}, {and} \bibinfo{person}{Jaakko
  Lehtinen}.} \bibinfo{year}{2018}\natexlab{}.
\newblock \showarticletitle{Differentiable monte carlo ray tracing through edge
  sampling}.
\newblock \bibinfo{journal}{\emph{ACM Transactions on Graphics (TOG)}}
  \bibinfo{volume}{37}, \bibinfo{number}{6} (\bibinfo{year}{2018}),
  \bibinfo{pages}{1--11}.
\newblock


\bibitem[Lin et~al\mbox{.}(2022)]%
        {misc_RVM}
\bibfield{author}{\bibinfo{person}{Shanchuan Lin}, \bibinfo{person}{Linjie
  Yang}, \bibinfo{person}{Imran Saleemi}, {and} \bibinfo{person}{Soumyadip
  Sengupta}.} \bibinfo{year}{2022}\natexlab{}.
\newblock \showarticletitle{Robust High-Resolution Video Matting with Temporal
  Guidance}. In \bibinfo{booktitle}{\emph{Proceedings of the IEEE/CVF Winter
  Conference on Applications of Computer Vision}}. \bibinfo{pages}{238--247}.
\newblock


\bibitem[Liu et~al\mbox{.}(2019)]%
        {mesh_dr1_softrasterizer}
\bibfield{author}{\bibinfo{person}{Shichen Liu}, \bibinfo{person}{Tianye Li},
  \bibinfo{person}{Weikai Chen}, {and} \bibinfo{person}{Hao Li}.}
  \bibinfo{year}{2019}\natexlab{}.
\newblock \showarticletitle{Soft rasterizer: A differentiable renderer for
  image-based 3d reasoning}. In \bibinfo{booktitle}{\emph{Proceedings of the
  IEEE/CVF International Conference on Computer Vision}}.
  \bibinfo{pages}{7708--7717}.
\newblock


\bibitem[Liu et~al\mbox{.}(2021)]%
        {NeRF_edit_cond}
\bibfield{author}{\bibinfo{person}{Steven Liu}, \bibinfo{person}{Xiuming
  Zhang}, \bibinfo{person}{Zhoutong Zhang}, \bibinfo{person}{Richard Zhang},
  \bibinfo{person}{Jun-Yan Zhu}, {and} \bibinfo{person}{Bryan Russell}.}
  \bibinfo{year}{2021}\natexlab{}.
\newblock \showarticletitle{Editing conditional radiance fields}. In
  \bibinfo{booktitle}{\emph{Proceedings of the IEEE/CVF International
  Conference on Computer Vision}}. \bibinfo{pages}{5773--5783}.
\newblock


\bibitem[Lombardi et~al\mbox{.}(2018)]%
        {mesh_appear_DAM}
\bibfield{author}{\bibinfo{person}{Stephen Lombardi}, \bibinfo{person}{Jason
  Saragih}, \bibinfo{person}{Tomas Simon}, {and} \bibinfo{person}{Yaser
  Sheikh}.} \bibinfo{year}{2018}\natexlab{}.
\newblock \showarticletitle{Deep appearance models for face rendering}.
\newblock \bibinfo{journal}{\emph{ACM Transactions on Graphics (ToG)}}
  \bibinfo{volume}{37}, \bibinfo{number}{4} (\bibinfo{year}{2018}),
  \bibinfo{pages}{1--13}.
\newblock


\bibitem[Lombardi et~al\mbox{.}(2019)]%
        {volume_NV}
\bibfield{author}{\bibinfo{person}{Stephen Lombardi}, \bibinfo{person}{Tomas
  Simon}, \bibinfo{person}{Jason Saragih}, \bibinfo{person}{Gabriel Schwartz},
  \bibinfo{person}{Andreas Lehrmann}, {and} \bibinfo{person}{Yaser Sheikh}.}
  \bibinfo{year}{2019}\natexlab{}.
\newblock \showarticletitle{Neural volumes: Learning dynamic renderable volumes
  from images}.
\newblock \bibinfo{journal}{\emph{arXiv preprint arXiv:1906.07751}}
  (\bibinfo{year}{2019}).
\newblock


\bibitem[Lombardi et~al\mbox{.}(2021)]%
        {volume_MVP}
\bibfield{author}{\bibinfo{person}{Stephen Lombardi}, \bibinfo{person}{Tomas
  Simon}, \bibinfo{person}{Gabriel Schwartz}, \bibinfo{person}{Michael
  Zollhoefer}, \bibinfo{person}{Yaser Sheikh}, {and} \bibinfo{person}{Jason
  Saragih}.} \bibinfo{year}{2021}\natexlab{}.
\newblock \showarticletitle{Mixture of volumetric primitives for efficient
  neural rendering}.
\newblock \bibinfo{journal}{\emph{ACM Transactions on Graphics (TOG)}}
  \bibinfo{volume}{40}, \bibinfo{number}{4} (\bibinfo{year}{2021}),
  \bibinfo{pages}{1--13}.
\newblock


\bibitem[Loper and Black(2014)]%
        {mesh_dr1_opendr}
\bibfield{author}{\bibinfo{person}{Matthew~M Loper} {and}
  \bibinfo{person}{Michael~J Black}.} \bibinfo{year}{2014}\natexlab{}.
\newblock \showarticletitle{OpenDR: An approximate differentiable renderer}. In
  \bibinfo{booktitle}{\emph{European Conference on Computer Vision}}. Springer,
  \bibinfo{pages}{154--169}.
\newblock


\bibitem[Loubet et~al\mbox{.}(2019)]%
        {mesh_dr2_repearameterizing}
\bibfield{author}{\bibinfo{person}{Guillaume Loubet}, \bibinfo{person}{Nicolas
  Holzschuch}, {and} \bibinfo{person}{Wenzel Jakob}.}
  \bibinfo{year}{2019}\natexlab{}.
\newblock \showarticletitle{Reparameterizing discontinuous integrands for
  differentiable rendering}.
\newblock \bibinfo{journal}{\emph{ACM Transactions on Graphics (TOG)}}
  \bibinfo{volume}{38}, \bibinfo{number}{6} (\bibinfo{year}{2019}),
  \bibinfo{pages}{1--14}.
\newblock


\bibitem[Luan et~al\mbox{.}(2021)]%
        {mesh_dr2_svbrdf}
\bibfield{author}{\bibinfo{person}{Fujun Luan}, \bibinfo{person}{Shuang Zhao},
  \bibinfo{person}{Kavita Bala}, {and} \bibinfo{person}{Zhao Dong}.}
  \bibinfo{year}{2021}\natexlab{}.
\newblock \showarticletitle{Unified shape and svbrdf recovery using
  differentiable monte carlo rendering}. In \bibinfo{booktitle}{\emph{Computer
  Graphics Forum}}, Vol.~\bibinfo{volume}{40}. Wiley Online Library,
  \bibinfo{pages}{101--113}.
\newblock


\bibitem[Luo et~al\mbox{.}(2019)]%
        {mesh_sfm_pmvsnet}
\bibfield{author}{\bibinfo{person}{Keyang Luo}, \bibinfo{person}{Tao Guan},
  \bibinfo{person}{Lili Ju}, \bibinfo{person}{Haipeng Huang}, {and}
  \bibinfo{person}{Yawei Luo}.} \bibinfo{year}{2019}\natexlab{}.
\newblock \showarticletitle{P-mvsnet: Learning patch-wise matching confidence
  aggregation for multi-view stereo}. In \bibinfo{booktitle}{\emph{Proceedings
  of the IEEE/CVF International Conference on Computer Vision}}.
  \bibinfo{pages}{10452--10461}.
\newblock


\bibitem[Ma et~al\mbox{.}(2021)]%
        {3dmm_nn_pixelcodec}
\bibfield{author}{\bibinfo{person}{Shugao Ma}, \bibinfo{person}{Tomas Simon},
  \bibinfo{person}{Jason Saragih}, \bibinfo{person}{Dawei Wang},
  \bibinfo{person}{Yuecheng Li}, \bibinfo{person}{Fernando De~La~Torre}, {and}
  \bibinfo{person}{Yaser Sheikh}.} \bibinfo{year}{2021}\natexlab{}.
\newblock \showarticletitle{Pixel codec avatars}. In
  \bibinfo{booktitle}{\emph{Proceedings of the IEEE/CVF Conference on Computer
  Vision and Pattern Recognition}}. \bibinfo{pages}{64--73}.
\newblock


\bibitem[Mescheder et~al\mbox{.}(2019)]%
        {occupancynetwork}
\bibfield{author}{\bibinfo{person}{Lars Mescheder}, \bibinfo{person}{Michael
  Oechsle}, \bibinfo{person}{Michael Niemeyer}, \bibinfo{person}{Sebastian
  Nowozin}, {and} \bibinfo{person}{Andreas Geiger}.}
  \bibinfo{year}{2019}\natexlab{}.
\newblock \showarticletitle{Occupancy networks: Learning 3d reconstruction in
  function space}. In \bibinfo{booktitle}{\emph{Proceedings of the IEEE/CVF
  Conference on Computer Vision and Pattern Recognition}}.
  \bibinfo{pages}{4460--4470}.
\newblock


\bibitem[Mildenhall et~al\mbox{.}(2019)]%
        {MPI_LLFF}
\bibfield{author}{\bibinfo{person}{Ben Mildenhall}, \bibinfo{person}{Pratul~P
  Srinivasan}, \bibinfo{person}{Rodrigo Ortiz-Cayon},
  \bibinfo{person}{Nima~Khademi Kalantari}, \bibinfo{person}{Ravi Ramamoorthi},
  \bibinfo{person}{Ren Ng}, {and} \bibinfo{person}{Abhishek Kar}.}
  \bibinfo{year}{2019}\natexlab{}.
\newblock \showarticletitle{Local light field fusion: Practical view synthesis
  with prescriptive sampling guidelines}.
\newblock \bibinfo{journal}{\emph{ACM Transactions on Graphics (TOG)}}
  \bibinfo{volume}{38}, \bibinfo{number}{4} (\bibinfo{year}{2019}),
  \bibinfo{pages}{1--14}.
\newblock


\bibitem[Mildenhall et~al\mbox{.}(2020)]%
        {NeRF}
\bibfield{author}{\bibinfo{person}{Ben Mildenhall}, \bibinfo{person}{Pratul~P
  Srinivasan}, \bibinfo{person}{Matthew Tancik}, \bibinfo{person}{Jonathan~T
  Barron}, \bibinfo{person}{Ravi Ramamoorthi}, {and} \bibinfo{person}{Ren Ng}.}
  \bibinfo{year}{2020}\natexlab{}.
\newblock \showarticletitle{Nerf: Representing scenes as neural radiance fields
  for view synthesis}. In \bibinfo{booktitle}{\emph{European conference on
  computer vision}}. Springer, \bibinfo{pages}{405--421}.
\newblock


\bibitem[M{\"u}ller et~al\mbox{.}(2022)]%
        {misc_instantNGP}
\bibfield{author}{\bibinfo{person}{Thomas M{\"u}ller}, \bibinfo{person}{Alex
  Evans}, \bibinfo{person}{Christoph Schied}, {and} \bibinfo{person}{Alexander
  Keller}.} \bibinfo{year}{2022}\natexlab{}.
\newblock \showarticletitle{Instant Neural Graphics Primitives with a
  Multiresolution Hash Encoding}.
\newblock \bibinfo{journal}{\emph{arXiv preprint arXiv:2201.05989}}
  (\bibinfo{year}{2022}).
\newblock


\bibitem[Niemeyer and Geiger(2021)]%
        {NeRF_GAN_giraffe}
\bibfield{author}{\bibinfo{person}{Michael Niemeyer} {and}
  \bibinfo{person}{Andreas Geiger}.} \bibinfo{year}{2021}\natexlab{}.
\newblock \showarticletitle{Giraffe: Representing scenes as compositional
  generative neural feature fields}. In \bibinfo{booktitle}{\emph{Proceedings
  of the IEEE/CVF Conference on Computer Vision and Pattern Recognition}}.
  \bibinfo{pages}{11453--11464}.
\newblock


\bibitem[Niemeyer et~al\mbox{.}(2020)]%
        {sdf_DVR}
\bibfield{author}{\bibinfo{person}{Michael Niemeyer}, \bibinfo{person}{Lars
  Mescheder}, \bibinfo{person}{Michael Oechsle}, {and} \bibinfo{person}{Andreas
  Geiger}.} \bibinfo{year}{2020}\natexlab{}.
\newblock \showarticletitle{Differentiable volumetric rendering: Learning
  implicit 3d representations without 3d supervision}. In
  \bibinfo{booktitle}{\emph{Proceedings of the IEEE/CVF Conference on Computer
  Vision and Pattern Recognition}}. \bibinfo{pages}{3504--3515}.
\newblock


\bibitem[Park et~al\mbox{.}(2021a)]%
        {NeRF_dyn_nerfies}
\bibfield{author}{\bibinfo{person}{Keunhong Park}, \bibinfo{person}{Utkarsh
  Sinha}, \bibinfo{person}{Jonathan~T Barron}, \bibinfo{person}{Sofien
  Bouaziz}, \bibinfo{person}{Dan~B Goldman}, \bibinfo{person}{Steven~M Seitz},
  {and} \bibinfo{person}{Ricardo Martin-Brualla}.}
  \bibinfo{year}{2021}\natexlab{a}.
\newblock \showarticletitle{Nerfies: Deformable neural radiance fields}. In
  \bibinfo{booktitle}{\emph{Proceedings of the IEEE/CVF International
  Conference on Computer Vision}}. \bibinfo{pages}{5865--5874}.
\newblock


\bibitem[Park et~al\mbox{.}(2021b)]%
        {NeRF_dyn_hypernerf}
\bibfield{author}{\bibinfo{person}{Keunhong Park}, \bibinfo{person}{Utkarsh
  Sinha}, \bibinfo{person}{Peter Hedman}, \bibinfo{person}{Jonathan~T Barron},
  \bibinfo{person}{Sofien Bouaziz}, \bibinfo{person}{Dan~B Goldman},
  \bibinfo{person}{Ricardo Martin-Brualla}, {and} \bibinfo{person}{Steven~M
  Seitz}.} \bibinfo{year}{2021}\natexlab{b}.
\newblock \showarticletitle{HyperNeRF: A Higher-Dimensional Representation for
  Topologically Varying Neural Radiance Fields}.
\newblock \bibinfo{journal}{\emph{arXiv preprint arXiv:2106.13228}}
  (\bibinfo{year}{2021}).
\newblock


\bibitem[Paysan et~al\mbox{.}(2009)]%
        {3dmm_bfm}
\bibfield{author}{\bibinfo{person}{Pascal Paysan}, \bibinfo{person}{Reinhard
  Knothe}, \bibinfo{person}{Brian Amberg}, \bibinfo{person}{Sami Romdhani},
  {and} \bibinfo{person}{Thomas Vetter}.} \bibinfo{year}{2009}\natexlab{}.
\newblock \showarticletitle{A 3D face model for pose and illumination invariant
  face recognition}. In \bibinfo{booktitle}{\emph{2009 sixth IEEE international
  conference on advanced video and signal based surveillance}}. Ieee,
  \bibinfo{pages}{296--301}.
\newblock


\bibitem[Pumarola et~al\mbox{.}(2021)]%
        {NeRF_dyn_dnerf}
\bibfield{author}{\bibinfo{person}{Albert Pumarola}, \bibinfo{person}{Enric
  Corona}, \bibinfo{person}{Gerard Pons-Moll}, {and} \bibinfo{person}{Francesc
  Moreno-Noguer}.} \bibinfo{year}{2021}\natexlab{}.
\newblock \showarticletitle{D-nerf: Neural radiance fields for dynamic scenes}.
  In \bibinfo{booktitle}{\emph{Proceedings of the IEEE/CVF Conference on
  Computer Vision and Pattern Recognition}}. \bibinfo{pages}{10318--10327}.
\newblock


\bibitem[Ramon et~al\mbox{.}(2021)]%
        {NeRFace_H3D}
\bibfield{author}{\bibinfo{person}{Eduard Ramon}, \bibinfo{person}{Gil
  Triginer}, \bibinfo{person}{Janna Escur}, \bibinfo{person}{Albert Pumarola},
  \bibinfo{person}{Jaime Garcia}, \bibinfo{person}{Xavier Giro-i Nieto}, {and}
  \bibinfo{person}{Francesc Moreno-Noguer}.} \bibinfo{year}{2021}\natexlab{}.
\newblock \showarticletitle{H3d-net: Few-shot high-fidelity 3d head
  reconstruction}. In \bibinfo{booktitle}{\emph{Proceedings of the IEEE/CVF
  International Conference on Computer Vision}}. \bibinfo{pages}{5620--5629}.
\newblock


\bibitem[Ravi et~al\mbox{.}(2020)]%
        {mesh_dr1_pytorch3d}
\bibfield{author}{\bibinfo{person}{Nikhila Ravi}, \bibinfo{person}{Jeremy
  Reizenstein}, \bibinfo{person}{David Novotny}, \bibinfo{person}{Taylor
  Gordon}, \bibinfo{person}{Wan-Yen Lo}, \bibinfo{person}{Justin Johnson},
  {and} \bibinfo{person}{Georgia Gkioxari}.} \bibinfo{year}{2020}\natexlab{}.
\newblock \showarticletitle{Accelerating 3d deep learning with pytorch3d}.
\newblock \bibinfo{journal}{\emph{arXiv preprint arXiv:2007.08501}}
  (\bibinfo{year}{2020}).
\newblock


\bibitem[Riegler and Koltun(2020a)]%
        {mesh_appear_freevs}
\bibfield{author}{\bibinfo{person}{Gernot Riegler} {and}
  \bibinfo{person}{Vladlen Koltun}.} \bibinfo{year}{2020}\natexlab{a}.
\newblock \showarticletitle{Free View Synthesis}.
\newblock \bibinfo{journal}{\emph{CoRR}}  \bibinfo{volume}{abs/2008.05511}
  (\bibinfo{year}{2020}).
\newblock
\showeprint[arXiv]{2008.05511}
\urldef\tempurl%
\url{https://arxiv.org/abs/2008.05511}
\showURL{%
\tempurl}


\bibitem[Riegler and Koltun(2020b)]%
        {mesh_appear_stablevs}
\bibfield{author}{\bibinfo{person}{Gernot Riegler} {and}
  \bibinfo{person}{Vladlen Koltun}.} \bibinfo{year}{2020}\natexlab{b}.
\newblock \showarticletitle{Stable View Synthesis}.
\newblock \bibinfo{journal}{\emph{CoRR}}  \bibinfo{volume}{abs/2011.07233}
  (\bibinfo{year}{2020}).
\newblock
\showeprint[arXiv]{2011.07233}
\urldef\tempurl%
\url{https://arxiv.org/abs/2011.07233}
\showURL{%
\tempurl}


\bibitem[Sch\"{o}nberger and Frahm(2016)]%
        {mesh_sfm_colmap1}
\bibfield{author}{\bibinfo{person}{Johannes~Lutz Sch\"{o}nberger} {and}
  \bibinfo{person}{Jan-Michael Frahm}.} \bibinfo{year}{2016}\natexlab{}.
\newblock \showarticletitle{Structure-from-Motion Revisited}. In
  \bibinfo{booktitle}{\emph{Conference on Computer Vision and Pattern
  Recognition (CVPR)}}.
\newblock


\bibitem[Sch\"{o}nberger et~al\mbox{.}(2016)]%
        {mesh_sfm_colmap2}
\bibfield{author}{\bibinfo{person}{Johannes~Lutz Sch\"{o}nberger},
  \bibinfo{person}{Enliang Zheng}, \bibinfo{person}{Marc Pollefeys}, {and}
  \bibinfo{person}{Jan-Michael Frahm}.} \bibinfo{year}{2016}\natexlab{}.
\newblock \showarticletitle{Pixelwise View Selection for Unstructured
  Multi-View Stereo}. In \bibinfo{booktitle}{\emph{European Conference on
  Computer Vision (ECCV)}}.
\newblock


\bibitem[Schwarz et~al\mbox{.}(2020)]%
        {NeRF_GAN_graf}
\bibfield{author}{\bibinfo{person}{Katja Schwarz}, \bibinfo{person}{Yiyi Liao},
  \bibinfo{person}{Michael Niemeyer}, {and} \bibinfo{person}{Andreas Geiger}.}
  \bibinfo{year}{2020}\natexlab{}.
\newblock \showarticletitle{Graf: Generative radiance fields for 3d-aware image
  synthesis}.
\newblock \bibinfo{journal}{\emph{arXiv preprint arXiv:2007.02442}}
  (\bibinfo{year}{2020}).
\newblock


\bibitem[Sorkine and Alexa(2007)]%
        {mesh_editing_asap}
\bibfield{author}{\bibinfo{person}{Olga Sorkine} {and} \bibinfo{person}{Marc
  Alexa}.} \bibinfo{year}{2007}\natexlab{}.
\newblock \showarticletitle{As-rigid-as-possible surface modeling}. In
  \bibinfo{booktitle}{\emph{Symposium on Geometry processing}},
  Vol.~\bibinfo{volume}{4}. \bibinfo{pages}{109--116}.
\newblock


\bibitem[Sorkine et~al\mbox{.}(2004)]%
        {mesh_editing_laplacian}
\bibfield{author}{\bibinfo{person}{Olga Sorkine}, \bibinfo{person}{Daniel
  Cohen-Or}, \bibinfo{person}{Yaron Lipman}, \bibinfo{person}{Marc Alexa},
  \bibinfo{person}{Christian R{\"o}ssl}, {and} \bibinfo{person}{H-P Seidel}.}
  \bibinfo{year}{2004}\natexlab{}.
\newblock \showarticletitle{Laplacian surface editing}. In
  \bibinfo{booktitle}{\emph{Proceedings of the 2004 Eurographics/ACM SIGGRAPH
  symposium on Geometry processing}}. \bibinfo{pages}{175--184}.
\newblock


\bibitem[Srinivasan et~al\mbox{.}(2021)]%
        {nerf_relit_nerv}
\bibfield{author}{\bibinfo{person}{Pratul~P Srinivasan},
  \bibinfo{person}{Boyang Deng}, \bibinfo{person}{Xiuming Zhang},
  \bibinfo{person}{Matthew Tancik}, \bibinfo{person}{Ben Mildenhall}, {and}
  \bibinfo{person}{Jonathan~T Barron}.} \bibinfo{year}{2021}\natexlab{}.
\newblock \showarticletitle{Nerv: Neural reflectance and visibility fields for
  relighting and view synthesis}. In \bibinfo{booktitle}{\emph{Proceedings of
  the IEEE/CVF Conference on Computer Vision and Pattern Recognition}}.
  \bibinfo{pages}{7495--7504}.
\newblock


\bibitem[Srinivasan et~al\mbox{.}(2019)]%
        {MPI_pushbound}
\bibfield{author}{\bibinfo{person}{Pratul~P Srinivasan},
  \bibinfo{person}{Richard Tucker}, \bibinfo{person}{Jonathan~T Barron},
  \bibinfo{person}{Ravi Ramamoorthi}, \bibinfo{person}{Ren Ng}, {and}
  \bibinfo{person}{Noah Snavely}.} \bibinfo{year}{2019}\natexlab{}.
\newblock \showarticletitle{Pushing the boundaries of view extrapolation with
  multiplane images}. In \bibinfo{booktitle}{\emph{Proceedings of the IEEE/CVF
  Conference on Computer Vision and Pattern Recognition}}.
  \bibinfo{pages}{175--184}.
\newblock


\bibitem[Sun et~al\mbox{.}(2022)]%
        {NeRF_IDE3D}
\bibfield{author}{\bibinfo{person}{Jingxiang Sun}, \bibinfo{person}{Xuan Wang},
  \bibinfo{person}{Yichun Shi}, \bibinfo{person}{Lizhen Wang},
  \bibinfo{person}{Jue Wang}, {and} \bibinfo{person}{Yebin Liu}.}
  \bibinfo{year}{2022}\natexlab{}.
\newblock \showarticletitle{IDE-3D: Interactive Disentangled Editing for
  High-Resolution 3D-aware Portrait Synthesis}.
\newblock \bibinfo{journal}{\emph{arXiv preprint arXiv:2205.15517}}
  (\bibinfo{year}{2022}).
\newblock


\bibitem[Sun et~al\mbox{.}(2021)]%
        {NeRF_edit_FENeRF}
\bibfield{author}{\bibinfo{person}{Jingxiang Sun}, \bibinfo{person}{Xuan Wang},
  \bibinfo{person}{Yong Zhang}, \bibinfo{person}{Xiaoyu Li},
  \bibinfo{person}{Qi Zhang}, \bibinfo{person}{Yebin Liu}, {and}
  \bibinfo{person}{Jue Wang}.} \bibinfo{year}{2021}\natexlab{}.
\newblock \showarticletitle{FENeRF: Face Editing in Neural Radiance Fields}.
\newblock \bibinfo{journal}{\emph{arXiv preprint arXiv:2111.15490}}
  (\bibinfo{year}{2021}).
\newblock


\bibitem[Tancik et~al\mbox{.}(2020)]%
        {fourier_embed}
\bibfield{author}{\bibinfo{person}{Matthew Tancik}, \bibinfo{person}{Pratul
  Srinivasan}, \bibinfo{person}{Ben Mildenhall}, \bibinfo{person}{Sara
  Fridovich-Keil}, \bibinfo{person}{Nithin Raghavan}, \bibinfo{person}{Utkarsh
  Singhal}, \bibinfo{person}{Ravi Ramamoorthi}, \bibinfo{person}{Jonathan
  Barron}, {and} \bibinfo{person}{Ren Ng}.} \bibinfo{year}{2020}\natexlab{}.
\newblock \showarticletitle{Fourier features let networks learn high frequency
  functions in low dimensional domains}.
\newblock \bibinfo{journal}{\emph{Advances in Neural Information Processing
  Systems}}  \bibinfo{volume}{33} (\bibinfo{year}{2020}),
  \bibinfo{pages}{7537--7547}.
\newblock


\bibitem[Thies et~al\mbox{.}(2019)]%
        {mesh_appear_deferred}
\bibfield{author}{\bibinfo{person}{Justus Thies}, \bibinfo{person}{Michael
  Zollh{\"o}fer}, {and} \bibinfo{person}{Matthias Nie{\ss}ner}.}
  \bibinfo{year}{2019}\natexlab{}.
\newblock \showarticletitle{Deferred neural rendering: Image synthesis using
  neural textures}.
\newblock \bibinfo{journal}{\emph{ACM Transactions on Graphics (TOG)}}
  \bibinfo{volume}{38}, \bibinfo{number}{4} (\bibinfo{year}{2019}),
  \bibinfo{pages}{1--12}.
\newblock


\bibitem[Wang et~al\mbox{.}(2021a)]%
        {NeRF_edit_clip}
\bibfield{author}{\bibinfo{person}{Can Wang}, \bibinfo{person}{Menglei Chai},
  \bibinfo{person}{Mingming He}, \bibinfo{person}{Dongdong Chen}, {and}
  \bibinfo{person}{Jing Liao}.} \bibinfo{year}{2021}\natexlab{a}.
\newblock \showarticletitle{CLIP-NeRF: Text-and-Image Driven Manipulation of
  Neural Radiance Fields}.
\newblock \bibinfo{journal}{\emph{arXiv preprint arXiv:2112.05139}}
  (\bibinfo{year}{2021}).
\newblock


\bibitem[Wang et~al\mbox{.}(2022)]%
        {3dmm_dr_faceverse}
\bibfield{author}{\bibinfo{person}{Lizhen Wang}, \bibinfo{person}{Zhiyuan
  Chen}, \bibinfo{person}{Tao Yu}, \bibinfo{person}{Chenguang Ma},
  \bibinfo{person}{Liang Li}, {and} \bibinfo{person}{Yebin Liu}.}
  \bibinfo{year}{2022}\natexlab{}.
\newblock \showarticletitle{FaceVerse: a Fine-grained and Detail-controllable
  3D Face Morphable Model from a Hybrid Dataset}.
\newblock \bibinfo{journal}{\emph{arXiv preprint arXiv:2203.14057}}
  (\bibinfo{year}{2022}).
\newblock


\bibitem[Wang et~al\mbox{.}(2021b)]%
        {sdf_NeuS}
\bibfield{author}{\bibinfo{person}{Peng Wang}, \bibinfo{person}{Lingjie Liu},
  \bibinfo{person}{Yuan Liu}, \bibinfo{person}{Christian Theobalt},
  \bibinfo{person}{Taku Komura}, {and} \bibinfo{person}{Wenping Wang}.}
  \bibinfo{year}{2021}\natexlab{b}.
\newblock \showarticletitle{Neus: Learning neural implicit surfaces by volume
  rendering for multi-view reconstruction}.
\newblock \bibinfo{journal}{\emph{arXiv preprint arXiv:2106.10689}}
  (\bibinfo{year}{2021}).
\newblock


\bibitem[Wei et~al\mbox{.}(2020)]%
        {mesh_sfm_deepsfm}
\bibfield{author}{\bibinfo{person}{Xingkui Wei}, \bibinfo{person}{Yinda Zhang},
  \bibinfo{person}{Zhuwen Li}, \bibinfo{person}{Yanwei Fu}, {and}
  \bibinfo{person}{Xiangyang Xue}.} \bibinfo{year}{2020}\natexlab{}.
\newblock \showarticletitle{Deepsfm: Structure from motion via deep bundle
  adjustment}. In \bibinfo{booktitle}{\emph{European conference on computer
  vision}}. Springer, \bibinfo{pages}{230--247}.
\newblock


\bibitem[Xia et~al\mbox{.}(2021)]%
        {2D_GANinversion}
\bibfield{author}{\bibinfo{person}{Weihao Xia}, \bibinfo{person}{Yulun Zhang},
  \bibinfo{person}{Yujiu Yang}, \bibinfo{person}{Jing-Hao Xue},
  \bibinfo{person}{Bolei Zhou}, {and} \bibinfo{person}{Ming-Hsuan Yang}.}
  \bibinfo{year}{2021}\natexlab{}.
\newblock \bibinfo{title}{GAN Inversion: A Survey}.
\newblock
\newblock
\showeprint[arxiv]{2101.05278}~[cs.CV]


\bibitem[Xian et~al\mbox{.}(2021)]%
        {xian2021space}
\bibfield{author}{\bibinfo{person}{Wenqi Xian}, \bibinfo{person}{Jia-Bin
  Huang}, \bibinfo{person}{Johannes Kopf}, {and} \bibinfo{person}{Changil
  Kim}.} \bibinfo{year}{2021}\natexlab{}.
\newblock \showarticletitle{Space-time neural irradiance fields for
  free-viewpoint video}. In \bibinfo{booktitle}{\emph{Proceedings of the
  IEEE/CVF Conference on Computer Vision and Pattern Recognition}}.
  \bibinfo{pages}{9421--9431}.
\newblock


\bibitem[Xiang et~al\mbox{.}(2021)]%
        {NeRF_neutex}
\bibfield{author}{\bibinfo{person}{Fanbo Xiang}, \bibinfo{person}{Zexiang Xu},
  \bibinfo{person}{Milos Hasan}, \bibinfo{person}{Yannick Hold-Geoffroy},
  \bibinfo{person}{Kalyan Sunkavalli}, {and} \bibinfo{person}{Hao Su}.}
  \bibinfo{year}{2021}\natexlab{}.
\newblock \showarticletitle{Neutex: Neural texture mapping for volumetric
  neural rendering}. In \bibinfo{booktitle}{\emph{Proceedings of the IEEE/CVF
  Conference on Computer Vision and Pattern Recognition}}.
  \bibinfo{pages}{7119--7128}.
\newblock


\bibitem[Yang et~al\mbox{.}(2020a)]%
        {3dmm_dr_facescape}
\bibfield{author}{\bibinfo{person}{Haotian Yang}, \bibinfo{person}{Hao Zhu},
  \bibinfo{person}{Yanru Wang}, \bibinfo{person}{Mingkai Huang},
  \bibinfo{person}{Qiu Shen}, \bibinfo{person}{Ruigang Yang}, {and}
  \bibinfo{person}{Xun Cao}.} \bibinfo{year}{2020}\natexlab{a}.
\newblock \showarticletitle{Facescape: a large-scale high quality 3d face
  dataset and detailed riggable 3d face prediction}. In
  \bibinfo{booktitle}{\emph{Proceedings of the IEEE/CVF Conference on Computer
  Vision and Pattern Recognition}}. \bibinfo{pages}{601--610}.
\newblock


\bibitem[Yang et~al\mbox{.}(2020b)]%
        {facescape}
\bibfield{author}{\bibinfo{person}{Haotian Yang}, \bibinfo{person}{Hao Zhu},
  \bibinfo{person}{Yanru Wang}, \bibinfo{person}{Mingkai Huang},
  \bibinfo{person}{Qiu Shen}, \bibinfo{person}{Ruigang Yang}, {and}
  \bibinfo{person}{Xun Cao}.} \bibinfo{year}{2020}\natexlab{b}.
\newblock \showarticletitle{FaceScape: a Large-scale High Quality 3D Face
  Dataset and Detailed Riggable 3D Face Prediction}. In
  \bibinfo{booktitle}{\emph{Proceedings of the IEEE Conference on Computer
  Vision and Pattern Recognition (CVPR)}}.
\newblock


\bibitem[Yariv et~al\mbox{.}(2021)]%
        {sdf_volsdf}
\bibfield{author}{\bibinfo{person}{Lior Yariv}, \bibinfo{person}{Jiatao Gu},
  \bibinfo{person}{Yoni Kasten}, {and} \bibinfo{person}{Yaron Lipman}.}
  \bibinfo{year}{2021}\natexlab{}.
\newblock \showarticletitle{Volume rendering of neural implicit surfaces}.
\newblock \bibinfo{journal}{\emph{Advances in Neural Information Processing
  Systems}}  \bibinfo{volume}{34} (\bibinfo{year}{2021}).
\newblock


\bibitem[Yariv et~al\mbox{.}(2020)]%
        {sdf_IDR}
\bibfield{author}{\bibinfo{person}{Lior Yariv}, \bibinfo{person}{Yoni Kasten},
  \bibinfo{person}{Dror Moran}, \bibinfo{person}{Meirav Galun},
  \bibinfo{person}{Matan Atzmon}, \bibinfo{person}{Basri Ronen}, {and}
  \bibinfo{person}{Yaron Lipman}.} \bibinfo{year}{2020}\natexlab{}.
\newblock \showarticletitle{Multiview neural surface reconstruction by
  disentangling geometry and appearance}.
\newblock \bibinfo{journal}{\emph{Advances in Neural Information Processing
  Systems}}  \bibinfo{volume}{33} (\bibinfo{year}{2020}),
  \bibinfo{pages}{2492--2502}.
\newblock


\bibitem[Yu et~al\mbox{.}(2021)]%
        {volume_plenoxels}
\bibfield{author}{\bibinfo{person}{Alex Yu}, \bibinfo{person}{Sara
  Fridovich-Keil}, \bibinfo{person}{Matthew Tancik}, \bibinfo{person}{Qinhong
  Chen}, \bibinfo{person}{Benjamin Recht}, {and} \bibinfo{person}{Angjoo
  Kanazawa}.} \bibinfo{year}{2021}\natexlab{}.
\newblock \showarticletitle{Plenoxels: Radiance Fields without Neural
  Networks}.
\newblock \bibinfo{journal}{\emph{arXiv preprint arXiv:2112.05131}}
  (\bibinfo{year}{2021}).
\newblock


\bibitem[Zhang et~al\mbox{.}(2022a)]%
        {zhang2022fdnerf}
\bibfield{author}{\bibinfo{person}{Jingbo Zhang}, \bibinfo{person}{Xiaoyu Li},
  \bibinfo{person}{Ziyu Wan}, \bibinfo{person}{Can Wang}, {and}
  \bibinfo{person}{Jing Liao}.} \bibinfo{year}{2022}\natexlab{a}.
\newblock \showarticletitle{FDNeRF: Few-shot Dynamic Neural Radiance Fields for
  Face Reconstruction and Expression Editing}.
\newblock \bibinfo{journal}{\emph{arXiv preprint arXiv:2208.05751}}
  (\bibinfo{year}{2022}).
\newblock


\bibitem[Zhang et~al\mbox{.}(2022b)]%
        {mesh_neuralasset}
\bibfield{author}{\bibinfo{person}{Longwen Zhang}, \bibinfo{person}{Chuxiao
  Zeng}, \bibinfo{person}{Qixuan Zhang}, \bibinfo{person}{Hongyang Lin},
  \bibinfo{person}{Ruixiang Cao}, \bibinfo{person}{Wei Yang},
  \bibinfo{person}{Lan Xu}, {and} \bibinfo{person}{Jingyi Yu}.}
  \bibinfo{year}{2022}\natexlab{b}.
\newblock \showarticletitle{Video-driven Neural Physically-based Facial Asset
  for Production}.
\newblock \bibinfo{journal}{\emph{arXiv preprint arXiv:2202.05592}}
  (\bibinfo{year}{2022}).
\newblock


\bibitem[Zheng and Xu(2021)]%
        {R2_DTexFusion}
\bibfield{author}{\bibinfo{person}{Chengwei Zheng} {and} \bibinfo{person}{Feng
  Xu}.} \bibinfo{year}{2021}\natexlab{}.
\newblock \showarticletitle{DTexFusion: Dynamic Texture Fusion using a Consumer
  RGBD Sensor}.
\newblock \bibinfo{journal}{\emph{IEEE Transactions on Visualization and
  Computer Graphics}} (\bibinfo{year}{2021}).
\newblock


\bibitem[Zheng et~al\mbox{.}(2022)]%
        {NeRFace_ImFace}
\bibfield{author}{\bibinfo{person}{Mingwu Zheng}, \bibinfo{person}{Hongyu
  Yang}, \bibinfo{person}{Di Huang}, {and} \bibinfo{person}{Liming Chen}.}
  \bibinfo{year}{2022}\natexlab{}.
\newblock \showarticletitle{ImFace: A Nonlinear 3D Morphable Face Model with
  Implicit Neural Representations}.
\newblock \bibinfo{journal}{\emph{arXiv preprint arXiv:2203.14510}}
  (\bibinfo{year}{2022}).
\newblock


\bibitem[Zheng et~al\mbox{.}(2021)]%
        {NeRFace_IMAvatar}
\bibfield{author}{\bibinfo{person}{Yufeng Zheng},
  \bibinfo{person}{Victoria~Fern{\'a}ndez Abrevaya}, \bibinfo{person}{Xu Chen},
  \bibinfo{person}{Marcel~C B{\"u}hler}, \bibinfo{person}{Michael~J Black},
  {and} \bibinfo{person}{Otmar Hilliges}.} \bibinfo{year}{2021}\natexlab{}.
\newblock \showarticletitle{IM Avatar: Implicit Morphable Head Avatars from
  Videos}.
\newblock \bibinfo{journal}{\emph{arXiv preprint arXiv:2112.07471}}
  (\bibinfo{year}{2021}).
\newblock


\bibitem[Zhou et~al\mbox{.}(2018)]%
        {MPI_stereomag}
\bibfield{author}{\bibinfo{person}{Tinghui Zhou}, \bibinfo{person}{Richard
  Tucker}, \bibinfo{person}{John Flynn}, \bibinfo{person}{Graham Fyffe}, {and}
  \bibinfo{person}{Noah Snavely}.} \bibinfo{year}{2018}\natexlab{}.
\newblock \showarticletitle{Stereo magnification: Learning view synthesis using
  multiplane images}.
\newblock \bibinfo{journal}{\emph{arXiv preprint arXiv:1805.09817}}
  (\bibinfo{year}{2018}).
\newblock


\end{thebibliography}

\appendix

% put supplementary here



\end{document}